\documentclass{article} %
\usepackage{iclr2027_conference,times}

\usepackage{amsmath,amsfonts,bm}

\def\eqref#1{equation~\ref{#1}}
\def\Eqref#1{Equation~\ref{#1}}

\def\1{\bm{1}}

\DeclareMathAlphabet{\mathsfit}{\encodingdefault}{\sfdefault}{m}{sl}
\SetMathAlphabet{\mathsfit}{bold}{\encodingdefault}{\sfdefault}{bx}{n}

\usepackage{amssymb}
\usepackage{amsfonts}
\usepackage{mathrsfs}
\usepackage{xspace}
\usepackage{bm}
\usepackage{upgreek}

\newcommand{\safemath}[2]{\newcommand{#1}{\ensuremath{#2}\xspace}}

\safemath{\bma}{\mathbf{a}}
\safemath{\bmb}{\mathbf{b}}
\safemath{\bmc}{\mathbf{c}}
\safemath{\bmd}{\mathbf{d}}
\safemath{\bme}{\mathbf{e}}
\safemath{\bmf}{\mathbf{f}}
\safemath{\bmg}{\mathbf{g}}
\safemath{\bmh}{\mathbf{h}}
\safemath{\bmi}{\mathbf{i}}
\safemath{\bmj}{\mathbf{j}}
\safemath{\bmk}{\mathbf{k}}
\safemath{\bml}{\mathbf{l}}
\safemath{\bmm}{\mathbf{m}}
\safemath{\bmn}{\mathbf{n}}
\safemath{\bmo}{\mathbf{o}}
\safemath{\bmp}{\mathbf{p}}
\safemath{\bmq}{\mathbf{q}}
\safemath{\bmr}{\mathbf{r}}
\safemath{\bms}{\mathbf{s}}
\safemath{\bmt}{\mathbf{t}}
\safemath{\bmu}{\mathbf{u}}
\safemath{\bmv}{\mathbf{v}}
\safemath{\bmw}{\mathbf{w}}
\safemath{\bmx}{\mathbf{x}}
\safemath{\bmy}{\mathbf{y}}
\safemath{\bmz}{\mathbf{z}}
\safemath{\bmzero}{\mathbf{0}}
\safemath{\bmone}{\mathbf{1}}

\safemath{\bmbeta}{\mathbf{beta}}

\bmdefine{\biad}{a}
\bmdefine{\bibd}{b}
\bmdefine{\bicd}{c}
\bmdefine{\bidd}{d}
\bmdefine{\bied}{e}
\bmdefine{\bifd}{f}
\bmdefine{\bigd}{g}
\bmdefine{\bihd}{h}
\bmdefine{\biid}{i}
\bmdefine{\bijd}{j}
\bmdefine{\bikd}{k}
\bmdefine{\bild}{l}
\bmdefine{\bimd}{m}
\bmdefine{\bind}{n}
\bmdefine{\biod}{o}
\bmdefine{\bipd}{p}
\bmdefine{\biqd}{q}
\bmdefine{\bird}{r}
\bmdefine{\bisd}{s}
\bmdefine{\bitd}{t}
\bmdefine{\biud}{u}
\bmdefine{\bivd}{v}
\bmdefine{\biwd}{w}
\bmdefine{\bixd}{x}
\bmdefine{\biyd}{y}
\bmdefine{\bizd}{z}

\bmdefine{\bixid}{\xi}
\bmdefine{\bilambdad}{\lambda}
\bmdefine{\bimud}{\mu}
\bmdefine{\bithetad}{\theta}
\bmdefine{\biphid}{\phi}
\bmdefine{\bialphad}{\alpha}

\safemath{\bmia}{\biad}
\safemath{\bmib}{\bibd}
\safemath{\bmic}{\bicd}
\safemath{\bmid}{\bidd}
\safemath{\bmie}{\bied}
\safemath{\bmif}{\bifd}
\safemath{\bmig}{\bigd}
\safemath{\bmih}{\bihd}
\safemath{\bmii}{\biid}
\safemath{\bmij}{\bijd}
\safemath{\bmik}{\bikd}
\safemath{\bmil}{\bild}
\safemath{\bmim}{\bimd}
\safemath{\bmin}{\bind}
\safemath{\bmio}{\biod}
\safemath{\bmip}{\bipd}
\safemath{\bmiq}{\biqd}
\safemath{\bmir}{\bird}
\safemath{\bmis}{\bisd}
\safemath{\bmit}{\bitd}
\safemath{\bmiu}{\biud}
\safemath{\bmiv}{\bivd}
\safemath{\bmiw}{\biwd}
\safemath{\bmix}{\bixd}
\safemath{\bmiy}{\biyd}
\safemath{\bmiz}{\bizd}

\safemath{\bmxi}{\bixid}
\safemath{\bmlambda}{\bilambdad}
\safemath{\bmalpha}{\bialphad}
\safemath{\bmmu}{\bimud}
\safemath{\bmtheta}{\bithetad}
\safemath{\bmphi}{\biphid}

\safemath{\bA}{\mathbf{A}}
\safemath{\bB}{\mathbf{B}}
\safemath{\bC}{\mathbf{C}}
\safemath{\bD}{\mathbf{D}}
\safemath{\bE}{\mathbf{E}}
\safemath{\bF}{\mathbf{F}}
\safemath{\bG}{\mathbf{G}}
\safemath{\bH}{\mathbf{H}}
\safemath{\bI}{\mathbf{I}}
\safemath{\bJ}{\mathbf{J}}
\safemath{\bK}{\mathbf{K}}
\safemath{\bL}{\mathbf{L}}
\safemath{\bM}{\mathbf{M}}
\safemath{\bN}{\mathbf{N}}
\safemath{\bO}{\mathbf{O}}
\safemath{\bP}{\mathbf{P}}
\safemath{\bQ}{\mathbf{Q}}
\safemath{\bR}{\mathbf{R}}
\safemath{\bS}{\mathbf{S}}
\safemath{\bT}{\mathbf{T}}
\safemath{\bU}{\mathbf{U}}
\safemath{\bV}{\mathbf{V}}
\safemath{\bW}{\mathbf{W}}
\safemath{\bX}{\mathbf{X}}
\safemath{\bY}{\mathbf{Y}}
\safemath{\bZ}{\mathbf{Z}}

\safemath{\bZero}{\mathbf{0}}
\safemath{\bOne}{\mathbf{1}}
\safemath{\bDelta}{\mathbf{\Delta}}
\safemath{\bLambda}{\mathbf{\UpLambda}}
\safemath{\bPhi}{\mathbf{\Upphi}}
\safemath{\bSigma}{\mathbf{\Upsigma}}
\safemath{\bOmega}{\mathbf{\Upomega}}
\safemath{\bTheta}{\mathbf{\Uptheta}}

\bmdefine{\biAd}{A}
\bmdefine{\biBd}{B}
\bmdefine{\biCd}{C}
\bmdefine{\biDd}{D}
\bmdefine{\biEd}{E}
\bmdefine{\biFd}{F}
\bmdefine{\biGd}{G}
\bmdefine{\biHd}{H}
\bmdefine{\biId}{I}
\bmdefine{\biJd}{J}
\bmdefine{\biKd}{K}
\bmdefine{\biLd}{L}
\bmdefine{\biMd}{M}
\bmdefine{\biOd}{N}
\bmdefine{\biPd}{O}
\bmdefine{\biQd}{P}
\bmdefine{\biRd}{R}
\bmdefine{\biSd}{S}
\bmdefine{\biTd}{T}
\bmdefine{\biUd}{U}
\bmdefine{\biVd}{V}
\bmdefine{\biWd}{W}
\bmdefine{\biXd}{X}
\bmdefine{\biYd}{Y}
\bmdefine{\biZd}{Z}

\bmdefine{\biDelta}{\Delta}
\bmdefine{\biLambda}{\Lambda}
\bmdefine{\biPhi}{\Phi}
\bmdefine{\biSigma}{\Sigma}
\bmdefine{\biOmega}{\Omega}
\bmdefine{\biTheta}{\Theta}

\safemath{\bimA}{\biAd}
\safemath{\bimB}{\biBd}
\safemath{\bimC}{\biCd}
\safemath{\bimD}{\biDd}
\safemath{\bimE}{\biEd}
\safemath{\bimF}{\biFd}
\safemath{\bimG}{\biGd}
\safemath{\bimH}{\biHd}
\safemath{\bimI}{\biId}
\safemath{\bimJ}{\biJd}
\safemath{\bimK}{\biKd}
\safemath{\bimL}{\biLd}
\safemath{\bimM}{\biMd}
\safemath{\bimN}{\biNd}
\safemath{\bimO}{\biOd}
\safemath{\bimP}{\biPd}
\safemath{\bimQ}{\biQd}
\safemath{\bimR}{\biRd}
\safemath{\bimS}{\biSd}
\safemath{\bimT}{\biTd}
\safemath{\bimU}{\biUd}
\safemath{\bimV}{\biVd}
\safemath{\bimW}{\biWd}
\safemath{\bimX}{\biXd}
\safemath{\bimY}{\biYd}
\safemath{\bimZ}{\biZd}

\safemath{\bimDelta}{\biDelta}
\safemath{\bimLambda}{\biLambda}
\safemath{\bimPhi}{\biPhi}
\safemath{\bimSigma}{\biSigma}
\safemath{\bimOmega}{\biOmega}
\safemath{\bimTheta}{\biTheta}

\safemath{\setA}{\mathcal{A}}
\safemath{\setB}{\mathcal{B}}
\safemath{\setC}{\mathcal{C}}
\safemath{\setD}{\mathcal{D}}
\safemath{\setE}{\mathcal{E}}
\safemath{\setF}{\mathcal{F}}
\safemath{\setG}{\mathcal{G}}
\safemath{\setH}{\mathcal{H}}
\safemath{\setI}{\mathcal{I}}
\safemath{\setJ}{\mathcal{J}}
\safemath{\setK}{\mathcal{K}}
\safemath{\setL}{\mathcal{L}}
\safemath{\setM}{\mathcal{M}}
\safemath{\setN}{\mathcal{N}}
\safemath{\setO}{\mathcal{O}}
\safemath{\setP}{\mathcal{P}}
\safemath{\setQ}{\mathcal{Q}}
\safemath{\setR}{\mathcal{R}}
\safemath{\setS}{\mathcal{S}}
\safemath{\setT}{\mathcal{T}}
\safemath{\setU}{\mathcal{U}}
\safemath{\setV}{\mathcal{V}}
\safemath{\setW}{\mathcal{W}}
\safemath{\setX}{\mathcal{X}}
\safemath{\setY}{\mathcal{Y}}
\safemath{\setZ}{\mathcal{Z}}
\safemath{\emptySet}{\varnothing}

\safemath{\colA}{\mathscr{A}}
\safemath{\colB}{\mathscr{B}}
\safemath{\colC}{\mathscr{C}}
\safemath{\colD}{\mathscr{D}}
\safemath{\colE}{\mathscr{E}}
\safemath{\colF}{\mathscr{F}}
\safemath{\colG}{\mathscr{G}}
\safemath{\colH}{\mathscr{H}}
\safemath{\colI}{\mathscr{I}}
\safemath{\colJ}{\mathscr{J}}
\safemath{\colK}{\mathscr{K}}
\safemath{\colL}{\mathscr{L}}
\safemath{\colM}{\mathscr{M}}
\safemath{\colN}{\mathscr{N}}
\safemath{\colO}{\mathscr{O}}
\safemath{\colP}{\mathscr{P}}
\safemath{\colQ}{\mathscr{Q}}
\safemath{\colR}{\mathscr{R}}
\safemath{\colS}{\mathscr{S}}
\safemath{\colT}{\mathscr{T}}
\safemath{\colU}{\mathscr{U}}
\safemath{\colV}{\mathscr{V}}
\safemath{\colW}{\mathscr{W}}
\safemath{\colX}{\mathscr{X}}
\safemath{\colY}{\mathscr{Y}}
\safemath{\colZ}{\mathscr{Z}}

\safemath{\opA}{\mathbb{A}}
\safemath{\opB}{\mathbb{B}}
\safemath{\opC}{\mathbb{C}}
\safemath{\opD}{\mathbb{D}}
\safemath{\opE}{\mathbb{E}}
\safemath{\opF}{\mathbb{F}}
\safemath{\opG}{\mathbb{G}}
\safemath{\opH}{\mathbb{H}}
\safemath{\opI}{\mathbb{I}}
\safemath{\opJ}{\mathbb{J}}
\safemath{\opK}{\mathbb{K}}
\safemath{\opL}{\mathbb{L}}
\safemath{\opM}{\mathbb{M}}
\safemath{\opN}{\mathbb{N}}
\safemath{\opO}{\mathbb{O}}
\safemath{\opP}{\mathbb{P}}
\safemath{\opQ}{\mathbb{Q}}
\safemath{\opR}{\mathbb{R}}
\safemath{\opS}{\mathbb{S}}
\safemath{\opT}{\mathbb{T}}
\safemath{\opU}{\mathbb{U}}
\safemath{\opV}{\mathbb{V}}
\safemath{\opW}{\mathbb{W}}
\safemath{\opX}{\mathbb{X}}
\safemath{\opY}{\mathbb{Y}}
\safemath{\opZ}{\mathbb{Z}}
\safemath{\opZero}{\mathbb{O}}
\safemath{\identityop}{\opI}

\safemath{\veca}{\bma}
\safemath{\vecb}{\bmb}
\safemath{\vecc}{\bmc}
\safemath{\vecd}{\bmd}
\safemath{\vece}{\bme}
\safemath{\vecf}{\bmf}
\safemath{\vecg}{\bmg}
\safemath{\vech}{\bmh}
\safemath{\veci}{\bmi}
\safemath{\vecj}{\bmj}
\safemath{\veck}{\bmk}
\safemath{\vecl}{\bml}
\safemath{\vecm}{\bmm}
\safemath{\vecn}{\bmn}
\safemath{\veco}{\bmo}
\safemath{\vecp}{\bmmp}
\safemath{\vecq}{\bmq}
\safemath{\vecr}{\bmr}
\safemath{\vecs}{\bms}
\safemath{\vect}{\bmt}
\safemath{\vecu}{\bmu}
\safemath{\vecv}{\bmv}
\safemath{\vecw}{\bmw}
\safemath{\vecx}{\bmx}
\safemath{\vecy}{\bmy}
\safemath{\vecz}{\bmz}

\safemath{\veczero}{\bmzero}
\safemath{\vecone}{\bmone}
\safemath{\vecxi}{\bmxi}
\safemath{\veclambda}{\bmlambda}
\safemath{\vecmu}{\bmmu}
\safemath{\vectheta}{\bmtheta}
\safemath{\vecphi}{\bmphi}

\safemath{\matA}{\bA}
\safemath{\matB}{\bB}
\safemath{\matC}{\bC}
\safemath{\matD}{\bD}
\safemath{\matE}{\bE}
\safemath{\matF}{\bF}
\safemath{\matG}{\bG}
\safemath{\matH}{\bH}
\safemath{\matI}{\bI}
\safemath{\matJ}{\bJ}
\safemath{\matK}{\bK}
\safemath{\matL}{\bL}
\safemath{\matM}{\bM}
\safemath{\matN}{\bN}
\safemath{\matO}{\bO}
\safemath{\matP}{\bP}
\safemath{\matQ}{\bQ}
\safemath{\matR}{\bR}
\safemath{\matS}{\bS}
\safemath{\matT}{\bT}
\safemath{\matU}{\bU}
\safemath{\matV}{\bV}
\safemath{\matW}{\bW}
\safemath{\matX}{\bX}
\safemath{\matY}{\bY}
\safemath{\matZ}{\bZ}
\safemath{\matzero}{\bmzero}

\safemath{\matDelta}{\bDelta}
\safemath{\matLambda}{\bLambda}
\safemath{\matPhi}{\bPhi}
\safemath{\matSigma}{\bSigma}
\safemath{\matOmega}{\bOmega}
\safemath{\matTheta}{\bTheta}

\safemath{\matidentity}{\matI}
\safemath{\matone}{\matO}

\safemath{\rnda}{A}
\safemath{\rndb}{B}
\safemath{\rndc}{C}
\safemath{\rndd}{D}
\safemath{\rnde}{E}
\safemath{\rndf}{F}
\safemath{\rndg}{G}
\safemath{\rndh}{H}
\safemath{\rndi}{I}
\safemath{\rndj}{J}
\safemath{\rndk}{K}
\safemath{\rndl}{L}
\safemath{\rndm}{M}
\safemath{\rndn}{N}
\safemath{\rndo}{O}
\safemath{\rndp}{P}
\safemath{\rndq}{Q}
\safemath{\rndr}{R}
\safemath{\rnds}{S}
\safemath{\rndt}{T}
\safemath{\rndu}{U}
\safemath{\rndv}{V}
\safemath{\rndw}{W}
\safemath{\rndx}{X}
\safemath{\rndy}{Y}
\safemath{\rndz}{Z}

\safemath{\rveca}{\bimA}
\safemath{\rvecb}{\bimB}
\safemath{\rvecc}{\bimC}
\safemath{\rvecd}{\bimD}
\safemath{\rvece}{\bimE}
\safemath{\rvecf}{\bimF}
\safemath{\rvecg}{\bimG}
\safemath{\rvech}{\bimH}
\safemath{\rveci}{\bimI}
\safemath{\rvecj}{\bimJ}
\safemath{\rveck}{\bimK}
\safemath{\rvecl}{\bimL}
\safemath{\rvecm}{\bimM}
\safemath{\rvecn}{\bimN}
\safemath{\rveco}{\bomO}
\safemath{\rvecp}{\bimP}
\safemath{\rvecq}{\bimQ}
\safemath{\rvecr}{\bimR}
\safemath{\rvecs}{\bimS}
\safemath{\rvect}{\bimT}
\safemath{\rvecu}{\bimU}
\safemath{\rvecv}{\bimV}
\safemath{\rvecw}{\bimW}
\safemath{\rvecx}{\bimX}
\safemath{\rvecy}{\bimY}
\safemath{\rvecz}{\bimZ}

\safemath{\rvecxi}{\bmxi}
\safemath{\rveclambda}{\bmlambda}
\safemath{\rvecmu}{\bmmu}
\safemath{\rvectheta}{\bmtheta}
\safemath{\rvecphi}{\bmphi}

\safemath{\rmatA}{\bimA}
\safemath{\rmatB}{\bimB}
\safemath{\rmatC}{\bimC}
\safemath{\rmatD}{\bimD}
\safemath{\rmatE}{\bimE}
\safemath{\rmatF}{\bimF}
\safemath{\rmatG}{\bimG}
\safemath{\rmatH}{\bimH}
\safemath{\rmatI}{\bimI}
\safemath{\rmatJ}{\bimJ}
\safemath{\rmatK}{\bimK}
\safemath{\rmatL}{\bimL}
\safemath{\rmatM}{\bimM}
\safemath{\rmatN}{\bimN}
\safemath{\rmatO}{\bimO}
\safemath{\rmatP}{\bimP}
\safemath{\rmatQ}{\bimQ}
\safemath{\rmatR}{\bimR}
\safemath{\rmatS}{\bimS}
\safemath{\rmatT}{\bimT}
\safemath{\rmatU}{\bimU}
\safemath{\rmatV}{\bimV}
\safemath{\rmatW}{\bimW}
\safemath{\rmatX}{\bimX}
\safemath{\rmatY}{\bimY}
\safemath{\rmatZ}{\bimZ}

\safemath{\rmatDelta}{\bimDelta}
\safemath{\rmatLambda}{\bimLambda}
\safemath{\rmatPhi}{\bimPhi}
\safemath{\rmatSigma}{\bimSigma}
\safemath{\rmatOmega}{\bimOmega}
\safemath{\rmatTheta}{\bimTheta}
\usepackage{amssymb}
\usepackage{amsfonts}
\usepackage{mathrsfs}
\usepackage{xspace}
\usepackage{bm}

\newcommand{\methodname}{LoReS}

\usepackage{scalerel}

\newenvironment{textbmatrix}{	\setlength{\arraycolsep}{2.5pt}%
								\big[\begin{matrix}}{\end{matrix}\big]%
								\raisebox{0.08ex}{\vphantom{M}}}

\def\be{\begin{equation}}
\def\een{\nonumber \end{equation}}
\def\mat{\begin{bmatrix}}
\def\emat{\end{bmatrix}}
\def\btm{\begin{textbmatrix}}
\def\etm{\end{textbmatrix}}

\def\ba#1\ea{\begin{align}#1\end{align}}
\def\bas#1\eas{\begin{align*}#1\end{align*}}
\def\bs#1\es{\begin{split}#1\end{split}} 
\def\bg#1\eg{\begin{gather}#1\end{gather}} 
\def\bi#1\ei{\begin{itemize}#1\end{itemize}}

\safemath{\dirac}{\delta}					%
\safemath{\krond}{\dirac}					%
\safemath{\upto}{\uparrow}
\safemath{\downto}{\downarrow}
\safemath{\iu}{j}							%
\safemath{\ev}{\lambda}						%
\safemath{\hilseqspace}{l^{2}}				%
\newcommand{\banachfunspace}[1]{\setL^{#1}}	%
\safemath{\hilfunspace}{\banachfunspace{2}}	%

\safemath{\SNR}{\text{\sc snr}} 				%
\safemath{\No}{N_0}							%
\safemath{\Es}{E_s}							%
\safemath{\Eb}{E_b}							%
\safemath{\EbNo}{\frac{\Eb}{\No}}
\safemath{\EsNo}{\frac{\Es}{\No}}

\DeclareMathOperator{\CHop}{\ensuremath{\opH}} %
\safemath{\tvir}{\rndh_{\CHop}}				%
\safemath{\tvtf}{\rndl_{\CHop}}				%
\safemath{\spf}{\rnds_{\CHop}}				%
\safemath{\bff}{H_{\CHop}}					%

\safemath{\ircf}{r_{h}}						%
\safemath{\tftvcf}{r_{s}}					%
\safemath{\tfcf}{r_{l}}						%
\safemath{\bfcf}{r_{H}}						%

\safemath{\tcorr}{c_h}						%
\safemath{\scf}{c_{s}}						%
\safemath{\tfcorr}{c_{l}}					%
\safemath{\fcorr}{c_{H}}						%

\safemath{\mi}{I}							%
\safemath{\capacity}{C}						%

\safemath{\normal}{\mathcal{N}}			%
\safemath{\jpg}{\mathcal{CN}}			%
\safemath{\mchain}{\leftrightarrow}		%

\safemath{\dB}{\,\mathrm{dB}}
\safemath{\dBm}{\,\mathrm{dBm}}
\safemath{\Hz}{\,\mathrm{Hz}}
\safemath{\kHz}{\,\mathrm{kHz}}
\safemath{\MHz}{\,\mathrm{MHz}}
\safemath{\GHz}{\,\mathrm{GHz}}
\safemath{\s}{\,\mathrm{s}}
\safemath{\ms}{\,\mathrm{ms}}
\safemath{\mus}{\,\mathrm{\mu s}}
\safemath{\ns}{\,\mathrm{ns}}
\safemath{\meter}{\,\mathrm{m}}
\safemath{\mm}{\,\mathrm{mm}}
\safemath{\cm}{\,\mathrm{cm}}
\safemath{\m}{\,\mathrm{m}}
\safemath{\W}{\,\mathrm{W}}
\safemath{\J}{\,\mathrm{J}}
\safemath{\K}{\,\mathrm{K}}
\safemath{\bit}{\,\mathrm{bit}}

\safemath{\define}{=}			%

\safemath{\equivalent}{\sim}
\safemath{\distas}{\sim}					%
\safemath{\sdiff}{\Delta}				%

\safemath{\reals}{\mathbb{R}}
\safemath{\positivereals}{\reals_{+}}
\safemath{\integers}{\mathbb{Z}}
\safemath{\posint}{\integers_{+}}
\safemath{\naturals}{\mathbb{N}}
\safemath{\posnaturals}{\naturals_{+}}
\safemath{\complexset}{\mathbb{C}}
\safemath{\rationals}{\mathbb{Q}}

\usepackage{xcolor}
\definecolor{linkblue}{RGB}{40,80,190}

\usepackage[
    colorlinks=true,
    linkcolor=linkblue,
    citecolor=linkblue,
    urlcolor=linkblue
]{hyperref}

\usepackage{url}
\usepackage{booktabs}
\usepackage{array}
\usepackage{multirow}
\usepackage{graphicx}
\usepackage{wrapfig}
\usepackage{pgf}
\usepackage{tikz}
\usetikzlibrary{arrows.meta,calc,positioning}
\graphicspath{{media/}}  %
\usepackage{amsmath}
\usepackage{amssymb}
\usepackage{amsthm}
\usepackage{mathtools}
\usepackage{pifont}
\usepackage{fontawesome5}
\usepackage{enumitem}

\newtheorem{proposition}{Proposition}

\usepackage{subcaption}
\usepackage{xcolor}
\usepackage{colortbl}
\usepackage[section]{placeins}
\usepackage{needspace}
\usepackage{tabularx}
\newcommand{\numhead}[1]{\multicolumn{1}{c}{#1}}

\title{\centering Quantizing Recurrent Language \\ Models Is Easier Than You Think}

\author{%
\parbox[t]{\dimexpr\textwidth-2\tabcolsep\relax}{\centering
Philipp Nazari\textsuperscript{*,1,2,3,4}\quad
Neehal Tumma\textsuperscript{*,5,6}\quad
Daniela Rus\textsuperscript{5,6}\quad
T. Konstantin Rusch\textsuperscript{1,3,4,6}\\[0.5em]
\normalfont\small
\textsuperscript{1}Max Planck Institute for Intelligent Systems\quad
\textsuperscript{2}ETH Z\"urich\quad
\textsuperscript{3}T\"ubingen AI Center\\
\textsuperscript{4}ELLIS Institute T\"ubingen\quad
\textsuperscript{5}MIT CSAIL \quad
\textsuperscript{6}Liquid AI\\[0.3em]
\textsuperscript{*}Equal contribution \quad Correspondence: {philipp.nazari@tuebingen.mpg.de}.
}}

\newcommand{\WT}{WikiText-2}

\newcommand{\rout}{\bR_{o}}

\iclrfinalcopy
\begin{document}

\maketitle
\fancyhead{}
\fancyhead[L]{\footnotesize\normalfont Quantizing Recurrent Language Models Is Easier Than You Think}
\renewcommand{\headrulewidth}{0.4pt}

\begin{abstract}
Linear recurrences are becoming an increasingly common building block in large language models, replacing the growing KV cache of softmax attention with a fixed-size recurrent state that enables efficient decoding. However, the quantization stack has largely been developed around softmax Transformers, whose numerical properties and inference state differ from those of recurrent models. Our goal is to understand these differences and study the design space of quantization methods for recurrent models. In our analysis, we differentiate between two targets. First, extra-state quantization concerns weights and activations outside the recurrent state, whose distributions are shaped indirectly by the sequence mixer during training. In controlled 1.3B parameter models, we find that decayed recurrences develop weaker residual outlier channels than softmax attention and are therefore easier to quantize without incoherence processing. Second, intra-state quantization concerns the recurrent state itself during decode, where repeated updates create a different numerical problem from KV cache quantization. We instantiate our findings as \methodname{}, a state quantization pipeline combining value-axis scale grouping to match output channel magnitude, decay-based mixed-precision to protect slowly decaying heads or channels, and stochastic rounding to limit error accumulation under repeated requantization. Across Qwen-3.5 9B, Nemotron-H 8B, and Kimi-Linear 48B-A3B, \methodname{} supports robust 8-bit state quantization and lower effective precision on shorter decode horizons. We implement \methodname{} in vLLM and demonstrate end-to-end decode speedups.\looseness-1
\end{abstract}

\vspace{-1em}
{\centering\small
\href{https://github.com/camail-official/LoReS}{\faGithub\enspace\nolinkurl{github.com/camail-official/LoReS}}\par}

\section{Introduction}
\label{sec:introduction}

Softmax attention has shaped the architecture of large language models, as well as the deployment stack built around them. This is particularly true for post-training quantization (PTQ). A large body of prior work investigates quantization of the weights and activations of softmax Transformers~\citep{frantar2022gptq,xiao2023smoothquant,ashkboos2024quarot,liu2025spinquant}, while a complementary line studies quantization of the key-value cache that grows during autoregressive generation~\citep{liu2024kivi,hooper2024kvquant,duanmu2024skvq}.

Concurrently, the space of sequence mixers used in foundation models is expanding. Linear attention and state-space models replace softmax attention with linear recurrences~\citep{katharopoulos2020transformers,dao2024transformers,yang2023gated,yang2025gated,pdn}, and recent large-scale hybrid models interleave recurrent and softmax layers within the same architecture~\citep{blakeman2025nemotron,team2025kimi,team2026qwen3}. As these hybrid models become increasingly common, the inference stack developed for softmax Transformers must be adapted accordingly.

Compared with softmax attention, however, the behavior of recurrent language models under quantization remains poorly understood. Existing work has begun to adapt quantization to recurrent architectures~\citep{chiang2025quamba,chiang2025quamba2,yue2025mambaquant,kim2026ssdi8,chandiramani2026nemotron}, but has focused on particular architectures and quantization targets. In this work, we instead ask the more general question: \emph{how does replacing softmax attention with a linear recurrent sequence mixer change the quantization problem?} We argue that this replacement matters along two distinct axes. First, the choice of sequence mixer changes the training dynamics of the model and therefore the distributions of the weights and activations learned throughout the network. We refer to quantization of these tensors as \emph{extra-state weight/activation (W/A) quantization}. Second, softmax attention and recurrent mixers maintain fundamentally different objects during inference. Softmax attention stores a growing cache of keys and values, whereas a recurrent mixer repeatedly updates a fixed-size state in place. We refer to quantization of this object as \emph{(intra-)state quantization}. Together, these two settings separate the implicit effect of the sequence mixer on the rest of the model from the explicit effect of quantizing the mixer state itself.
Our main contributions can be summarized as follows:\looseness-1
\begin{itemize}[leftmargin=15pt, labelindent=0pt, parsep=1pt, topsep=1pt, partopsep=1pt, itemsep=2pt]
    \item For extra-state W/A quantization, we show that decayed linear recurrences develop substantially weaker residual outlier channels than softmax attention, making them more robust to W/A quantization before incoherence processing. Incoherence processing largely closes this gap.

    \item For intra-state quantization, we analyze how scale grouping, recurrent decay, rounding rule, and storage format impact quantization error, and reveal failure modes of random rotations.\looseness-1

    \item We validate these findings in controlled 1.3B-parameter experiments, and demonstrate their transfer to the hybrids Qwen-3.5 9B, Nemotron-H 8B, and Kimi-Linear 48B-A3B. We instantiate our findings as \methodname{} (\textbf{Lo}w-bit \textbf{Re}current \textbf{S}tate), combining value-axis grouping, stochastic rounding and decay-based mixed precision for integer state quantization.

    \item We implement \methodname{} in vLLM, demonstrating end-to-end decode speedups at scale.
\end{itemize}

\definecolor{decaycol}{RGB}{0,114,178}    %
\definecolor{gatecol}{RGB}{230,159,0}     %
\definecolor{deltacol}{RGB}{213,94,0}     %
\newcommand{\dcol}[1]{{\color{decaycol}#1}}
\newcommand{\gcol}[1]{{\color{gatecol}#1}}
\newcommand{\ecol}[1]{{\color{deltacol}#1}}

\begin{table}[t!]
\centering
\small
\setlength{\tabcolsep}{3pt}
\renewcommand{\arraystretch}{1.45}
\caption{State-update rules for recurrent models in the mixer-zoo design space. Colors identify \dcol{decay}, \gcol{write gating}, and \ecol{delta correction}.}
\label{tab:zoo-updates}
\begin{tabular}{ll}
\toprule
\multicolumn{1}{c}{model} & \multicolumn{1}{c}{update}\\
\midrule
Linear attention~\citep{katharopoulos2020transformers}
    & $\bS_t = \bS_{t-1} + \bmv_t \bmk_t^\top$ \\

Mamba-2~\citep{dao2024transformers}
    & $\bS_t = \dcol{\alpha_t}\, \bS_{t-1} + \bmv_t \bmk_t^\top$ \\

Gated Linear Attention (GLA)~\citep{yang2023gated}
    & $\bS_t = \bS_{t-1}\, \dcol{\operatorname{diag}(\bmalpha_t)} + \bmv_t \bmk_t^\top$ \\

DeltaNet~\citep{schlag2021linear,yang2024parallelizing}
    & $\bS_t = \bS_{t-1}\, \ecol{\bigl(\bI - \beta_t \bmk_t \bmk_t^\top\bigr)} + \gcol{\beta_t}\, \bmv_t \bmk_t^\top$ \\

Gated DeltaNet~\citep{yang2025gated}
    & $\bS_t = \dcol{\alpha_t}\, \bS_{t-1}\, \ecol{\bigl(\bI - \beta_t \bmk_t \bmk_t^\top\bigr)} + \gcol{\beta_t}\, \bmv_t \bmk_t^\top$ \\

Kimi Delta Attention (KDA)~\citep{team2025kimi}
    & $\bS_t = \bS_{t-1}\, \dcol{\operatorname{diag}(\bmalpha_t)}\, \ecol{\bigl(\bI - \beta_t \bmk_t \bmk_t^\top\bigr)} + \gcol{\beta_t}\, \bmv_t \bmk_t^\top$ \\
\bottomrule
\end{tabular}
\vspace{-1em}
\end{table}

\section{Background and preliminaries}
\label{sec:setup}

\textbf{Softmax attention.}
Given queries $\bQ \in \mathbb{R}^{T, d_k}$, keys
$\bK \in \mathbb{R}^{T, d_k}$, and values
$\bV \in \mathbb{R}^{T, d_v}$, softmax self attention~\citep{vaswani2017attention}
computes
\begin{equation*}
    \bmo_t
    =
    \bV_{1:t}^{\top}
    \operatorname{softmax}
    \left(
        \bK_{1:t}\bmq_t/\sqrt{d_k}
    \right).
\vspace{-0.2em}
\end{equation*}
During autoregressive generation, the model stores previously computed keys and values in a key-value (KV) cache and reuses them at subsequent decode steps. As a result, the sequence mixer state grows linearly with sequence length (Figure~\ref{fig:sequence-mixer-quantization}).

\textbf{Linear attention.}
Linear attention replaces the nonlinear softmax operator with an unnormalized linear kernel~\citep{katharopoulos2020transformers,schlag2021linear}. In its simplest form, associativity allows keys and values to be summarized into a fixed-size matrix-valued state $\bS_t \in \mathbb{R}^{d_v, d_k}$, satisfying
\begin{equation*}
    \bS_t
    =
    \bS_{t-1}
    +
    \bmv_t \bmk_t^\top,
    \qquad
    \bmo_t
    =
    \bS_t \bmq_t.
\end{equation*}
Unlike the KV cache, this recurrent state $\mathbf S_t$ has constant size and is updated in place after every token. Modern recurrent sequence mixers refine this basic recurrence through mechanisms such as forgetting, gated writes, and delta-rule corrections~\citep{dao2024transformers,yang2023gated,yang2025gated,team2025kimi}. To isolate these architectural choices, we construct a \emph{mixer zoo} in which the sequence mixer varies while the surrounding layer design and training recipe are fixed. In particular, all models share the backbone around the mixer, including the query, key, and value projections, causal convolutions, normalization, RoPE~\citep{su2024roformer}, and output projection, apart from the explicitly varied query/key normalization (which we denote by $\|\mathbf Q \mathbf K\|$). The recurrent variants in the zoo follow\looseness-1
\begin{equation}
\label{eq:zoo-recurrence}
    \bS_t
    =
    \bS_{t-1}\,
    \dcol{\operatorname{diag}(\bmalpha_t)}\,
    \ecol{\bigl(\bI - \delta\,\beta_t\,\bmk_t\bmk_t^\top\bigr)}
    +
    \gcol{\beta_t}\,\bmv_t\bmk_t^\top, \qquad \mathbf o_t = \mathbf S_t \mathbf q_t,
\end{equation}
where $\|\bmk_t\|_2 = \|\mathbf q_t\|_2 = 1$, $\beta_t\in(0,2)$, and the vector $\dcol{\bmalpha_t}$ has entries $\dcol{\alpha_{t,i}\in(0,1]}$. In our notation, the indicator $\delta\in\{0,1\}$ controls delta correction. The recurrence composes three mechanisms:\looseness-1
\begin{itemize}[leftmargin=15pt, labelindent=0pt, parsep=1pt, topsep=1pt, partopsep=1pt, itemsep=2pt]
    \item \dcol{\textbf{Decay}} is a multiplicative forget gate applied to the state before each write. A scalar gate defines one decay rate per attention head, as in Mamba-2~\citep{dao2024transformers} and Gated DeltaNet~\citep{yang2025gated}, while a diagonal gate defines a separate rate per key channel, as in Gated Linear Attention~\citep{yang2023gated} and Kimi Delta Attention~\citep{team2025kimi}.

    \item \gcol{\textbf{Write gating}} controls the strength of the outer-product update $\bmv_t\bmk_t^\top$.

    \item \ecol{\textbf{Delta correction}}, introduced in DeltaNet~\citep{schlag2021linear,yang2024parallelizing}, applies a rank-one correction along the current key through $\ecol{(\bI-\beta_t\bmk_t\bmk_t^\top)}$ before writing the new association.
\end{itemize}

Table~\ref{tab:zoo-updates} classifies linear attention models using this taxonomy. Our mixer zoo additionally contains softmax attention with and without query/key RMSNorm~\citep{zhang2019root}.

\subsection{Quantizing Linear Attention}
\label{sec:quantizing-mixers}

\begin{wrapfigure}[25]{r}{0.60\textwidth}
\vspace{-\intextsep}
\centering
\input{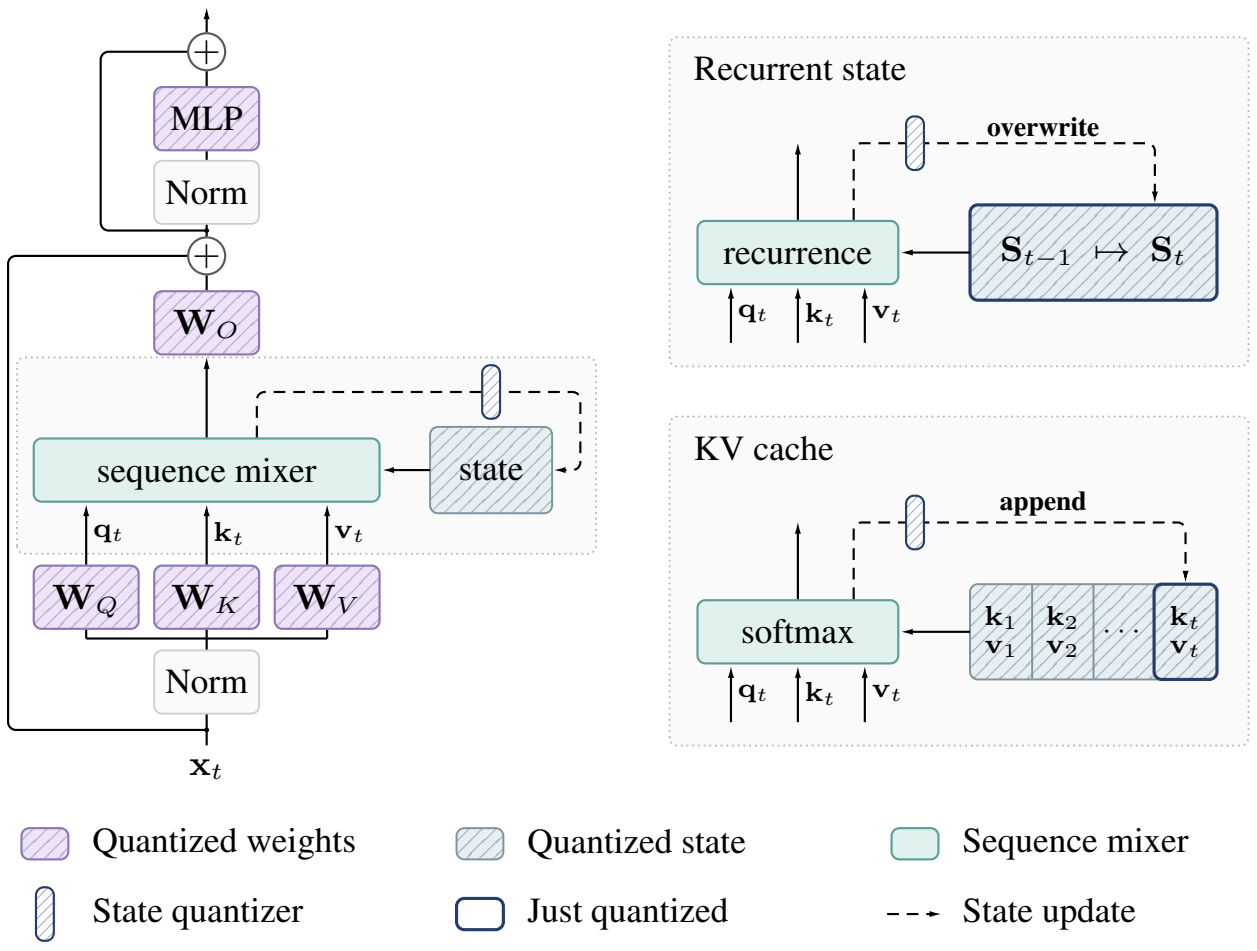}
\caption{We consider two quantization targets. Extra-state quantization acts on weights and activations outside the mixer state, while intra-state quantization acts on the state used during inference. A recurrent state is repeatedly updated and requantized in place, whereas entries in a KV cache are quantized once and subsequently reused.}
\label{fig:sequence-mixer-quantization}
\end{wrapfigure}

Figure~\ref{fig:sequence-mixer-quantization} distinguishes two quantization targets. \emph{Extra-state quantization} concerns the weights and activations (W/A) outside the mixer state, while \emph{intra-state quantization} concerns the state maintained during decode. The distinction between quantizing softmax attention and linear recurrences is implicit in the former, through the representations learned during training, and explicit in the latter, through the object that is stored and updated during inference.\looseness-1

\textbf{Extra-state quantization.}
Softmax and recurrent models expose the same class of weights and activations to a W/A quantizer, so the same general PTQ machinery can be applied to both ~\citep{frantar2022gptq,xiao2023smoothquant,ashkboos2024quarot, liu2025spinquant}. However, changing the sequence mixer can change the distributions of these tensors learned during training. Prior work on Mamba models has consequently developed architecture-specific treatments for their activation statistics~\citep{chiang2025quamba,chiang2025quamba2, yue2025mambaquant}. We instead study this dependence directly: using a common quantization framework and controlled architecture, Section~\ref{sec:external} asks how the sequence mixer itself changes W/A quantization.

\textbf{Intra-state quantization.}
Here the distinction between sequence mixers is explicit. KV cache quantization methods are designed around keys and values that are appended once and then reused. They exploit this structure through choices such as separate scaling and grouping for keys and values, outlier treatment, channel reordering, and a recent token window kept at higher precision~\citep{liu2024kivi,hooper2024kvquant,duanmu2024skvq}. A linear recurrence instead stores a fixed-size matrix $\bS_t$ that is updated after every decoded token. Quantizing this state therefore requires repeatedly dequantizing, updating, and requantizing the same object, allowing quantization error to propagate temporally through subsequent updates. % The state also has two channel axes and may attenuate previously introduced error through recurrent decay.
Previous work on recurrent state quantization during decode has focused primarily on Mamba models. Q-Mamba~\citep{tianqi2025qmamba} develops scaling methods that account for channel distributions in quantized Mamba states, while Quamba2~\citep{chiang2025quamba2} stores states in INT8 during generation. Nemotron 3 Super~\citep{chandiramani2026nemotron} studies low precision state storage using stochastic rounding. SSDi8~\citep{kim2026ssdi8} considers a related but different target, quantizing the chunked Mamba-2 SSD computation used during prefill to maintain a low-precision execution path.\looseness-1

Concurrent work proposes state-quantization methods using decay-based mixed precision~\citep{zhang2026damp,yao2026stepquant}, optimized scaling~\citep{yao2026stepquant}, and reduced requantization frequency~\citep{pan2026leapquant}. Our work instead focuses on a broader analysis of quantization in linear recurrent language models, covering both extra-state W/A and intra-state quantization. Section~\ref{ch:state-quant} studies how scale grouping, recurrent decay, and repeated requantization influence quantization error. It furthermore reveals failure modes of random rotations and explores integer versus floating-point formats.\looseness-1

\section{Extra-State Weight and Activation Quantization}
\label{sec:external}
\definecolor{ezHl}{HTML}{E8F1FB}

\begin{wraptable}[21]{R}{0.58\textwidth}
\vspace{-\intextsep}
\centering
\footnotesize
\setlength{\tabcolsep}{2pt}
\renewcommand{\arraystretch}{1.12}
\caption{Relative perplexity change (\%) on \WT{} in the $1.3$B mixer zoo. The first two rows use softmax attention, $\|\mathbf Q \mathbf K\|$ denotes $QK$-norm. Subsequent rows are recurrent mixers. Rotation columns use W4A4.}
\label{tab:external-ladder}
\begin{tabular}{@{}lccc r>{\columncolor{ezHl}}r rr>{\columncolor{ezHl}[\tabcolsep][0pt]}r@{}}
\toprule
\multicolumn{4}{c}{Architectural choices} & \multicolumn{2}{c}{No rotation} & \multicolumn{3}{c}{With rotation (W4A4)} \\
\cmidrule(lr){1-4}\cmidrule(lr){5-6}\cmidrule(lr){7-9}
\dcol{Decay} & $\|\mathbf Q \mathbf K\|$ & \gcol{$\beta$} & \ecol{$\Delta$} & \numhead{W4A8} & \numhead{W4A4} & \numhead{$\bR_1$} & \numhead{$\bR_4$} & \numhead{$\bR_1\bR_4\rout$} \\
\midrule
$-$ & $-$ & $-$ & $-$ & 5.8 & 98.4 & 36.7 & 47.9 & 10.6 \\
$-$ & $\checkmark$ & $-$ & $-$ & 5.9 & 92.3 & 37.8 & 46.5 & 11.1 \\
\midrule
$-$ & $\checkmark$ & $-$ & $-$ & 6.8 & 111.4 & 48.1 & 54.0 & 12.5 \\
$-$ & $\checkmark$ & \gcol{\checkmark} & $-$ & 6.9 & 82.4 & 47.2 & 35.9 & 12.4 \\
$-$ & $\checkmark$ & \gcol{\checkmark} & \ecol{\checkmark} & 6.1 & 68.2 & 46.5 & 28.5 & 12.7 \\
\addlinespace[2pt]
\dcol{Scal} & $\checkmark$ & $-$ & $-$ & 6.0 & 56.6 & 40.3 & 24.7 & 11.9 \\
\dcol{Scal} & $\checkmark$ & \gcol{\checkmark} & $-$ & 5.5 & 53.5 & 41.3 & 21.3 & 11.3 \\
\dcol{Scal} & $\checkmark$ & \gcol{\checkmark} & \ecol{\checkmark} & 6.0 & 55.4 & 43.2 & 21.5 & 11.8 \\
\addlinespace[2pt]
\dcol{Diag} & $\checkmark$ & $-$ & $-$ & 4.8 & 45.0 & 32.6 & 18.3 & 9.3 \\
\dcol{Diag} & $\checkmark$ & \gcol{\checkmark} & $-$ & 5.4 & 49.2 & 35.7 & 20.7 & 10.6 \\
\dcol{Diag} & $\checkmark$ & \gcol{\checkmark} & \ecol{\checkmark} & 5.8 & 47.8 & 37.0 & 19.5 & 11.2 \\
\bottomrule
\end{tabular}
\end{wraptable}

\textbf{Experimental setup.}
In this section, we investigate how the choice of sequence mixer affects W/A quantization using a $1.3$B parameter mixer zoo (training recipe in Appendix~\ref{app:zoo-training}). We apply the same quantization stack to every model: GPTQ~\citep{frantar2022gptq} with per-output-channel weight scales and dynamic activation scales~\citep{liu2025spinquant} (details in Appendix~\ref{app:wa-setup}). Following rotation-based PTQ methods~\citep{ashkboos2024quarot,liu2025spinquant}, we evaluate three orthogonal rotations to reduce activation outliers. $\bR_1$ rotates the residual stream, $\bR_4$ rotates the input to the MLP down projection, and $\rout$ the input to the mixer's output projection. We report relative \WT{}~\citep{merity2016pointer} perplexity changes, with absolute perplexities in Appendix~\ref{app:ladder-full}.

\textbf{Decayed recurrences are robust.}
Without rotations, we find that decayed linear recurrences are more robust to 4-bit activation quantization than their softmax counterpart (Table~\ref{tab:external-ladder}). Holding the other recurrent degrees of freedom fixed, decay consistently improves robustness, with diagonal decay performing best. Applying the full rotation stack reduces W4A4 degradation to a similar level across the zoo, suggesting that the advantage of decayed recurrences arises primarily from the representation they learn. The individual rotations further localize this difference: $\bR_1$, which acts on the residual stream, is more effective for softmax models. In decayed recurrences, $\bR_4$ becomes relatively more important. This motivates the following residual stream analysis:

\begin{wrapfigure}[14]{r}{0.58\textwidth}
\vspace{-1.5\intextsep}
\centering
\input{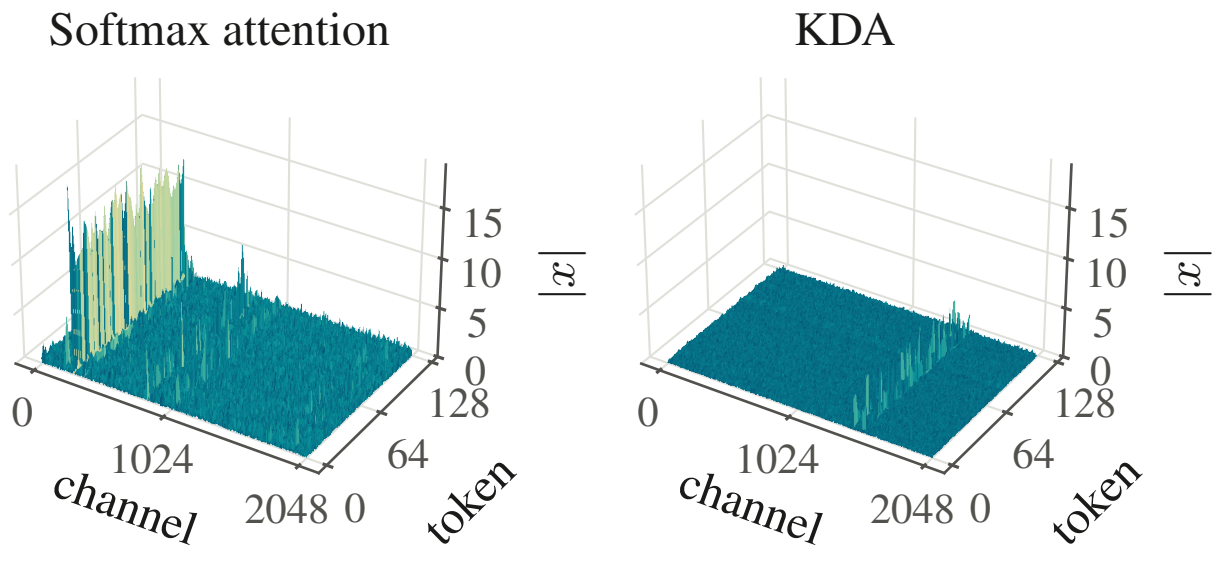}
\caption{Residual stream of a representative layer.}
\label{fig:softmax-mechanism}
\end{wrapfigure}

\textbf{Residual stream outliers.} Figure~\ref{fig:softmax-mechanism} shows that softmax attention develops a dominant residual channel whose magnitude is approximately constant across tokens (and layers, Figure~\ref{fig:skyline-layers}). Decayed recurrences exhibit weaker outliers. Across the zoo, residual stream kurtosis closely tracks unrotated W4A4 degradation (Figure~\ref{fig:external-law-curtosis}), showing that models with stronger residual outliers are harder to quantize. We next test how strongly the models rely on this channel (Table~\ref{tab:external-outlier}). Removing it before the query and key projections sharply increases perplexity in the softmax models, while removing it for the value projection is less detrimental.\looseness-1

\begin{wraptable}[14]{r}{0.44\textwidth}
\centering
\small
\setlength{\tabcolsep}{1.8pt}
\renewcommand{\arraystretch}{1.12}
\caption{Removing or replacing the dominant channel by its mean for QK, V, or the mixer input. Reported as relative \WT{} perplexity changes (\%). The first two rows use softmax attention.}
\label{tab:external-outlier}
\begin{tabular}{@{}lccc rrrr@{}}
\toprule
\multicolumn{4}{c}{Architectural choices} & \multicolumn{3}{c}{removed} & \multicolumn{1}{c}{mean} \\
\cmidrule(lr){1-4}\cmidrule(lr){5-7}\cmidrule(lr){8-8}
\dcol{Decay} & $\|\mathbf Q \mathbf K\|$ & \gcol{$\beta$} & \ecol{$\Delta$} & \numhead{QK} & \numhead{V} & \numhead{QKV} & \numhead{QKV} \\
\midrule
$-$             & $-$          & $-$               & $-$               & 1691.3 & 7.6 & 2765.2 & 7.9 \\
$-$             & $\checkmark$ & $-$               & $-$               & 419.3 & 9.8 & 389.3 & 2.7 \\
\midrule
\dcol{Diag}     & $\checkmark$ & \gcol{\checkmark} & \ecol{\checkmark} & 2.2 & 1.1 & 9.9 & 1.7 \\
\bottomrule
\end{tabular}
\end{wraptable}
Replacing the channel at the mixer input by its mean over tokens is far less harmful than removing it entirely. This approximately constant component thus provides a useful bias to the queries and keys. Decayed recurrent models are substantially less sensitive to the same intervention. Prior work connects large activation channels to attention sinks and the ability of softmax attention to concentrate probability mass on particular tokens ~\citep{sun2024massive,sun2026spike,chen2026attention}. We hypothesize that decayed recurrences have less need to develop the same residual structure because decay already provides a mechanism for focusing on a subset of tokens. We leave further investigation to future work.

\section{State Quantization}
\label{ch:state-quant}

\definecolor{sqValueInk}{HTML}{1F4E79}
\definecolor{sqKeyInk}{HTML}{216E39}
\newcommand{\sqVBest}[1]{\textcolor{sqValueInk}{\textbf{#1}}}
\newcommand{\sqKBest}[1]{\textcolor{sqKeyInk}{\textbf{#1}}}
We next study low precision storage of the recurrent state while keeping state updates in full precision. Our evaluation covers five $1.3$B mixer-zoo models and the hybrids using teacher-forced decode over eight sequences of $2048$ \WT{} tokens. Our core metric $\Delta \ell$ is the median token-level increase in cross-entropy relative to a full precision 32 bit trajectory. Appendix~\ref{app:rank-transfer-rtn} shows that $\Delta\ell$ correlates with generation accuracy, supporting its use as a useful proxy loss. We find that $\Delta \ell$ allows for efficient iteration, making it a practical metric for the study of future quantization methods. Throughout this section, we refer to zoo variants by the architectures they model.

We organize our analysis around properties that distinguish recurrent state quantization from KV cache quantization. First, the state has two channel axes, whereas a KV cache has one channel axis and one time axis. Section~\ref{sec:sq-grouping} addresses how this affects scale grouping. Second, linear recurrences typically employ a decay factor that suppresses error accumulation over time. Section~\ref{sec:sq-mixedprec} uses this structure to allocate precision according to decay strength. Third, each KV cache entry is written and quantized once, whereas the recurrent state is dequantized, updated, and requantized after every decode step. Section~\ref{sec:sq-rounding} studies how the rounding rule affects this accumulation. Finally, Section~\ref{sec:sq-rotations} reveals failure modes of random rotations under repeated state quantization, and Section~\ref{sec:sq-formats} compares integer and floating point formats. The proofs of all statements are in Appendix~\ref{app:proofs}.

\begin{figure}[t]
\centering
\input{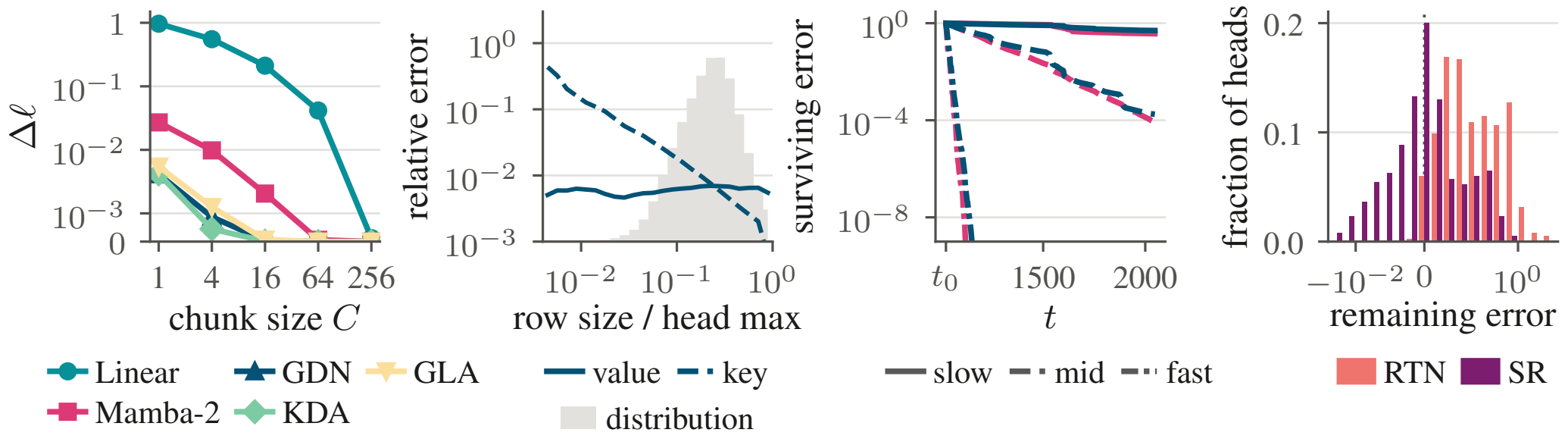}
\caption{(a)~$\Delta\ell$ under INT6 end-of-chunk state quantization with chunk size $C$. (b)~Relative readout error under INT8 value- and key-axis grouping, against the row's size relative to the largest row of its head. (c)~Surviving fraction $\lVert\bE_t\rVert_F^2/\lVert\bm{\varepsilon}_{t_0}\rVert_F^2$ of a single INT4 error injected at $t_0=1024$. Slow, mid, and fast denote heads at the 98th, 85th, and 60th percentile of mean decay. (d)~Distribution over heads of the remaining error $\lVert\bE_t\rVert_F^2-\sum_{\tau=1}^{t}\gamma_{\tau,t}^2\lVert\bm{\varepsilon}_\tau\rVert_F^2$ for Mamba-2 under INT6 value-axis grouping with RTN and SR (the latter averaging eight draws).\looseness-1}
\label{fig:sq-overview}
\vspace{-1em}
\end{figure}

For the analysis below, we keep the keys, values, and gates fixed and let $\bE_t$ denote the difference between the quantized and full-precision states. By~\Eqref{eq:zoo-recurrence}, this error consists of a propagated component and fresh quantization error $\bm{\varepsilon}_t$,
\begin{equation}
    \bE_t
    =
    \bE_{t-1}
    \dcol{\operatorname{diag}(\bm\alpha_t)}
    \ecol{(\bI - \delta\,\beta_t \bmk_t \bmk_t^\top)}
    +
    \bm{\varepsilon}_t.
\label{eq:sq-error-recurrence}
\end{equation}
Figure~\ref{fig:sq-overview}(a) shows how $\Delta \ell$ increases sharply as the state is quantized more often, demonstrating decode is the most sensitive regime.

\subsection{Scale grouping}

\label{sec:sq-grouping}
We compare two choices for scale grouping of the state $\bS_t \in \mathbb R^{d_v, d_k}$: value-axis grouping, which assigns one scale to each value row, and key-axis grouping, which assigns one scale to each key column. Both reduce degradation relative to using a single scale for the entire state (Table~\ref{tab:sq-perstate} in Appendix~\ref{app:state-quant}). Towards comparing key- and value-axis grouping, consider the readout error introduced by a single quantization step. At $b$ bits, let $r \coloneqq 2^{b-1}\!-\!1$ and let group $G$ have quantization scale \mbox{$s_G \coloneqq \lVert \bS_G\rVert_\infty/r$}. We approximate errors as independent and uniform on $[-s_G/2,s_G/2]$.
\begin{proposition}[Per-channel readout error]
\label{prop:sq-prices}
Define the readout error $\bme\coloneqq\bE\bmq$ and the row and column ranges $M_i \coloneqq \max_j |\bS_{ij}|$ and $C_j \coloneqq \max_i |\bS_{ij}|$. The expected per-channel squared readout errors for value- and key-axis grouping are, respectively,
\begin{equation}
  \mathbb{E}_{\bE}\!\left[(e_i^{\mathrm{val}})^2\right]
  = {M_i^2} / {12r^2},
  \qquad
  \mathbb{E}_{\bE}\!\left[(e_i^{\mathrm{key}})^2\right]
  = {\sum_j q_j^2 C_j^2} / {12r^2}.
  \label{eq:sq-prices}
\end{equation}
\end{proposition}
\vspace{-1em}
To interpret this result, recall that each value row of the state produces one output channel. Under value-axis grouping, the expected squared error in that output channel is proportional to the square of the row’s largest absolute entry. Under key-axis grouping, every channel has the same expected squared error, determined by the column ranges and the query. Large entries in one row can therefore set coarse quantization steps for other rows. Figure~\ref{fig:sq-overview}(b) supports this view empirically.\looseness-1

\begin{wraptable}[13]{r}{0.53\textwidth}
\vspace{-\intextsep}
\centering
\small
\setlength{\tabcolsep}{2.9pt}
\caption{Median cross-entropy increase $\Delta\ell$ for value- and key-axis grouping under RTN.}
\label{tab:sq-rot}
\begin{tabular}{lrrrrrr}\toprule
& \multicolumn{2}{c}{INT8} & \multicolumn{2}{c}{INT6} & \multicolumn{2}{c}{INT4}\\
\cmidrule(lr){2-3}\cmidrule(lr){4-5}\cmidrule(lr){6-7}
model & \numhead{value} & \numhead{key} & \numhead{value} & \numhead{key} & \numhead{value} & \numhead{key}\\\midrule
Linear & \textbf{0.125} & 0.332 & 1.230 & \textbf{0.967} & 3.647 & \textbf{2.479}\\
Mamba-2 & \textbf{0.003} & 0.011 & \textbf{0.027} & 0.052 & 0.521 & \textbf{0.400}\\
GLA & \textbf{0.000} & 0.005 & \textbf{0.005} & 0.034 & \textbf{0.269} & 0.309\\
GDN & \textbf{0.000} & 0.005 & \textbf{0.004} & 0.025 & \textbf{0.202} & 0.313\\
KDA & \textbf{0.000} & 0.007 & \textbf{0.004} & 0.033 & 0.276 & \textbf{0.266}\\
\midrule
Qwen-3.5 & \textbf{0.001} & 0.010 & \textbf{0.004} & 0.157 & \textbf{0.825} & 1.037\\
Nemotron-H & \textbf{0.001} & 0.019 & \textbf{0.010} & 0.183 & \textbf{0.763} & 1.227\\
Kimi-Linear & \textbf{0.001} & 0.001 & \textbf{0.005} & 0.009 & 0.262 & \textbf{0.056}\\
\bottomrule
\end{tabular}
\end{wraptable}

\textbf{Value grouping is robust.}
Table~\ref{tab:sq-rot} compares the two grouping axes. Value-axis grouping is preferred for every model at INT8 and for all models except Linear at INT6. At INT4, errors increase substantially, indicating that the bit budget is insufficient. This also exposes a limitation of the local approximation above, which does not consider error accumulation from repeated requantization (a phenomenon that is exacerbated in low-bit regimes). We discuss and mitigate this effect in Section~\ref{sec:sq-rounding}.

\subsection{Decay-Based Mixed Precision}
\label{sec:sq-mixedprec}

\begin{wraptable}[14]{r}{0.34\textwidth}
\vspace{-\intextsep}
\centering
\small
\setlength{\tabcolsep}{4pt}
\caption{Decay-based mixed precision. INT5 promotes the slowest-decaying half of heads (or channels for GLA, KDA and Kimi). Entries are $\Delta\ell$ under RTN.}
\label{tab:sq-mp}
\begin{tabular}{lrrr}\toprule
model & \numhead{INT4} & \numhead{INT5} & \numhead{INT6}\\\midrule
Mamba-2 & 0.400 & 0.033 & 0.027\\
GLA & 0.269 & 0.009 & 0.005\\
GDN & 0.202 & 0.005 & 0.004\\
KDA & 0.266 & 0.005 & 0.004\\
\midrule
Qwen-3.5 & 0.825 & 0.006 & 0.004\\
Nemotron-H & 0.763 & 0.045 & 0.010\\
Kimi-Linear & 0.056 & 0.006 & 0.005\\
\bottomrule
\end{tabular}
\end{wraptable}

Unlike static entries in a KV cache, a recurrent state can suppress previously introduced quantization error through decay. Unrolling~\Eqref{eq:sq-error-recurrence} yields the following result.
\begin{proposition}[Decaying error]
\label{prop:sq-decay}
Assume the decay is bounded by $\lVert\dcol{\bm\alpha_t}\rVert_\infty\leq\dcol{\alpha}<1$ for all $t\geq1$. Then
\begin{equation*}
\lVert\bE_t\rVert_F\le\sum_{\tau=1}^t \dcol{\alpha}^{t-\tau}\lVert\bm{\varepsilon}_\tau\rVert_F.
\end{equation*}
\end{proposition}
\vspace{-0.5em}
This result shows that decay limits error accumulation by exponentially suppressing the contribution of earlier state quantization steps. Figure~\ref{fig:sq-overview}(c) tracks a single injected error and shows faster suppression in faster-decaying heads.

\textbf{Decay based precision allocation.}
We use the above observation that decay attenuates injected error to assign more precision to more slowly decaying components. Using a calibration set, we rank heads (or channels for per-channel decayed models) by their average decay and promote the slowest-decaying half from the best INT4 configuration to the best INT6 configuration (Table~\ref{tab:sq-rot}). Table~\ref{tab:sq-mp} shows that the resulting INT5 effective precision recovers most of the gap from INT4 to INT6.\looseness-1

\subsection{Rounding}
\label{sec:sq-rounding}
Under repeated requantization, errors can reinforce or cancel one another. Unlike round-to-nearest (RTN), stochastic rounding (SR) is conditionally unbiased~\citep{gupta2015deep,croci2022stochastic} and can therefore inhibit error reinforcement. Indeed, \citet{chandiramani2026nemotron} find that SR helps state quantization in Mamba-2.  The following result, obtained by unrolling~\Eqref{eq:sq-error-recurrence}, formalizes this observation:

\begin{proposition}[Repeated quantization]
\label{prop:rtn-dead-zone}
For scalar decay $\alpha_t$ and $\delta=0$, define $\gamma_{\tau,t}=\prod_{j=\tau+1}^{t}\alpha_j$. Under RTN and SR, respectively, the error at time $t$ satisfies
\begin{equation*}
\begin{aligned}
\lVert\bE_t^{\mathrm{RTN}}\rVert_F^2
&= \sum_{\tau=1}^t \gamma_{\tau,t}^2
\lVert\bm{\varepsilon}_\tau^{\mathrm{RTN}}\rVert_F^2
+2\!\!\!\!\!\!\sum_{1\le\tau<s\le t}\!\!\!\!\!\!\gamma_{\tau,t}\gamma_{s,t}
\bigl\langle\bm{\varepsilon}_\tau^{\mathrm{RTN}},
\bm{\varepsilon}_s^{\mathrm{RTN}}\bigr\rangle_F,\\
\mathbb{E}\!\left[\lVert\bE_t^{\mathrm{SR}}\rVert_F^2\right]
&= \sum_{\tau=1}^t \gamma_{\tau,t}^2
\mathbb{E}\!\left[\lVert\bm{\varepsilon}_\tau^{\mathrm{SR}}\rVert_F^2\right].
\end{aligned}
\end{equation*}
\end{proposition}

\begin{wraptable}[16]{r}{0.46\textwidth}
\vspace{-\intextsep}
\centering
\small
\setlength{\tabcolsep}{2.9pt}
\caption{Median cross-entropy increase $\Delta\ell$ for RTN and SR with value-axis grouping and symmetric grids. Bold marks best per precision. Table~\ref{tab:sq-rounding-full} contains more data.}
\label{tab:sq-rounding}
\begin{tabular}{@{}lrrrr@{}}\toprule
& \multicolumn{2}{c}{INT8} & \multicolumn{2}{c}{INT6}\\
\cmidrule(lr){2-3}\cmidrule(lr){4-5}
model & \numhead{RTN} & \numhead{SR} & \numhead{RTN} & \numhead{SR}\\\midrule
Mamba-2 & 0.0026 & \textbf{0.0015} & \textbf{0.0270} & 0.0352\\
GLA & 0.0002 & \textbf{0.0001} & 0.0055 & \textbf{0.0020}\\
GDN & \textbf{0.0000} & 0.0001 & \textbf{0.0041} & 0.0048\\
KDA & \textbf{0.0002} & \textbf{0.0002} & \textbf{0.0042} & 0.0043\\
\midrule
Qwen-3.5 & 0.0005 & \textbf{0.0001} & 0.0039 & \textbf{0.0016}\\
Nemotron-H & 0.0006 & \textbf{0.0001} & 0.0105 & \textbf{0.0026}\\
Kimi-Linear & 0.0005 & \textbf{0.0000} & 0.0046 & \textbf{0.0007}\\
\bottomrule
\end{tabular}
\end{wraptable}

This result shows that rounding errors from different steps can reinforce one another under RTN. Under SR, these interactions vanish in expectation. Figure~\ref{fig:sq-overview}(d) evaluates the remainder \mbox{$\|\bE_t\|_F^2 - \sum_{\tau=1}^t \gamma_{\tau,t}^2 \|\bm{\varepsilon}_\tau\|_F^2$} empirically, isolating the interaction terms. It is predominantly positive under RTN, indicating reinforcement, but clusters around zero under SR, consistent with its vanishing expectation.

\textbf{SR shows benefits in practice.}
Table~\ref{tab:sq-rounding} compares RTN and SR under value-axis grouping. It shows that SR consistently lowers $\Delta \ell$ for all three hybrids at both INT8 and INT6. On the mixer zoo, the results are mixed, likely because SR can introduce larger fresh rounding errors than RTN. We will later see that real decode tasks also favor SR for all three hybrids (Appendix~\ref{app:end-to-end}, Table~\ref{tab:aime-cube}).

\subsection{Rotations}
\label{sec:sq-rotations}

\begin{figure*}[t]
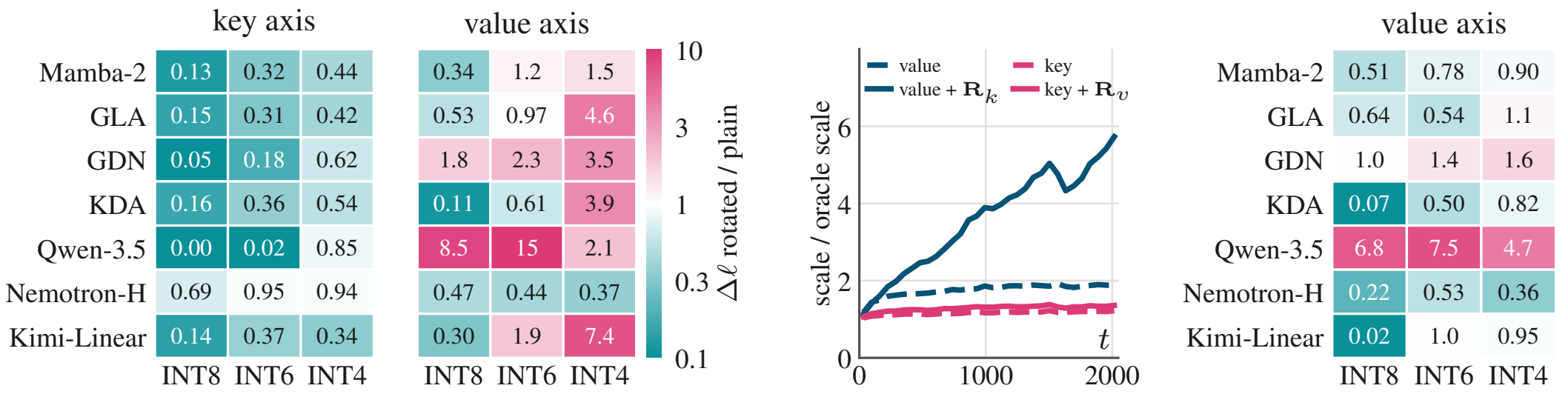

\centering
\begin{subfigure}[b]{0.47\linewidth}
\centering
\input{media/sq_rotation_effect.pgf}
\caption{$\Delta\ell$ rotated / plain}
\label{fig:sq-rotation-effect-a}
\end{subfigure}\hfill
\begin{subfigure}[b]{0.25\linewidth}
\centering
\input{media/sq_rotation_feedback_b.pgf}
\caption{Scale inflation}
\label{fig:sq-rotation-effect-b}
\end{subfigure}\hfill
\begin{subfigure}[b]{0.25\linewidth}
\centering
\input{media/sq_rotation_effect_oracle_value.pgf}
\caption{$\Delta\ell$ rotated/plain, oracle}
\label{fig:sq-rotation-effect-c}
\end{subfigure}
\caption{(a) $\Delta\ell$ with rotation divided by $\Delta\ell$ without it ($\bR_v$ for key-axis grouping, $\bR_k$ for value-axis grouping), under SR. Below one: rotation lowers the error. Data in Table~\ref{tab:sq-rot-sr-full}. (b) INT4 SR quantization scale over the scale of an auxiliary full-precision (an ``oracle'') copy of the recurrence against the decode step for GLA, mean within a head, median over 64 steps, 90th percentile over heads. (c) As (a) for value-axis grouping, with every scale taken from an oracle copy of the state (Appendix~\ref{app:sq-scale-feedback}).\looseness-1}
\label{fig:sq-rotation-effect}
%\vspace{-0.5em}
\end{figure*}

KV cache quantization methods such as QuaRot~\citep{ashkboos2024quarot} and SpinQuant~\citep{liu2025spinquant} use rotations to reduce outliers before quantization. Linear attention with scalar or no decay also admits the required symmetries~\citep{nazari2026key,tumma2026linquantfunctionalinvarianceslowprecision}. Diagonally decayed models, like KDA, require explicit rotations of the key axis before quantization. During decode, the setting of this paper, rotations furthermore influence temporal error propagation. We investigate this effect below.\looseness-1

For value-axis grouping, we denote by $\bR_k$ a within scale-group rotation of the key-axis. For key-axis grouping, $\bR_v$ rotates the value dimension. Figure~\ref{fig:sq-rotation-effect}(a) compares these within scale-group rotations under SR. While $\bR_v$ lowers $\Delta\ell$ with key-axis grouping, $\bR_k$ can increase it with value-axis grouping, particularly at low bit widths. LinQuant~\citep{tumma2026linquantfunctionalinvarianceslowprecision} observes a similar rotation penalty during chunked prefill. During decode, this penalty can be exacerbated by the recurrent state being requantized after every token, not just at the chunk boundaries. We characterize the resulting error accumulation below.\looseness-1

\begin{wrapfigure}[16]{r}{0.48\textwidth}
\vspace{-1.2\intextsep}
\centering
\input{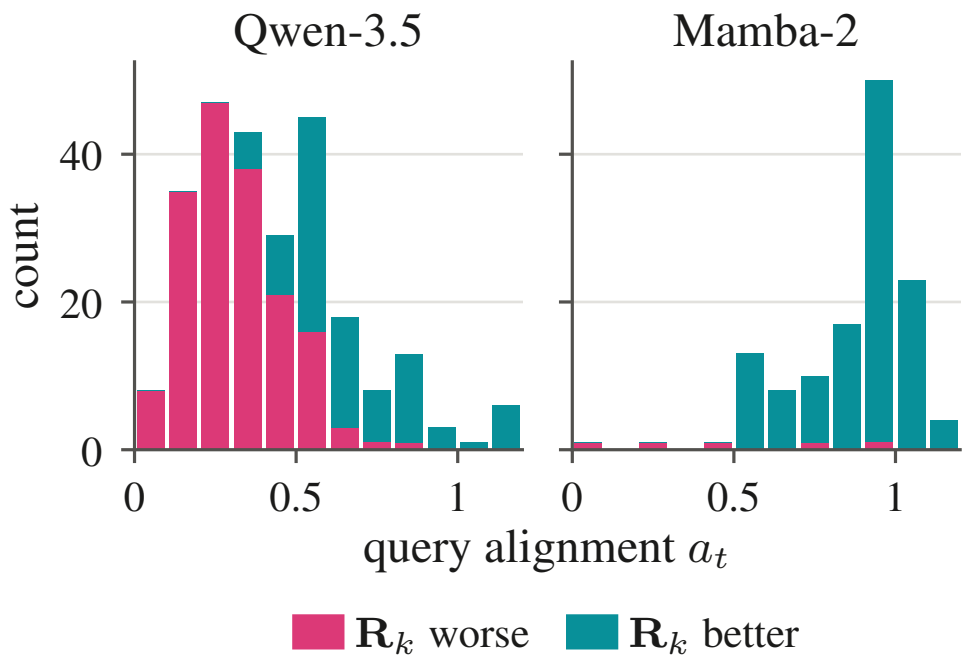}
\caption{Query alignment $a_t$ under INT4 SR with oracle scales. Heads are split by whether $\bR_k$ increases or lowers readout error.\looseness-1}
\label{fig:sq-rot-query-alignment}
\end{wrapfigure}

\paragraph{Scale feedback loops.}
The accumulated state error $\bE_t$ consists of the fresh rounding error $\bm{\varepsilon}_t$ and propagated earlier errors (~\Eqref{eq:sq-error-recurrence}). Following~\citet{chee2023quip}, we find random Hadamard rotations reduce one-step quantization error $\bm{\varepsilon}_t$ (Appendix~\ref{app:sq-rotations-formats}, Table~\ref{tab:sq-oneshot-rot}). This benefit need not persist under repeated requantization.\looseness-1

During decode, quantization scales are derived from an already noisy state. This can lead to growing scales and thus coarsen the quantization grid, which in turn can introduce larger rounding errors and increase the next scale. This feedback loop can outweigh the one-step benefit of rotations. Figure~\ref{fig:sq-rotation-effect}(b) illustrates this effect in GLA, where $\bR_k$ increases scale inflation under value-axis grouping. Figure~\ref{fig:sq-rot-inflation-time} in Appendix~\ref{app:sq-scale-feedback} extends this comparison to other recurrences. In order to isolate the influence of scale inflation on $\Delta \ell$, we carry along a full precision copy of the state. We then derive oracle scales from this full precision state (Appendix~\ref{app:sq-scale-feedback}). This intervention substantially reduces the additional $\Delta\ell$ penalty caused by $\bR_k$ (Figure~\ref{fig:sq-rotation-effect}(c)). However, such oracle scales still leave a rotation penalty in some models:

\paragraph{Query alignment.}
The effect of fresh rounding error on the readout depends on its alignment with the query. For unit-norm queries, we define
\begin{equation*}
a_t\coloneqq d_k\,\frac{\mathbb{E}\lVert\bm{\varepsilon}_t\bmq_t\rVert_2^2}
{\mathbb{E}\lVert\bm{\varepsilon}_t\rVert_F^2},
\end{equation*}
with expectations over fresh SR draws. Rotations influence this alignment. Figure~\ref{fig:sq-rot-query-alignment} showcases the effect at INT4 with oracle scales. For Qwen-3.5, which has a remaining rotation penalty even with oracle scales, unrotated error is often less aligned with the query. Rotation removes this advantage, leaking quantization error into usually suppressed query dimensions.

\subsection{Format}
\label{sec:sq-formats}

\begin{wraptable}[13]{r}{0.46\textwidth}
\vspace{-\intextsep}
\centering
\small
\setlength{\tabcolsep}{2.9pt}
\caption{Median cross-entropy increase $\Delta \ell$ for INT8 and E4M3 under SR.}
\label{tab:sq-fp-tf}
\begin{tabular}{@{}lrrrr@{}}\toprule
& \multicolumn{2}{c}{value} & \multicolumn{2}{c}{key}\\
\cmidrule(lr){2-3}\cmidrule(lr){4-5}
model & \numhead{INT8} & \numhead{E4M3} & \numhead{INT8} & \numhead{E4M3}\\\midrule
Mamba-2 & 0.0015 & 0.0019 & \textbf{0.0005} & 0.0010\\
GLA & \textbf{0.0001} & 0.0007 & 0.0005 & 0.0007\\
GDN & \textbf{0.0001} & 0.0016 & 0.0007 & 0.0013\\
KDA & \textbf{0.0002} & 0.0006 & 0.0005 & 0.0009\\
\midrule
Qwen-3.5 & \textbf{0.0001} & 0.0002 & 0.8767 & 0.0004\\
Nemotron-H & \textbf{0.0001} & 0.0015 & 6.5132 & 0.0429\\
Kimi-Linear & \textbf{0.0000} & 0.0001 & 0.0002 & 0.0001\\
\bottomrule
\end{tabular}
\end{wraptable}

The integer quantization grids considered in the above experiments are uniformly spaced. We compare them with FP8 in E4M3 format, which gives finer spacing near zero and coarser spacing at large magnitudes. Under SR with value-axis grouping, INT8 consistently yields lower $\Delta\ell$ than E4M3 (Table~\ref{tab:sq-fp-tf}). With key-axis grouping, E4M3 improves over INT8 for the three hybrids but worsens all four decayed mixer-zoo models. Across both axes and formats, value-grouped INT8 is best for almost all models, including all three hybrids.

\definecolor{e2eWAFull}{HTML}{0072B2}
\definecolor{e2eW8A16}{HTML}{009E73}
\definecolor{e2eW8A8}{HTML}{E69F00}
\definecolor{e2eW4A8}{HTML}{D55E00}

\begin{table}[t!]
\centering
\caption{End-to-end accuracy (\%). W/A/S gives weight, activation, and state precision in bits. GPQA and AIME are avg@32.}
\label{tab:end-to-end-summary}
\centering
\small
\setlength{\tabcolsep}{2.3pt}
\renewcommand{\arraystretch}{1.10}
\resizebox{\linewidth}{!}{%
\begin{tabular}{@{}lrrrrrrrrrrrrrrr@{}}
\toprule
& \multicolumn{5}{c}{Qwen-3.5}
& \multicolumn{5}{c}{Nemotron-H}
& \multicolumn{5}{c}{Kimi-Linear} \\
\cmidrule(lr){2-6}
\cmidrule(lr){7-11}
\cmidrule(lr){12-16}
W/A/S
    & \numhead{GSM8K} & \numhead{MMLU-Pro} & \numhead{GPQA} & \numhead{AIME} & \numhead{RULER}
    & \numhead{GSM8K} & \numhead{MMLU-Pro} & \numhead{GPQA} & \numhead{AIME} & \numhead{RULER}
    & \numhead{GSM8K} & \numhead{MMLU-Pro} & \numhead{GPQA} & \numhead{AIME} & \numhead{RULER} \\
\midrule
\rowcolor{e2eWAFull!12}
$16/16/32$
    & 95.9 & 82.6 & 80.1 & 82.3 & 94.6
    & 92.3 & 70.6 & 54.1 & 51.2 & 88.0
    & 89.1 & 70.3 & 63.5 & 63.6 & 93.7 \\
\rowcolor{e2eWAFull!12}
$16/16/8$
    & 96.4 & 82.2 & 79.6 & 79.5 & 94.7
    & 92.0 & 70.1 & 53.5 & 48.4 & 88.3
    & 89.2 & 70.2 & 63.8 & 62.6 & 93.8 \\
$8/16/32$
    & 96.5 & 82.5 & 80.1 & 83.1 & 94.6
    & 92.9 & 70.4 & 54.0 & 51.7 & 88.4
    & 89.8 & 70.4 & 63.9 & 64.3 & 93.5 \\
$8/16/8$
    & 96.0 & 82.3 & 79.1 & 80.1 & 94.5
    & 92.5 & 70.0 & 53.1 & 49.4 & 88.2
    & 89.6 & 70.2 & 61.9 & 62.4 & 94.0 \\
\rowcolor{e2eWAFull!12}
$8/8/32$
    & 96.8 & 82.4 & 80.3 & 82.1 & 94.6
    & 93.3 & 70.3 & 54.8 & 50.9 & 88.4
    & 88.9 & 70.1 & 63.3 & 62.3 & 93.7 \\
\rowcolor{e2eWAFull!12}
$8/8/8$
    & 96.5 & 82.2 & 79.8 & 78.6 & 94.7
    & 92.6 & 70.1 & 53.5 & 49.5 & 88.5
    & 88.4 & 69.7 & 62.6 & 63.6 & 93.9 \\
$4/8/32$
    & 97.1 & 80.7 & 75.0 & 73.5 & 94.9
    & 90.9 & 67.8 & 49.0 & 43.3 & 85.4
    & 86.4 & 65.9 & 53.6 & 56.9 & 93.4 \\
$4/8/8$
    & 96.1 & 80.1 & 73.6 & 70.9 & 94.8
    & 91.8 & 67.0 & 49.7 & 41.0 & 85.1
    & 88.9 & 65.5 & 52.6 & 55.2 & 93.3 \\
\bottomrule
\end{tabular}}
\end{table}

\section{End-to-End Evaluation}
\label{sec:recipes}

To demonstrate the practical utility of our findings, we combine extra-state W/A quantization (Section~\ref{sec:external}) with \methodname{}, our intra-state quantization (Section~\ref{ch:state-quant}) recipe. \methodname{} stores recurrent states on unrotated integer grids with value-axis grouping, symmetric absmax scales, and stochastic rounding. We use INT8 as the default state precision. For lower effective precisions, we use decay-based precision allocation to assign fewer bits to quickly decaying heads or channels. We evaluate the framework on the three large post-trained hybrid models. Appendix~\ref{app:gen-protocol} gives the complete protocol.\looseness-1

\begin{wrapfigure}[16]{R}{0.5\textwidth}
\vspace{-1.5\intextsep}
\centering
\input{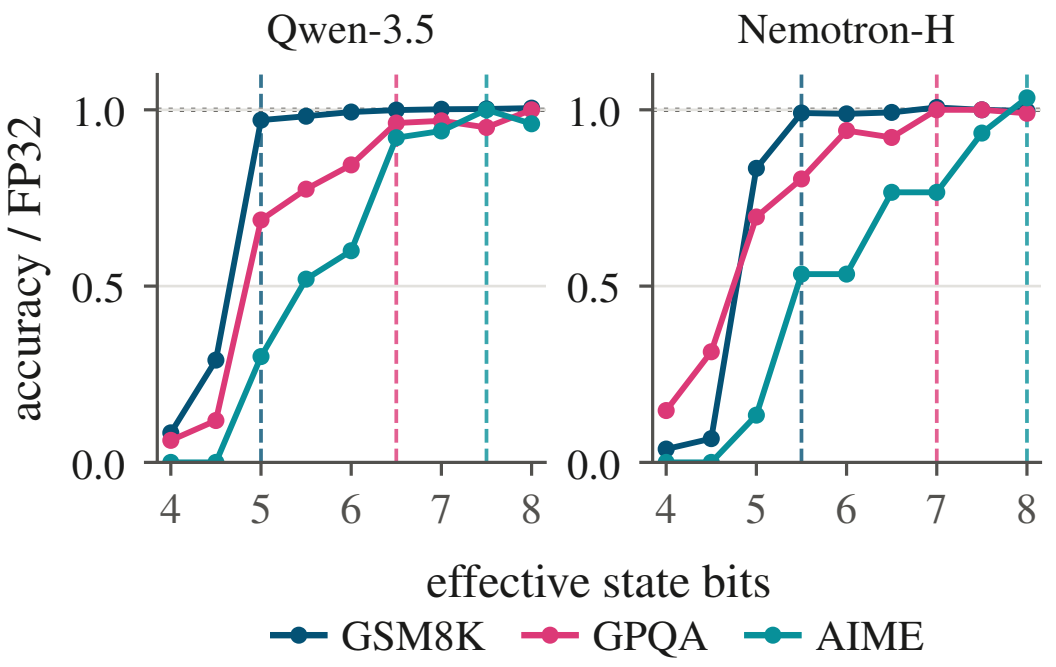}
\caption{Decay-ranked promotion. Dotted lines mark the minimum bit budget within 5\% of FP32 accuracy. GPQA and AIME are pass@1.}
\label{fig:promo-relative}
\end{wrapfigure}

\textbf{Accuracy.}
We evaluate 8-shot GSM8K~\citep{cobbe2021training}, MMLU-Pro~\citep{wang2024mmlu}, GPQA~\citep{rein2023gpqa}, and AIME (combining the 2024 and 2025 competitions for 60 questions), together with RULER~\citep{hsieh2024ruler} at a $32$k context. Generation is capped at $8$k tokens for GSM8K and $64$k for MMLU-Pro, GPQA, and AIME.

Weight quantization uses GPTQ~\citep{frantar2022gptq} with groups of $128$ input channels, except that W8A8 uses per-output-channel weight scales (Appendix~\ref{app:wa-setup}). We apply W/A quantization with llm-compressor~\citep{llmcompressor2024} and serve the resulting models with vLLM~\citep{kwon2023efficient}.

Table~\ref{tab:end-to-end-summary} compares INT8 and FP32 states across W/A precisions. INT8 states preserve most of the accuracy at each W/A setting, while larger losses generally arise in configurations with 4-bit weights. The additional loss from state quantization does not increase consistently as W/A precision decreases, supporting the combination of extra-state and intra-state quantization.

Next, we examine decode length, which helps interpret the variation across benchmarks. Median response lengths follow AIME $>$ GPQA $>$ MMLU-Pro $>$ GSM8K $>$ RULER (Table~\ref{tab:genlen}). GSM8K and MMLU-Pro generally show small changes under state quantization. RULER likewise remains stable when state precision is reduced, despite its $32$k-token input, because generation is short. GPQA and AIME tend to show somewhat larger losses, consistent with greater sensitivity over longer decode trajectories. We believe that incorporating quantization during training (e.g. via quantization-aware training or low-precision training) could reduce the remaining degradation, but leave further exploration of this to future work.

As discussed in Section~\ref{ch:state-quant}, we use decay-based allocation to reduce state precision below INT8, promoting linear attention heads of Qwen-3.5 and Nemotron-H between adjacent precisions based on decay strength. Figure~\ref{fig:promo-relative} shows that both models support lower effective precision on GSM8K and GPQA. On GSM8K, this permits $5$ effective bits for Qwen-3.5 and $5.5$ for Nemotron-H with less than a $5\%$ relative drop in accuracy from the FP32-state baseline.

\begin{figure}[t!]
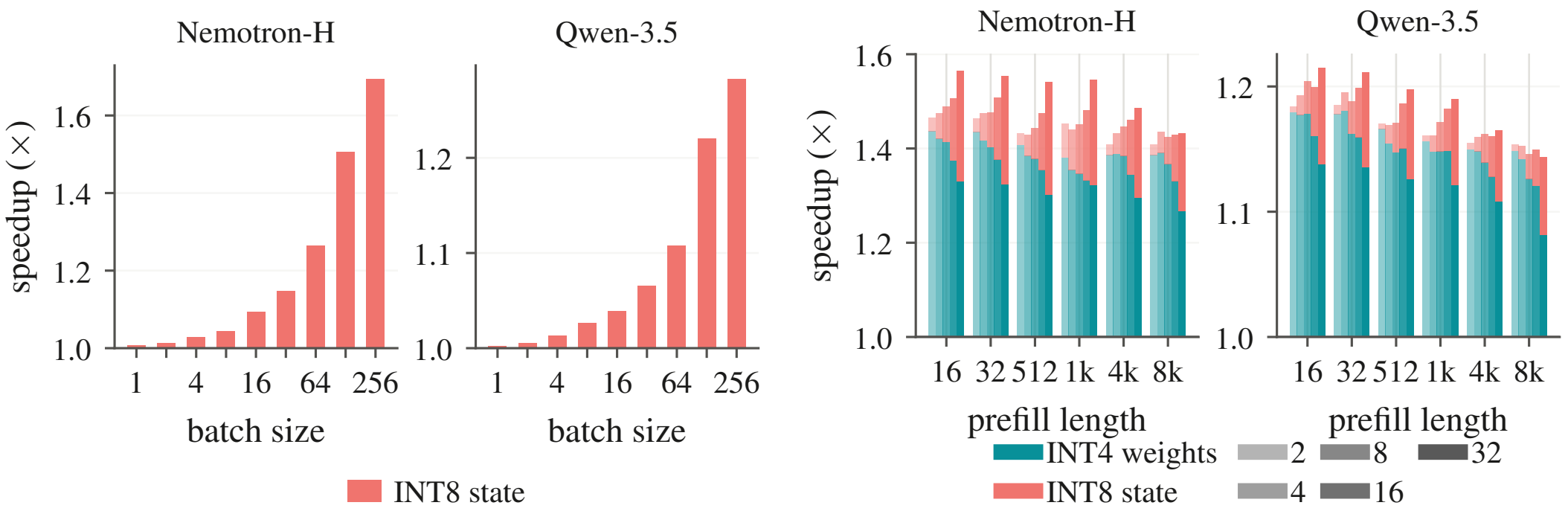

\centering
\begin{subfigure}[b]{0.49\textwidth}
\centering
\resizebox{\linewidth}{!}{\input{media/speed_vllm_state_only.pgf}}
\caption{BF16 Weights.}
\label{fig:speed-vllm-w4}
\end{subfigure}\hfill
\begin{subfigure}[b]{0.49\textwidth}
\centering
\resizebox{\linewidth}{!}{\input{media/speed_vllm_b3_stack_wrap.pgf}}
\caption{INT4 Weights}
\label{fig:speed-vllm-dilution}
\end{subfigure}
\caption{Time-per-output-token (TPOT) speedup. Subfigure~(\subref{fig:speed-vllm-w4}) shows the speedup of an INT8 state over an FP32 state with BF16 weights across batch sizes. Subfigure~(\subref{fig:speed-vllm-dilution}) shows the speedup relative to BF16 weights with an FP32 state across prefill lengths at several batch sizes. The lower segment shows the contribution of INT4 weights; the upper segment shows the additional contribution of INT8 states. Appendix~\ref{app:speed-raw} reports contributions separately as well as absolute TPOT.}
\label{fig:speed-vllm-joint}
\vspace{-1em}
\end{figure}

\begin{wrapfigure}[15]{r}{0.34\textwidth}
\vspace{-\intextsep}
\centering
\input{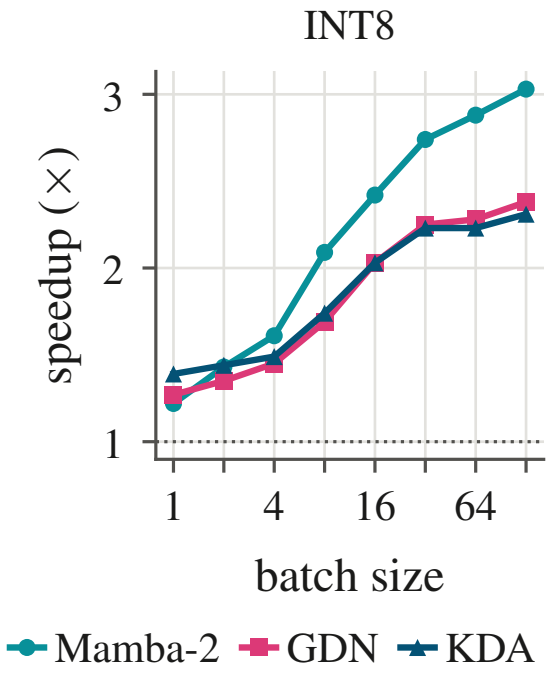}
\caption{H100 speedup of vLLM mixer kernels using INT8 versus FP32 states.}
\label{fig:speed-kernel}
\end{wrapfigure}

\textbf{Speedup.} We evaluate how state quantization affects decode speed in individual sequence mixer kernels and in end-to-end vLLM serving. Figure~\ref{fig:speed-kernel} shows that INT8 state storage accelerates all three sequence mixer kernels as the decode batch size increases. At a batch size of 8, Mamba-2 more than doubles throughput, while GDN and KDA also improve significantly. Appendix~\ref{app:kernel-design} describes the measurement and reports the same experiment on an RTX~6000.

To evaluate the end-to-end efficiency of \methodname, we measure fixed-batch decoding of Nemotron-H and Qwen-3.5 with vLLM on one H100. We report time per output token (TPOT), excluding prefill. Appendix~\ref{app:speed-protocol} gives the complete measurement protocol. Appendix~\ref{app:throughput} reports additional results. Figure~\ref{fig:speed-vllm-w4} shows that the speedup from INT8 state quantization grows with batch size, since each sequence carries its own recurrent state. Weight quantization contributes most at small batch sizes because weight traffic is shared across active sequences. Combining weight and state quantization therefore provides complementary speedups across serving loads (Figure~\ref{fig:speed-vllm-dilution}). Finally, since Qwen and Nemotron are hybrid models, longer prefill grows the KV cache and reduces the relative contribution of the recurrent state to total memory traffic.\looseness-1

\section{Conclusion}
In this work, we study PTQ of linear recurrent and hybrid language models, separating weight/activation quantization from state quantization during decode. First, we focus on extra-state quantization, and in controlled experiments find that decayed recurrences develop weaker residual outliers than softmax attention, making their activations easier to quantize. Next, we characterize the effects of scale grouping, decay, and repeated rounding on state error. We furthermore reveal failure modes of random rotations under repeated state-quantization and compare integer and floating point formats for state quantization. We showcase the practical utility of our analysis by combining extra-state W/A quantization with \methodname{}, an intra-state quantization scheme which combines value-axis grouping, stochastic rounding, and decay-based precision allocation. Empirically, \methodname{} supports low bit states on Qwen-3.5, Nemotron-H, and Kimi-Linear with modest accuracy loss, with lower precision at shorter decode horizons. We implement \methodname{} in vLLM and show that combining state and weight quantization improves decode throughput. Future work could explore quantization-aware or low-precision training as well as GPU-native microscaling formats to further improve the performance and efficiency of linear recurrences. Furthermore, the effect of rotations on state quantization warrants further investigation.

\clearpage
\newpage

\subsection*{Acknowledgments}
This work was supported by the Hector Foundation, Max Planck ETH Center for Learning Systems, and the Department of the Air Force Artificial Intelligence Accelerator and was accomplished under Cooperative Agreement Number FA8750-19-2-1000. The views and conclusions contained in this document are those of the authors and should not be interpreted as representing the official policies, either expressed or implied, of the Department of the Air Force or the U.S. Government. The U.S. Government is authorized to reproduce and distribute reprints for Government purposes notwithstanding any copyright notation herein.

\bibliography{iclr2027_conference}

@article{yang2024parallelizing,
  title={Parallelizing linear transformers with the delta rule over sequence length},
  author={Yang, Songlin and Wang, Bailin and Zhang, Yu and Shen, Yikang and Kim, Yoon},
  journal={Advances in neural information processing systems},
  volume={37},
  pages={115491--115522},
  year={2024}
}

@inproceedings{tianqi2025qmamba,
  title={Q-Mamba: Towards more efficient Mamba models via post-training quantization},
  author={Tianqi, Chen and Chen, Yuanteng and Wang, Peisong and Xu, Weixiang and Zhu, Zeyu and Cheng, Jian},
  booktitle={Findings of the Association for Computational Linguistics: ACL 2025},
  pages={10594--10610},
  year={2025}
}

@article{pdn,
  title={Preconditioned DeltaNet: Curvature-aware Sequence Modeling for Linear Recurrences},
  author={Tumma, Neehal and Loo, Noel and Rus, Daniela},
  journal={arXiv preprint arXiv:2604.21100},
  year={2026}
}

@article{parnichkun2025quantifying,
  title={Quantifying memory utilization with effective state-size},
  author={Parnichkun, Rom N and Tumma, Neehal and Thomas, Armin W and Moro, Alessandro and An, Qi and Suzuki, Taiji and Yamashita, Atsushi and Poli, Michael and Massaroli, Stefano},
  journal={arXiv preprint arXiv:2504.19561},
  year={2025}
}

@inproceedings{chahine2026curious,
  title={The curious case of in-training compression of state space models},
  author={Chahine, Makram and Nazari, Philipp and Rus, Daniela and Rusch, T Konstantin},
  booktitle={International Conference on Learning Representations},
  volume={2026},
  pages={111759--111798},
  year={2026}
}

@article{vaswani2017attention,
  title={Attention is all you need},
  author={Vaswani, Ashish and Shazeer, Noam and Parmar, Niki and Uszkoreit, Jakob and Jones, Llion and Gomez, Aidan N and Kaiser, {\L}ukasz and Polosukhin, Illia},
  journal={Advances in neural information processing systems},
  volume={30},
  year={2017}
}

@InProceedings{schlag2021linear,
  title = 	 {Linear Transformers Are Secretly Fast Weight Programmers},
  author =       {Schlag, Imanol and Irie, Kazuki and Schmidhuber, J{\"u}rgen},
  booktitle = 	 {Proceedings of the 38th International Conference on Machine Learning},
  pages = 	 {9355--9366},
  year = 	 {2021},
  editor = 	 {Meila, Marina and Zhang, Tong},
  volume = 	 {139},
  series = 	 {Proceedings of Machine Learning Research},
  month = 	 {18--24 Jul},
  publisher =    {PMLR},
  url = 	 {https://proceedings.mlr.press/v139/schlag21a.html}
}

@inproceedings{yang2025gated,
  title={Gated delta networks: Improving mamba2 with delta rule},
  author={Yang, Songlin and Kautz, Jan and Hatamizadeh, Ali},
  booktitle={International Conference on Learning Representations},
  volume={2025},
  pages={29687--29707},
  year={2025}
}

@article{yang2023gated,
  title={Gated linear attention transformers with hardware-efficient training},
  author={Yang, Songlin and Wang, Bailin and Shen, Yikang and Panda, Rameswar and Kim, Yoon},
  journal={arXiv preprint arXiv:2312.06635},
  year={2023}
}

@misc{gu2024mamba,
      title={Mamba: Linear-Time Sequence Modeling with Selective State Spaces}, 
      author={Albert Gu and Tri Dao},
      year={2024},
      eprint={2312.00752},
      archivePrefix={arXiv},
      primaryClass={cs.LG},
      url={https://arxiv.org/abs/2312.00752}, 
}

@article{dao2024transformers,
  title={Transformers are SSMs: Generalized models and efficient algorithms through structured state space duality},
  author={Dao, Tri and Gu, Albert},
  journal={arXiv preprint arXiv:2405.21060},
  year={2024}
}

@InProceedings{katharopoulos2020transformers,
  title = 	 {Transformers are {RNN}s: Fast Autoregressive Transformers with Linear Attention},
  author =       {Katharopoulos, Angelos and Vyas, Apoorv and Pappas, Nikolaos and Fleuret, Fran{\c{c}}ois},
  booktitle = 	 {Proceedings of the 37th International Conference on Machine Learning},
  pages = 	 {5156--5165},
  year = 	 {2020},
  editor = 	 {III, Hal Daumé and Singh, Aarti},
  volume = 	 {119},
  series = 	 {Proceedings of Machine Learning Research},
  month = 	 {13--18 Jul},
  publisher =    {PMLR},
  url = 	 {https://proceedings.mlr.press/v119/katharopoulos20a.html}
}

@article{fu2022hungry,
  title={Hungry hungry hippos: Towards language modeling with state space models},
  author={Fu, Daniel Y and Dao, Tri and Saab, Khaled K and Thomas, Armin W and Rudra, Atri and R{\'e}, Christopher},
  journal={arXiv preprint arXiv:2212.14052},
  year={2022}
}

@article{so2021searching,
  title={Searching for efficient transformers for language modeling},
  author={So, David and Ma{\'n}ke, Wojciech and Liu, Hanxiao and Dai, Zihang and Shazeer, Noam and Le, Quoc V},
  journal={Advances in neural information processing systems},
  volume={34},
  pages={6010--6022},
  year={2021}
}

@article{ashkboos2024quarot,
  title={Quarot: Outlier-free 4-bit inference in rotated llms},
  author={Ashkboos, Saleh and Mohtashami, Amirkeivan and Croci, Maximilian L and Li, Bo and Cameron, Pashmina and Jaggi, Martin and Alistarh, Dan and Hoefler, Torsten and Hensman, James},
  journal={Advances in Neural Information Processing Systems},
  volume={37},
  pages={100213--100240},
  year={2024}
}

@inproceedings{liu2025spinquant,
  title={Spinquant: Llm quantization with learned rotations},
  author={Liu, Zechun and Zhao, Changsheng and Fedorov, Igor and Soran, Bilge and Choudhary, Dhruv and Krishnamoorthi, Raghuraman and Chandra, Vikas and Tian, Yuandong and Blankevoort, Tijmen},
  booktitle={International Conference on Learning Representations},
  volume={2025},
  pages={92009--92032},
  year={2025}
}

@inproceedings{chollet2017xception,
  title={Xception: Deep learning with depthwise separable convolutions},
  author={Chollet, Fran{\c{c}}ois},
  booktitle={2017 IEEE conference on computer vision and pattern recognition (CVPR)},
  pages={1800--1807},
  year={2017},
  organization={Ieee}
}

@inproceedings{chiang2025quamba,
  title={Quamba: A post-training quantization recipe for selective state space models},
  author={Chiang, Hung-Yueh and Chang, Chi-Chih and Frumkin, Natalia and Wu, Kai-Chiang and Marculescu, Diana},
  booktitle={International Conference on Learning Representations},
  volume={2025},
  pages={101328--101354},
  year={2025}
}

@misc{tumma2023leveraginglowranksparserecurrent,
      title={Leveraging Low-Rank and Sparse Recurrent Connectivity for Robust Closed-Loop Control}, 
      author={Neehal Tumma and Mathias Lechner and Noel Loo and Ramin Hasani and Daniela Rus},
      year={2023},
      eprint={2310.03915},
      archivePrefix={arXiv},
      primaryClass={cs.LG},
      url={https://arxiv.org/abs/2310.03915}, 
}

@misc{amini2025lfm2technicalreport,
      title={LFM2 Technical Report}, 
      author={Alexander Amini and Anna Banaszak and Harold Benoit and Arthur Böök and Tarek Dakhran and Song Duong and Alfred Eng and Fernando Fernandes and Marc Härkönen and Anne Harrington and Ramin Hasani and Saniya Karwa and Yuri Khrustalev and Maxime Labonne and Mathias Lechner and Valentine Lechner and Simon Lee and Zetian Li and Noel Loo and Jacob Marks and Edoardo Mosca and Samuel J. Paech and Paul Pak and Rom N. Parnichkun and Alex Quach and Ryan Rogers and Daniela Rus and Nayan Saxena and Bettina Schlager and Tim Seyde and Jimmy T. H. Smith and Aditya Tadimeti and Neehal Tumma},
      year={2025},
      eprint={2511.23404},
      archivePrefix={arXiv},
      primaryClass={cs.LG},
      url={https://arxiv.org/abs/2511.23404}, 
}

@article{chiang2025quamba2,
  title={Quamba2: A robust and scalable post-training quantization framework for selective state space models},
  author={Chiang, Hung-Yueh and Chang, Chi-Chih and Frumkin, Natalia and Wu, Kai-Chiang and Abdelfattah, Mohamed S and Marculescu, Diana},
  journal={arXiv preprint arXiv:2503.22879},
  year={2025}
}

@inproceedings{
yue2025mambaquant,
title={MambaQuant: Quantizing the Mamba Family with Variance Aligned Rotation Methods},
author={Zukang Xu and Yuxuan Yue and Xing Hu and Dawei Yang and Zhihang Yuan and Zixu Jiang and Zhixuan Chen and JiangyongYu and XUCHEN and Sifan Zhou},
booktitle={The Thirteenth International Conference on Learning Representations},
year={2025},
url={https://openreview.net/forum?id=KI45uDnmzv}
}

@inproceedings{kim2026ssdi8,
  title={SSDi8: Accurate and Efficient 8-bit Quantization for State Space Duality},
  author={Kim, Hyunwoo and Ko, Byoungchan and Kang, Minseok and Kim, Minwoo and Lee, Dongjin and Lee, Jaehoon and Yoon, Sungroh and Jung, Dahuin},
  booktitle={International Conference on Learning Representations},
  volume={2026},
  pages={97014--97042},
  year={2026}
}

@misc{team2026qwen3,
    title  = {{Qwen3.5}: Towards Native Multimodal Agents},
    author = {{Qwen Team}},
    month  = {February},
    year   = {2026},
    url    = {https://qwen.ai/blog?id=qwen3.5}
}

@article{hsieh2024ruler,
  title={RULER: What's the real context size of your long-context language models?},
  author={Hsieh, Cheng-Ping and Sun, Simeng and Kriman, Samuel and Acharya, Shantanu and Rekesh, Dima and Jia, Fei and Zhang, Yang and Ginsburg, Boris},
  journal={arXiv preprint arXiv:2404.06654},
  year={2024}
}

@article{sun2024massive,
  title={Massive activations in large language models},
  author={Sun, Mingjie and Chen, Xinlei and Kolter, J Zico and Liu, Zhuang},
  journal={arXiv preprint arXiv:2402.17762},
  year={2024}
}

@article{sun2026spike,
  title={The spike, the sparse and the sink: Anatomy of massive activations and attention sinks},
  author={Sun, Shangwen and Canziani, Alfredo and LeCun, Yann and Zhu, Jiachen},
  journal={arXiv preprint arXiv:2603.05498},
  year={2026}
}

@article{chen2026attention,
  title={Attention Sinks Induce Gradient Sinks: Massive Activations as Gradient Regulators in Transformers},
  author={Chen, Yihong and Lin, Zhouchen and Yao, Quanming},
  journal={arXiv preprint arXiv:2603.17771},
  year={2026}
}

@inproceedings{xiao2023smoothquant,
  title={Smoothquant: Accurate and efficient post-training quantization for large language models},
  author={Xiao, Guangxuan and Lin, Ji and Seznec, Mickael and Wu, Hao and Demouth, Julien and Han, Song},
  booktitle={International conference on machine learning},
  pages={38087--38099},
  year={2023},
  organization={PMLR}
}

@misc{nazari2026key,
      title={On State Reduction in Linear Attention}, 
      author={Philipp Nazari and T. Konstantin Rusch},
      year={2026},
      eprint={2602.04852},
      archivePrefix={arXiv},
      primaryClass={cs.LG},
      url={https://arxiv.org/abs/2602.04852}, 
}

@article{chandiramani2026nemotron,
  title={Nemotron 3 super: Open, efficient mixture-of-experts hybrid mamba-transformer model for agentic reasoning},
  author={Chandiramani, Aakshita and Blakeman, Aaron and Olaoye, Abdullahi and Gupta, Abhibha and Somasamudramath, Abhilash and Khattar, Abhinav and Adesoba, Adeola and Renduchintala, Adi and Asif, Adil and Agrawal, Aditya and others},
  journal={arXiv preprint arXiv:2604.12374},
  year={2026}
}

@article{cobbe2021training,
  title={Training verifiers to solve math word problems},
  author={Cobbe, Karl and Kosaraju, Vineet and Bavarian, Mohammad and Chen, Mark and Jun, Heewoo and Kaiser, Lukasz and Plappert, Matthias and Tworek, Jerry and Hilton, Jacob and Nakano, Reiichiro and others},
  journal={arXiv preprint arXiv:2110.14168},
  year={2021}
}

@article{wang2024mmlu,
  title={Mmlu-pro: A more robust and challenging multi-task language understanding benchmark},
  author={Wang, Yubo and Ma, Xueguang and Zhang, Ge and Ni, Yuansheng and Chandra, Abhranil and Guo, Shiguang and Ren, Weiming and Arulraj, Aaran and He, Xuan and Jiang, Ziyan and others},
  journal={Advances in Neural Information Processing Systems},
  volume={37},
  pages={95266--95290},
  year={2024}
}

@article{rein2023gpqa,
  title={Gpqa: A graduate-level google-proof q\&a benchmark},
  author={Rein, David and Hou, Betty Li and Stickland, Asa Cooper and Petty, Jackson and Pang, Richard Yuanzhe and Dirani, Julien and Michael, Julian and Bowman, Samuel R},
  journal={arXiv preprint arXiv:2311.12022},
  year={2023}
}

@article{merity2016pointer,
  title={Pointer sentinel mixture models},
  author={Merity, Stephen and Xiong, Caiming and Bradbury, James and Socher, Richard},
  journal={arXiv preprint arXiv:1609.07843},
  year={2016}
}

@article{frantar2022gptq,
  title={Gptq: Accurate post-training quantization for generative pre-trained transformers},
  author={Frantar, Elias and Ashkboos, Saleh and Hoefler, Torsten and Alistarh, Dan},
  journal={arXiv preprint arXiv:2210.17323},
  year={2022}
}

@software{llmcompressor2024,
    title={{LLM Compressor}},
    author={Red Hat AI and vLLM Project},
    year={2024},
    month={8},
    url={https://github.com/vllm-project/llm-compressor},
}

@article{blakeman2025nemotron,
  title={Nemotron-h: A family of accurate and efficient hybrid mamba-transformer models},
  author={Blakeman, Aaron and Basant, Aarti and Khattar, Abhinav and Renduchintala, Adithya and Bercovich, Akhiad and Ficek, Aleksander and Bjorlin, Alexis and Taghibakhshi, Ali and Deshmukh, Amala Sanjay and Mahabaleshwarkar, Ameya Sunil and others},
  journal={arXiv preprint arXiv:2504.03624},
  year={2025}
}

@article{team2025kimi,
  title={Kimi linear: An expressive, efficient attention architecture},
  author={Team, Kimi and Zhang, Yu and Lin, Zongyu and Yao, Xingcheng and Hu, Jiaxi and Meng, Fanqing and Liu, Chengyin and Men, Xin and Yang, Songlin and Li, Zhiyuan and others},
  journal={arXiv preprint arXiv:2510.26692},
  year={2025}
}

@inproceedings{kwon2023efficient,
  title={Efficient Memory Management for Large Language Model Serving with PagedAttention},
  author={Woosuk Kwon and Zhuohan Li and Siyuan Zhuang and Ying Sheng and Lianmin Zheng and Cody Hao Yu and Joseph E. Gonzalez and Hao Zhang and Ion Stoica},
  booktitle={Proceedings of the ACM SIGOPS 29th Symposium on Operating Systems Principles},
  year={2023}
}

@article{su2024roformer,
  title={Roformer: Enhanced transformer with rotary position embedding},
  author={Su, Jianlin and Ahmed, Murtadha and Lu, Yu and Pan, Shengfeng and Bo, Wen and Liu, Yunfeng},
  journal={Neurocomputing},
  volume={568},
  pages={127063},
  year={2024},
  publisher={Elsevier}
}

@article{zhang2019root,
  title={Root mean square layer normalization},
  author={Zhang, Biao and Sennrich, Rico},
  journal={Advances in neural information processing systems},
  volume={32},
  year={2019}
}

@article{liu2024kivi,
  title={Kivi: A tuning-free asymmetric 2bit quantization for kv cache},
  author={Liu, Zirui and Yuan, Jiayi and Jin, Hongye and Zhong, Shaochen and Xu, Zhaozhuo and Braverman, Vladimir and Chen, Beidi and Hu, Xia},
  journal={arXiv preprint arXiv:2402.02750},
  year={2024}
}

@article{hooper2024kvquant,
  title={Kvquant: Towards 10 million context length llm inference with kv cache quantization},
  author={Hooper, Coleman and Kim, Sehoon and Mohammadzadeh, Hiva and Mahoney, Michael W and Shao, Yakun S and Keutzer, Kurt and Gholami, Amir},
  journal={Advances in Neural Information Processing Systems},
  volume={37},
  pages={1270--1303},
  year={2024}
}

@article{duanmu2024skvq,
  title={Skvq: Sliding-window key and value cache quantization for large language models},
  author={Duanmu, Haojie and Yuan, Zhihang and Li, Xiuhong and Duan, Jiangfei and Zhang, Xingcheng and Lin, Dahua},
  journal={arXiv preprint arXiv:2405.06219},
  year={2024}
}

@inproceedings{gupta2015deep,
  title={Deep learning with limited numerical precision},
  author={Gupta, Suyog and Agrawal, Ankur and Gopalakrishnan, Kailash and Narayanan, Pritish},
  booktitle={International conference on machine learning},
  pages={1737--1746},
  year={2015},
  organization={PMLR}
}

@article{croci2022stochastic,
  title={Stochastic rounding: implementation, error analysis and applications},
  author={Croci, Matteo and Fasi, Massimiliano and Higham, Nicholas J and Mary, Theo and Mikaitis, Mantas},
  journal={Royal Society Open Science},
  volume={9},
  number={3},
  pages={211631},
  year={2022}
}

@article{chee2023quip,
  title={Quip: 2-bit quantization of large language models with guarantees},
  author={Chee, Jerry and Cai, Yaohui and Kuleshov, Volodymyr and De Sa, Christopher M},
  journal={Advances in neural information processing systems},
  volume={36},
  pages={4396--4429},
  year={2023}
}

@article{tseng2024quip,
  title={Quip\#: Even better llm quantization with hadamard incoherence and lattice codebooks},
  author={Tseng, Albert and Chee, Jerry and Sun, Qingyao and Kuleshov, Volodymyr and De Sa, Christopher},
  journal={Proceedings of machine learning research},
  volume={235},
  pages={48630},
  year={2024}
}

@article{yao2026stepquant,
  title={STEPQuant: When and Where Errors Matter in Delta-Rule Recurrent State Quantization},
  author={Yao, Bingchen and Xu, Haobo and Lin, Haokun and Wu, Yichen and Guo, Ziyu and Zhang, Renrui and Lu, Zhichao and Sun, Zhenan and Wei, Ying},
  journal={arXiv preprint arXiv:2609.38169},
  year={2026}
}

@article{pan2026leapquant,
  title={LeapQuant: Efficient Linear Attention with Accurate Recurrent State Quantization},
  author={Pan, Yi and Xi, Haocheng and Zhu, Kan and Li, Xingyang and Wu, Yibo and Mishra, Mayank and Zhang, Hongtao and Zheng, William X and Kasikci, Baris and Han, Song and others},
  journal={arXiv preprint arXiv:2609.38166},
  year={2026}
}

@article{zhang2026damp,
  title={Damp: Decay-aware mixed-precision recurrent-state quantization},
  author={Zhang, Tao and Tan, Jianchao and Sun, Pingwei and Yu, Yanqi and Su, Zunhai and Jiang, Zixu and Xie, Yuchen and Cai, Xunliang and Zeng, Ziqian},
  journal={arXiv preprint arXiv:2608.27513},
  year={2026}
}

@misc{loo2026paratttparallelizingtesttimetraining,
      title={ParaTTT: Parallelizing Test-Time Training Layers with Deep Memories},
      author={Noel Loo and Neehal Tumma and T. Konstantin Rusch and Mathias Lechner and Alexander Amini and Daniela Rus},
      year={2026},
      eprint={2610.38255},
      archivePrefix={arXiv},
      primaryClass={cs.LG},
      url={https://arxiv.org/abs/2610.38255},
}

@misc{tumma2026linquantfunctionalinvarianceslowprecision,
      title={LinQuant: Functional Invariances for Low-Precision Linear Recurrences}, 
      author={Neehal Tumma and Philipp Nazari and Rohan Cherukuri and Daniela Rus and T. Konstantin Rusch},
      year={2026},
      eprint={2610.39418},
      archivePrefix={arXiv},
      primaryClass={cs.LG},
      url={https://arxiv.org/abs/2610.39418}, 
}
\bibliographystyle{iclr2027_conference}
\newpage

\appendix
\section{Extended Related Work}
\label{sec:related-work}

\textbf{Linear recurrent and hybrid language models.}
A growing line of work replaces softmax attention with recurrent sequence mixers whose state size and per-token decode cost do not grow with context length. Linear attention uses associativity to express attention as a recurrent update over a fixed-size matrix state~\citep{katharopoulos2020transformers,schlag2021linear}. Selective state-space models such as Mamba and Mamba-2 introduce input-dependent recurrent dynamics and hardware-efficient implementations~\citep{gu2024mamba,dao2024transformers}.

More recent linear recurrences increase the expressivity of the state update. Gated Linear Attention introduces data-dependent forgetting~\citep{yang2023gated}. DeltaNet uses the delta rule to update stored associations~\citep{schlag2021linear,yang2024parallelizing}, while Gated DeltaNet combines delta updates with forgetting~\citep{yang2025gated}. Preconditioned DeltaNet instead introduces curvature-aware preconditioning into DeltaNet-style updates~\citep{pdn}. Kimi Delta Attention extends this family with a diagonal decay term and is deployed in Kimi-Linear, which interleaves recurrent and attention layers~\citep{team2025kimi}.  ParaTTT~\citep{loo2026paratttparallelizingtesttimetraining} composes these linear recurrent primitives to enhance expressivity. Large-scale hybrid models such as Nemotron-H and Qwen-3.5 likewise combine recurrent and softmax attention layers~\citep{blakeman2025nemotron,team2026qwen3}. LFM2 is a hardware-aware hybrid architecture that combines gated short convolutions with grouped-query attention for efficient edge inference~\citep{amini2025lfm2technicalreport}. 

\textbf{Post-training quantization of softmax Transformers.}
Post-training quantization (PTQ) of softmax Transformers has been studied extensively. Weight-only methods such as GPTQ use layerwise reconstruction to reduce low-bit weight quantization error~\citep{frantar2022gptq}. Joint weight and activation quantization must additionally contend with heavy-tailed activation distributions. SmoothQuant transfers activation scale into the weights~\citep{xiao2023smoothquant}, while incoherence-based methods apply a change of basis to spread concentrated energy across channels. QuIP and QuIP\# apply randomized incoherence processing for low-bit weight quantization~\citep{chee2023quip,tseng2024quip}. QuaRot extends rotational invariances to weights, activations, and the KV cache~\citep{ashkboos2024quarot}, while SpinQuant learns rotations from calibration data~\citep{liu2025spinquant}.

The activation outliers that motivate these methods have also been studied independently of quantization. \citet{sun2024massive} identify massive activations that can act as implicit bias terms. More recent work studies their relationship with attention sinks~\citep{sun2026spike,chen2026attention}. Our extra-state analysis is complementary: instead of introducing another outlier treatment, we ask whether changing the sequence mixer alters the outlier structure learned during training in the first place.

A second line of Transformer PTQ targets the KV cache. KIVI observes different quantization geometry for keys and values and uses per-channel key quantization and per-token value quantization~\citep{liu2024kivi}. KVQuant similarly exploits asymmetric key/value structure together with outlier-aware quantization~\citep{hooper2024kvquant}, while SKVQ combines channel reordering with groupwise quantization and a recent-token window retained at higher precision~\citep{duanmu2024skvq}. Rotation-based methods can also transform keys and values before cache quantization~\citep{ashkboos2024quarot,liu2025spinquant}. These methods operate on cache entries that are written once and subsequently reused, in contrast to a recurrent state that is modified in place after every generated token.

\textbf{Quantization of recurrent language models.}
Several recent methods adapt weight and activation PTQ to Mamba-family architectures. Quamba identifies sensitive feature maps inside the selective scan and large output activation outliers, and develops an 8-bit weight and activation quantization recipe around these statistics~\citep{chiang2025quamba}. MambaQuant likewise targets Mamba-specific activation distributions using variance-aligned rotations~\citep{yue2025mambaquant}. Quamba2 extends PTQ to Mamba and Mamba-2 across several weight and activation precisions ~\citep{chiang2025quamba2}. These works develop effective quantization recipes for particular recurrent architectures. Our extra-state study instead holds the surrounding architecture and quantization procedure fixed and asks how the sequence mixer itself changes the numerical properties of the learned weights and activations. LinQuant~\citep{tumma2026linquantfunctionalinvarianceslowprecision} studies functional invariances of linear recurrences for low-precision prefill, complementing our analysis of repeated state quantization during decode.

Q-Mamba directly quantizes Mamba state caches in addition to weights and activations. Its decoupled scale quantization accounts for outliers along multiple state dimensions and supports low-bit state caches ~\citep{tianqi2025qmamba}. Nemotron~3 Super studies reduced-precision Mamba state storage during autoregressive decode. It evaluates FP16 storage with stochastic rounding and block-scaled INT16, and shows that stochastic rounding helps control error accumulation across recurrent updates ~\citep{chandiramani2026nemotron}. These works establish that low-precision state storage is feasible for Mamba-family models.

SSDi8 considers a related but distinct setting. It maintains an INT8 representation through the internal Mamba-2 SSD computation and quantizes intermediate activations inside the chunked SSD algorithm~\citep{kim2026ssdi8}. Its primary target is therefore low-precision execution of the SSD computation.

Our intra-state study takes a different perspective from these architecture-specific recipes. We treat recurrent state quantization as a numerical problem across several recurrence families and isolate three sources of error: the geometry of scale grouping, accumulation under repeated requantization, and attenuation through recurrent decay. We then use these
properties to construct \methodname{} and evaluate its transfer to large hybrid language models.

Concurrent works DAMP~\citep{zhang2026damp} and STEPQuant~\citep{yao2026stepquant} allocate precision based on decay, with STEPQuant optimizing scales to reduce output error. Concurrent work LeapQuant~\citep{pan2026leapquant} reduces requantization frequency and preserves outliers in higher precision.\looseness-1

\textbf{Recurrent-state compression beyond quantization.}
Quantization reduces the precision of each stored state, while an orthogonal line of work improves recurrent efficiency by changing the dimensionality or structure of the recurrence. \citet{chahine2026curious} use ideas from balanced truncation and Hankel singular values to reduce state dimension during training, while \citet{parnichkun2025quantifying} investigate effective state size in linear operators. For pretrained linear attention models, \citet{nazari2026key} identify low-rank recurrent states and structurally prune query and key dimensions after training. Related work has also studied low-rank and sparse recurrent structure outside language modeling~\citep{tumma2023leveraginglowranksparserecurrent}.

\section{Experimental Setup}
\label{app:expdetails}

\subsection{Models and training}
\label{app:zoo-training}
All mixer zoo checkpoints share the same $1.3$B-parameter backbone: $24$ layers, hidden size $2048$, an MLP ratio of $4$, and $16$ mixer heads of head dimension $128$ per layer, both along the key- and the value axis. All sequence mixer layers share the same scaffold, using short convolutions on queries, keys and values~\citep{chollet2017xception,so2021searching,fu2022hungry} followed by SiLU and RoPE. The output is generated using a gated RMSNorm before an output projection. All models are trained for $100$B tokens with the same recipe and a learning rate of $4\times10^{-4}$.

\subsection{Quantization and calibration}
\label{app:wa-setup}

For hybrid models, we use GPTQ for weight quantization~\citep{frantar2022gptq} through llm-compressor~\citep{llmcompressor2024}. We use groups of $128$ input channels, except that W8A8 uses per-output-channel weight scales, and calibrate on $128$ sequences of $2048$ tokens from the \WT{} \emph{train} split. Activations are quantized per token. The rotations described in Section~\ref{sec:external} are active during calibration, and foldable rotations are absorbed into the weights before quantization. The language-model head remains in BF16. For Kimi-Linear, every expert is calibrated on all calibration tokens, while the router remains in BF16.

\paragraph{Mixer zoo.}
We use GPTQ with symmetric per-output-channel weight scales and calibrate on $128$ sequences of $2048$ tokens from the \WT{} \emph{train} split. Dynamic activation scales are computed per token, with groups of $128$ channels at mixer output-projection inputs. Decay and write-gate projections use the same W/A precision as the other block linears, and the rotations use Hadamard matrices. The unquantized baseline uses BF16. Embeddings and the language-model head remain in BF16. We evaluate word perplexity on the full \WT{} test split with a maximum sequence length of $2048$.

\subsection{Evaluation protocols}
\label{app:gen-protocol}

Our generative evaluations follow a common protocol, with model-specific sampling parameters and reasoning modes. Qwen-3.5 uses temperature $0.6$, top-$p$ $0.95$, top-$k$ $20$, following the ``precise'' thinking-mode preset in its model card~\citep{team2026qwen3}. Nemotron-H uses temperature $0.6$ and top-$p$ $0.95$, as recommended in the model card~\citep{blakeman2025nemotron}. Kimi-Linear uses temperature $1.0$ and top-$p$ $0.95$. While Qwen-3.5 has reasoning enabled by default, we explicitly activate it for all evals but RULER for Nemotron-H.

\paragraph{Generation budgets.} GSM8K decodes up to $8192$ new tokens. AIME, GPQA (diamond), and MMLU-Pro decode up to $65536$. GSM8K uses the 8-shot chain-of-thought prompt.

\paragraph{Answer-format instructions.} Our GPQA evaluations use the following NeMo-Skills multiple-choice instruction for all models: ``Answer the following multiple choice question. The last line of your response should be in the following format: 'Answer: A/B/C/D' (e.g.\ 'Answer: A').'' The question and four lettered choices follow this instruction. For AIME, all models use the NeMo-Skills math instruction ``Solve the following math problem. Make sure to put the answer (and only answer) inside \texttt{\textbackslash boxed\{\}}.'' MMLU-Pro uses the corresponding ten-choice instruction with the format line ``Answer: A/B/C/D/E/F/G/H/I/J,'' options listed in dataset order, and no few-shot examples. We omit the math harness's base-model stop strings (\texttt{Q:}, \texttt{Question:}, and \texttt{Problem:}) because we find they can also occur within a reasoning trace.

\paragraph{Scoring.} AIME results pool the 2024 and 2025 competitions, each containing $30$ problems. AIME responses are graded with the NeMo-Skills math grader: the integer in the last valid \texttt{\textbackslash boxed\{\}} expression, compared with the reference by Math-Verify. For GSM8K, we select the last boxed answer, then the last explicit answer cue, and otherwise fall back to Math-Verify's mathematical extraction. Numeric answers are compared exactly, without integer rounding. Other expressions use the NeMo-Skills math grader~\citep{blakeman2025nemotron}. For GPQA and MMLU-Pro, a rule modeled on the NeMo-Skills multiple-choice grader extracts the choice from the last valid \texttt{Answer: X} line and otherwise uses the last \texttt{\textbackslash boxed\{X\}} expression, where X is A--D for GPQA and A--J for MMLU-Pro. Our implementation also accepts a parenthesized choice after the answer cue, as in \texttt{Answer: (C)}. Responses without a valid choice are marked incorrect.

\section{Proofs}
\label{app:proofs}
\begin{proof}[\textbf{Proof of Proposition~\ref{prop:sq-prices}}]
\label{app:proof-prop-sq-prices}
For fixed $\bS$ and $\bmq$, the readout error in component $i$ is $e_i=\sum_j\bE_{ij}q_j$. If an entry belongs to a group with scale $s_G$, the high-resolution approximation gives $\bE_{ij}\sim\operatorname{Unif}[-s_G/2,s_G/2]$, so
\begin{equation*}
\mathbb{E}_{\bE}[\bE_{ij}]=0,
\qquad
\mathbb{E}_{\bE}[\bE_{ij}^2]=\frac{s_G^2}{12}.
\end{equation*}
Independence and zero means imply $\mathbb{E}_{\bE}[\bE_{ij}\bE_{ik}]=0$ for $j\ne k$. Expanding the square therefore yields
\begin{equation}
\mathbb{E}_{\bE}[e_i^2]
=\sum_jq_j^2\mathbb{E}_{\bE}[\bE_{ij}^2]
+2\sum_{j<k}q_jq_k\mathbb{E}_{\bE}[\bE_{ij}\bE_{ik}]
=\sum_jq_j^2\mathbb{E}_{\bE}[\bE_{ij}^2].
\label{eq:readout-second-moment}
\end{equation}
Under value-axis grouping, every entry in row $i$ has scale $s_i^{\mathrm{val}}=M_i/r$. Substitution into~\eqref{eq:readout-second-moment} gives
\begin{equation*}
\mathbb{E}_{\bE}\!\left[(e_i^{\mathrm{val}})^2\right]
=\frac{M_i^2}{12r^2}\sum_jq_j^2
=\frac{M_i^2}{12r^2},
\end{equation*}
Linear attention uses unit-norm queries, $\lVert\bmq\rVert_2=1$, which gives the final equality. Under key-axis grouping, every entry in column $j$ has scale $s_j^{\mathrm{key}}=C_j/r$, hence
\begin{equation*}
\mathbb{E}_{\bE}\!\left[(e_i^{\mathrm{key}})^2\right]
=\frac{1}{12r^2}\sum_jq_j^2C_j^2.
\end{equation*}
These are the two expressions in~\eqref{eq:sq-prices}.
\end{proof}

\paragraph{Shared error propagation setup for Propositions~\ref{prop:sq-decay} and~\ref{prop:rtn-dead-zone}.}
For the fixed update operands in~\Eqref{eq:sq-error-recurrence}, define
\begin{equation*}
\mathbf A_t=\operatorname{diag}(\bm\alpha_t)
(\bI-\delta\,\beta_t\bmk_t\bmk_t^\top),
\qquad
\mathbf P_{\tau,t}=\mathbf A_{\tau+1}\cdots\mathbf A_t,
\qquad
\mathbf P_{t,t}=\bI.
\end{equation*}
The states start identically, so $\bE_0=0$. Unrolling~\Eqref{eq:sq-error-recurrence} yields
\begin{equation}
\bE_t=\sum_{\tau=1}^t\bm{\varepsilon}_\tau\mathbf P_{\tau,t}.
\label{eq:sq-error-unrolled}
\end{equation}
Each product $\mathbf P_{\tau,t}$ propagates an injection from step $\tau$ to step $t$.

\begin{proof}[\textbf{Proof of Proposition~\ref{prop:sq-decay}}]
\label{app:proof-sq-decay}
Because $\lVert\bmk_t\rVert_2=1$, the matrix $\bI-\delta\,\beta_t\bmk_t\bmk_t^\top$ acts as the identity on the orthogonal complement of $\bmk_t$ and has eigenvalue $1-\delta\,\beta_t$ along $\bmk_t$. The ranges $\delta\in\{0,1\}$ and $\beta_t\in(0,2)$ therefore give $\lVert\bI-\delta\,\beta_t\bmk_t\bmk_t^\top\rVert_2\le1$. Since $\lVert\operatorname{diag}(\bm\alpha_t)\rVert_2\le\alpha$, we have $\lVert\mathbf A_t\rVert_2\le\alpha$ and $\lVert\mathbf P_{\tau,t}\rVert_2\le\alpha^{t-\tau}$. Applying the triangle inequality to~\Eqref{eq:sq-error-unrolled} gives
\begin{equation*}
\lVert\bE_t\rVert_F
\le\sum_{\tau=1}^t\lVert\bm{\varepsilon}_\tau\mathbf P_{\tau,t}\rVert_F
\le\sum_{\tau=1}^t\alpha^{t-\tau}\lVert\bm{\varepsilon}_\tau\rVert_F.
\end{equation*}
This argument imposes no independence or alignment assumptions on the injections.
\end{proof}

\begin{proof}[\textbf{Proof of Proposition~\ref{prop:rtn-dead-zone}}]
\label{app:proof-rtn-dead-zone}
Expanding the squared Frobenius norm in~\Eqref{eq:sq-error-unrolled} gives, for any rounding rule,
\begin{equation}
\begin{aligned}
\lVert\bE_t\rVert_F^2
&=\sum_{\tau=1}^t\lVert\bm{\varepsilon}_\tau\mathbf P_{\tau,t}\rVert_F^2\\
&\quad+2\sum_{1\le\tau<s\le t}
\bigl\langle\bm{\varepsilon}_\tau\mathbf P_{\tau,t},
\bm{\varepsilon}_s\mathbf P_{s,t}\bigr\rangle_F.
\end{aligned}
\label{eq:sq-error-interactions}
\end{equation}
In particular, this is the RTN decomposition for the full recurrence, including diagonal decay and delta correction.

Let $\mathcal H_s$ denote the history immediately before rounding at step $s$, including the quantization scales. Under conditionally unbiased SR, $\mathbb E[\bm{\varepsilon}_s\mid\mathcal H_s]=0$. For $\tau<s$, the earlier injection $\bm{\varepsilon}_\tau$ is $\mathcal H_s$-measurable, and the propagation matrices are fixed independently of rounding. Thus
\begin{equation*}
\begin{aligned}
\mathbb E\bigl[
\langle\bm{\varepsilon}_\tau\mathbf P_{\tau,t},
\bm{\varepsilon}_s\mathbf P_{s,t}\rangle_F\bigr]
&=\mathbb E\bigl[
\langle\bm{\varepsilon}_\tau\mathbf P_{\tau,t},
\mathbb E[\bm{\varepsilon}_s\mid\mathcal H_s]\mathbf P_{s,t}\rangle_F\bigr]\\
&=0.
\end{aligned}
\end{equation*}
Taking expectations in~\Eqref{eq:sq-error-interactions} therefore yields
\begin{equation}
\mathbb E\!\left[\lVert\bE_t^{\mathrm{SR}}\rVert_F^2\right]
=\sum_{\tau=1}^t
\mathbb E\!\left[\lVert\bm{\varepsilon}_\tau^{\mathrm{SR}}\mathbf P_{\tau,t}\rVert_F^2\right].
\label{eq:sq-sr-propagation}
\end{equation}
The quantization scales may depend on previous rounding errors; conditional unbiasedness suffices and does not require independent injections. For scalar decay without delta correction, $\mathbf A_t=\alpha_t\bI$ and $\mathbf P_{\tau,t}=\gamma_{\tau,t}\bI$, where $\gamma_{\tau,t}=\prod_{j=\tau+1}^t\alpha_j$. Substituting into~\Eqref{eq:sq-error-interactions} and~\Eqref{eq:sq-sr-propagation} proves the two identities in the proposition.

Under the contraction assumption of Proposition~\ref{prop:sq-decay}, the general SR identity also gives
\begin{equation*}
\mathbb E\!\left[\lVert\bE_t^{\mathrm{SR}}\rVert_F^2\right]
\le\sum_{\tau=1}^t\alpha^{2(t-\tau)}
\mathbb E\!\left[\lVert\bm{\varepsilon}_\tau^{\mathrm{SR}}\rVert_F^2\right].
\end{equation*}

To show that RTN reinforcement can persist under decay, let $\bS_t=\alpha\bS_{t-1}+u\mathbf B$, where $0<\alpha<1$ and $\mathbf B$ has a single unit entry and is zero elsewhere. Start both states at zero and round elementwise to the fixed grid $h\mathbb Z$, with $0<u<h/2$. RTN discards the write at every step, so $\widehat{\bS}_t=0$ and $\bm{\varepsilon}_t^{\mathrm{RTN}}=-u\mathbf B$. Consequently,
\begin{equation*}
\bigl\langle\bm{\varepsilon}_\tau^{\mathrm{RTN}},
\bm{\varepsilon}_s^{\mathrm{RTN}}\bigr\rangle_F=u^2>0
\qquad(\tau<s).
\end{equation*}
Thus positive cross terms can occur over arbitrarily many successive steps, even with strict decay.
\end{proof}

\section{Extra-State Quantization}
\label{app:extra-state}

\subsection{Quantization results and rotation ablations}
\label{app:ladder-full}

Table~\ref{tab:external-ladder-absolute} reports the full weight--activation precision ladder in absolute \WT{} perplexity, including the unquantized baseline. The main text (Table~\ref{tab:external-ladder}) shows relative changes for unrotated W4A8 and W4A4, together with W4A4 rotation ablations.

\begin{table*}[htbp]
\centering
\footnotesize
\setlength{\tabcolsep}{1.3pt}
\renewcommand{\arraystretch}{1.12}
\caption{Complete absolute \WT{} perplexity ladder. Baseline is the unquantized BF16 model. $16$-bit entries denote BF16.}
\label{tab:external-ladder-absolute}
\begin{tabular}{@{}lccc r rrrr>{\columncolor{ezHl}}r rrrr>{\columncolor{ezHl}[\tabcolsep][0pt]}r@{}}
\toprule
\multicolumn{4}{c}{Architectural choices} & & \multicolumn{5}{c}{No rotation} & \multicolumn{5}{c}{Full stack $\bR_1\bR_4\rout$} \\
\cmidrule(lr){1-4}\cmidrule(lr){6-10}\cmidrule(lr){11-15}
\dcol{Decay} & $\|\mathbf Q \mathbf K\|$ & \gcol{$\beta$} & \ecol{$\Delta$}
 & \numhead{Baseline} & \numhead{W8A8} & \numhead{W4A16} & \numhead{W4A8} & \numhead{W16A4} & \numhead{W4A4} & \numhead{W8A8} & \numhead{W4A16} & \numhead{W4A8} & \numhead{W16A4} & \numhead{W4A4} \\
\midrule
$-$ & $-$ & $-$ & $-$ & 15.66 & 15.71 & 16.52 & 16.57 & 26.47 & 31.07 & 15.67 & 16.34 & 16.34 & 16.44 & 17.33 \\
$-$ & $\checkmark$ & $-$ & $-$ & 15.66 & 15.70 & 16.53 & 16.58 & 25.87 & 30.11 & 15.67 & 16.40 & 16.40 & 16.46 & 17.40 \\
\midrule
$-$ & $\checkmark$ & $-$ & $-$ & 17.93 & 17.98 & 19.06 & 19.15 & 31.72 & 37.91 & 17.93 & 18.84 & 18.86 & 18.95 & 20.16 \\
$-$ & $\checkmark$ & \gcol{\checkmark} & $-$ & 18.05 & 18.10 & 19.24 & 19.30 & 28.54 & 32.93 & 18.06 & 19.01 & 19.02 & 19.12 & 20.30 \\
$-$ & $\checkmark$ & \gcol{\checkmark} & \ecol{\checkmark} & 17.72 & 17.77 & 18.72 & 18.80 & 26.32 & 29.80 & 17.72 & 18.61 & 18.61 & 18.73 & 19.97 \\

\addlinespace[2pt]
\dcol{Scal} & $\checkmark$ & $-$ & $-$ & 16.94 & 16.99 & 17.93 & 17.97 & 23.68 & 26.53 & 16.95 & 17.78 & 17.80 & 17.87 & 18.96 \\
\dcol{Scal} & $\checkmark$ & \gcol{\checkmark} & $-$ & 17.19 & 17.23 & 18.09 & 18.15 & 23.82 & 26.40 & 17.19 & 18.01 & 18.02 & 18.07 & 19.13 \\
\dcol{Scal} & $\checkmark$ & \gcol{\checkmark} & \ecol{\checkmark} & 17.10 & 17.15 & 18.09 & 18.12 & 23.90 & 26.58 & 17.12 & 17.94 & 17.94 & 18.03 & 19.12 \\

\addlinespace[2pt]
\dcol{Diag} & $\checkmark$ & $-$ & $-$ & 15.86 & 15.88 & 16.58 & 16.62 & 20.95 & 22.99 & 15.86 & 16.50 & 16.50 & 16.55 & 17.33 \\
\dcol{Diag} & $\checkmark$ & \gcol{\checkmark} & $-$ & 16.08 & 16.12 & 16.89 & 16.95 & 21.75 & 23.99 & 16.09 & 16.83 & 16.82 & 16.83 & 17.78 \\
\dcol{Diag} & $\checkmark$ & \gcol{\checkmark} & \ecol{\checkmark} & 16.55 & 16.58 & 17.46 & 17.51 & 22.13 & 24.46 & 16.55 & 17.30 & 17.29 & 17.42 & 18.40 \\
\bottomrule
\end{tabular}
\end{table*}

Table~\ref{tab:external-rotation-ablation} ablates the full rotation stack $\bR_1\bR_4\rout$ on the $1.3$B mixer zoo. Among the individual rotations, $\bR_1$ is more effective for the softmax models and for plain linear attention, whereas $\bR_4$ is more effective for every recurrent cell that carries a decay mechanism, a write gate, or a delta-rule correction. This shows that decayed linear recurrence PTQ error is not dominated by outliers in the residual stream.

\begin{table*}[htbp]
\centering
\small
\setlength{\tabcolsep}{4.0pt}
\caption{Relative \WT{} perplexity increase (\%) under W4A4 with different rotation subsets on the $1.3$B mixer zoo.}
\label{tab:external-rotation-ablation}
\begin{tabular}{@{}lccc rrrrrrrr@{}}
\toprule
\multicolumn{4}{c}{Architectural choices} & \multicolumn{8}{c}{$\Delta_{\mathrm{PPL}}$ (\%)} \\
\cmidrule(lr){1-4}\cmidrule(lr){5-12}
\dcol{Decay} & $\|\mathbf Q \mathbf K\|$ & \gcol{$\beta$} & \ecol{$\Delta$}
 & \numhead{none} & \numhead{$\bR_1$} & \numhead{$\bR_4$} & \numhead{$\rout$} & \numhead{$\bR_1\bR_4$} & \numhead{$\bR_1\rout$} & \numhead{$\bR_4\rout$} & \numhead{$\bR_1\bR_4\rout$} \\
\midrule
$-$                   & $-$          & $-$               & $-$               & 98.4 & 36.7 & 47.9 & 90.1 & 12.2 & 34.4 & 44.3 & \textbf{10.6} \\
$-$                   & $\checkmark$ & $-$               & $-$               & 92.3 & 37.8 & 46.5 & 86.0 & 12.9 & 35.7 & 42.8 & \textbf{11.1} \\
\midrule
$-$                   & $\checkmark$ & $-$               & $-$               & 111.4 & 48.1 & 54.0 & 105.9 & 14.0 & 45.5 & 50.5 & \textbf{12.5} \\
$-$                   & $\checkmark$ & \gcol{\checkmark} & $-$               & 82.4 & 47.2 & 35.9 & 77.9 & 13.9 & 44.8 & 33.2 & \textbf{12.4} \\
$-$                   & $\checkmark$ & \gcol{\checkmark} & \ecol{\checkmark} & 68.2 & 46.5 & 28.5 & 65.0 & 13.9 & 43.7 & 26.3 & \textbf{12.7} \\

\addlinespace[2pt]
\dcol{Scal}         & $\checkmark$ & $-$               & $-$               & 56.6 & 40.3 & 24.7 & 54.8 & 13.0 & 38.1 & 23.3 & \textbf{11.9} \\
\dcol{Scal}         & $\checkmark$ & \gcol{\checkmark} & $-$               & 53.5 & 41.3 & 21.3 & 51.2 & 12.6 & 38.8 & 19.5 & \textbf{11.3} \\
\dcol{Scal}         & $\checkmark$ & \gcol{\checkmark} & \ecol{\checkmark} & 55.4 & 43.2 & 21.5 & 52.8 & 13.1 & 40.6 & 19.8 & \textbf{11.8} \\

\addlinespace[2pt]
\dcol{Diag}       & $\checkmark$ & $-$               & $-$               & 45.0 & 32.6 & 18.3 & 42.8 & 10.4 & 31.0 & 16.7 & \textbf{9.3} \\
\dcol{Diag}       & $\checkmark$ & \gcol{\checkmark} & $-$               & 49.2 & 35.7 & 20.7 & 46.8 & 11.6 & 34.4 & 19.3 & \textbf{10.6} \\
\dcol{Diag}       & $\checkmark$ & \gcol{\checkmark} & \ecol{\checkmark} & 47.8 & 37.0 & 19.5 & 45.6 & 12.5 & 34.8 & 18.1 & \textbf{11.2} \\
\bottomrule
\end{tabular}
\end{table*}

\subsection{Residual outliers and interventions}

\begin{figure*}[htbp]
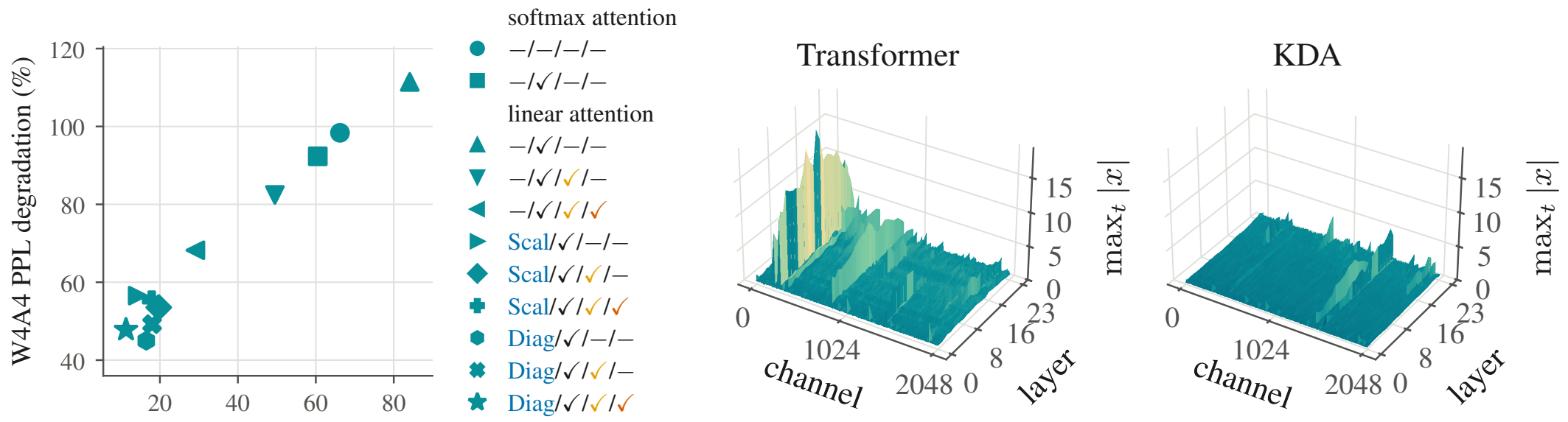

\centering
\begin{subfigure}[b]{0.44\textwidth}
\centering
\resizebox{\linewidth}{!}{\input{media/kurtosis_law.pgf}}
\caption{Kurtosis versus W4A4 degradation.}
\label{fig:external-law-curtosis}
\end{subfigure}\hfill
\begin{subfigure}[b]{0.54\textwidth}
\centering
\resizebox{\linewidth}{!}{\input{media/residual_skyline_layers.pgf}}
\caption{Mixer-input magnitude by channel and layer.}
\label{fig:skyline-layers}
\end{subfigure}
\caption{Subfigure~(\subref{fig:external-law-curtosis}) relates residual-stream kurtosis to unrotated W4A4 perplexity degradation. Legend entries denote \dcol{decay}/$\|\mathbf Q \mathbf K\|$/\gcol{$\beta$}/\ecol{$\Delta$}, as in Table~\ref{tab:external-ladder}. Subfigure~(\subref{fig:skyline-layers}) shows the mixer-input magnitude over $128$ tokens.}
\label{fig:external-outliers-joint}
\end{figure*}

Figure~\ref{fig:external-law-curtosis} relates residual-stream kurtosis to unrotated W4A4 degradation across the mixer zoo.

Figure~\ref{fig:skyline-layers} shows that the Transformer builds an outlier channel that persists across layers and complements Figure~\ref{fig:softmax-mechanism}.

Table~\ref{tab:external-outlier-full} extends Table~\ref{tab:external-outlier} to every cell of the $1.3$B mixer zoo. Removing the dominant channel, either at the mixer input or before individual projections, has a much larger effect in the softmax cells than in the recurrent cells.

\begin{table}[htbp]
\centering
\small
\setlength{\tabcolsep}{4.0pt}
\renewcommand{\arraystretch}{1.12}
\caption{Absolute \WT{} perplexity under the dominant-channel interventions of Table~\ref{tab:external-outlier}. \emph{None} is the unmodified BF16 model. Under \emph{removed}, QK/V delete the channel only before the corresponding projections and QKV deletes it at the mixer input. \emph{mean} replaces QKV at the mixer input by its token average.}
\label{tab:external-outlier-full}
\begin{tabular}{@{}lccc rrrrr@{}}
\toprule
\multicolumn{4}{c}{Architectural choices} & \multicolumn{1}{c}{none} & \multicolumn{3}{c}{removed} & \multicolumn{1}{c}{mean} \\
\cmidrule(lr){1-4}\cmidrule(lr){5-5}\cmidrule(lr){6-8}\cmidrule(lr){9-9}
\dcol{Decay} & $\|\mathbf Q \mathbf K\|$ & \gcol{$\beta$} & \ecol{$\Delta$} & & \numhead{QK} & \numhead{V} & \numhead{QKV} & \numhead{QKV} \\
\midrule
$-$             & $-$          & $-$               & $-$               & 15.66 & 280.51 & 16.86 & 448.70 & 16.89 \\
$-$             & $\checkmark$ & $-$               & $-$               & 15.66 & 81.30 & 17.18 & 76.61 & 16.08 \\
\midrule
$-$             & $\checkmark$ & $-$               & $-$               & 17.93 & 21.23 & 18.20 & 23.62 & 19.81 \\
$-$             & $\checkmark$ & \gcol{\checkmark} & $-$               & 18.05 & 18.89 & 18.21 & 20.64 & 18.62 \\
$-$             & $\checkmark$ & \gcol{\checkmark} & \ecol{\checkmark} & 17.72 & 18.42 & 18.10 & 21.91 & 17.96 \\

\addlinespace[2pt]
\dcol{Scal}     & $\checkmark$ & $-$               & $-$               & 16.94 & 17.41 & 17.12 & 18.21 & 17.16 \\
\dcol{Scal}     & $\checkmark$ & \gcol{\checkmark} & $-$               & 17.19 & 17.76 & 17.36 & 19.65 & 17.38 \\
\dcol{Scal}     & $\checkmark$ & \gcol{\checkmark} & \ecol{\checkmark} & 17.10 & 17.52 & 17.22 & 18.78 & 17.22 \\

\addlinespace[2pt]
\dcol{Diag}     & $\checkmark$ & $-$               & $-$               & 15.86 & 16.64 & 15.98 & 18.81 & 16.26 \\
\dcol{Diag}     & $\checkmark$ & \gcol{\checkmark} & $-$               & 16.08 & 16.85 & 16.35 & 19.93 & 16.27 \\
\dcol{Diag}     & $\checkmark$ & \gcol{\checkmark} & \ecol{\checkmark} & 16.55 & 16.91 & 16.74 & 18.19 & 16.83 \\
\bottomrule
\end{tabular}
\end{table}
\section{State Quantization}
\label{app:state-quant}

\subsection{Grouping, mixed precision, and rounding}

This subsection reports the additional ablations referenced in Section~\ref{ch:state-quant}.

Table~\ref{tab:sq-perstate} compares per-axis with per-state scale grouping. Both value- and key-axis grouping outperform per-state grouping by a wide margin.

\begin{table}[htbp]
\centering
\small
\caption{Median cross-entropy increase $\Delta\ell$ under INT6 with per-state, value-axis, and key-axis scaling.}
\label{tab:sq-perstate}
\begin{tabular}{lrrr}\toprule
model & \numhead{state} & \numhead{value} & \numhead{key}\\\midrule
Linear & 2.358 & 1.230 & 0.967\\
Mamba-2 & 0.516 & 0.027 & 0.052\\
GLA & 0.377 & 0.005 & 0.034\\
GDN & 0.158 & 0.004 & 0.025\\
KDA & 0.361 & 0.004 & 0.033\\
\bottomrule
\end{tabular}
\end{table}

Table~\ref{tab:sq-mp-random} compares the decay-based INT5 allocation from Section~\ref{sec:sq-mixedprec} with a matched random allocation.

\begin{table}[htbp]
\centering
\small
\caption{Median cross-entropy increase $\Delta\ell$ for decay-based and random INT5 allocations under RTN.}
\label{tab:sq-mp-random}
\begin{tabular}{lrr}\toprule
model & \numhead{decay} & \numhead{random}\\\midrule
Mamba-2 & \textbf{0.033} & 0.203\\
GLA & \textbf{0.009} & 0.073\\
GDN & \textbf{0.005} & 0.064\\
KDA & \textbf{0.005} & 0.108\\
\midrule
Qwen-3.5 & \textbf{0.006} & 0.267\\
Nemotron-H & \textbf{0.045} & 0.158\\
Kimi-Linear & \textbf{0.006} & 0.028\\
\bottomrule
\end{tabular}
\end{table}

\begin{table}[htbp]
\centering
\footnotesize
\setlength{\tabcolsep}{3pt}
\caption{Median cross-entropy increase $\Delta\ell$ for RTN and SR with value- and key-axis grouping and symmetric grids. Bold marks best per precision.}
\label{tab:sq-rounding-full}
\begin{tabular}{@{}lrrrrrrrrrrrr@{}}\toprule
& \multicolumn{4}{c}{INT8} & \multicolumn{4}{c}{INT6} & \multicolumn{4}{c}{INT4}\\
\cmidrule(lr){2-5}\cmidrule(lr){6-9}\cmidrule(lr){10-13}
& \multicolumn{2}{c}{value} & \multicolumn{2}{c}{key} & \multicolumn{2}{c}{value} & \multicolumn{2}{c}{key} & \multicolumn{2}{c}{value} & \multicolumn{2}{c}{key}\\
\cmidrule(lr){2-3}\cmidrule(lr){4-5}\cmidrule(lr){6-7}\cmidrule(lr){8-9}\cmidrule(lr){10-11}\cmidrule(lr){12-13}
model & \numhead{RTN} & \numhead{SR} & \numhead{RTN} & \numhead{SR} & \numhead{RTN} & \numhead{SR} & \numhead{RTN} & \numhead{SR} & \numhead{RTN} & \numhead{SR} & \numhead{RTN} & \numhead{SR}\\\midrule
Linear & \textbf{0.125} & 0.223 & 0.332 & 0.135 & 1.230 & 1.654 & \textbf{0.967} & 1.019 & 3.647 & 4.338 & \textbf{2.479} & 3.884\\
Mamba-2 & 0.003 & \textbf{0.001} & 0.011 & \textbf{0.001} & 0.027 & 0.035 & 0.052 & \textbf{0.011} & 0.521 & 0.813 & 0.400 & \textbf{0.342}\\
GLA & \textbf{0.000} & \textbf{0.000} & 0.005 & 0.001 & 0.005 & \textbf{0.002} & 0.034 & 0.009 & 0.269 & \textbf{0.103} & 0.309 & 0.266\\
GDN & \textbf{0.000} & \textbf{0.000} & 0.005 & 0.001 & \textbf{0.004} & 0.005 & 0.025 & 0.019 & \textbf{0.202} & 0.391 & 0.313 & 0.650\\
KDA & \textbf{0.000} & \textbf{0.000} & 0.007 & \textbf{0.000} & \textbf{0.004} & \textbf{0.004} & 0.033 & 0.010 & 0.276 & \textbf{0.218} & 0.266 & 0.328\\
\midrule
Qwen-3.5 & 0.001 & \textbf{0.000} & 0.010 & 0.877 & 0.004 & \textbf{0.002} & 0.157 & 7.493 & \textbf{0.825} & 2.632 & 1.037 & 8.818\\
Nemotron-H & 0.001 & \textbf{0.000} & 0.019 & 6.513 & 0.010 & \textbf{0.003} & 0.183 & 7.574 & 0.763 & \textbf{0.520} & 1.227 & 9.453\\
Kimi-Linear & 0.001 & \textbf{0.000} & 0.001 & \textbf{0.000} & 0.005 & \textbf{0.001} & 0.009 & \textbf{0.001} & 0.262 & 0.153 & \textbf{0.056} & 0.141\\
\bottomrule
\end{tabular}
\end{table}

\subsection{Rotations and formats}
\label{app:sq-rotations-formats}

\begin{table}[htbp]
\centering
\footnotesize
\setlength{\tabcolsep}{2pt}
\caption{Median cross-entropy increase $\Delta\ell$ under within-group rotations for SR with symmetric grids. Bold marks best per precision.}
\label{tab:sq-rot-sr-full}
\begin{tabular}{@{}lrrrrrrrrrrrr@{}}\toprule
& \multicolumn{4}{c}{INT8} & \multicolumn{4}{c}{INT6} & \multicolumn{4}{c}{INT4}\\
\cmidrule(lr){2-5}\cmidrule(lr){6-9}\cmidrule(lr){10-13}
model & \numhead{value} & \numhead{$+\bR_k$} & \numhead{key} & \numhead{$+\bR_v$} & \numhead{value} & \numhead{$+\bR_k$} & \numhead{key} & \numhead{$+\bR_v$} & \numhead{value} & \numhead{$+\bR_k$} & \numhead{key} & \numhead{$+\bR_v$}\\\midrule
Linear & 0.2229 & 0.0380 & 0.1349 & \textbf{0.0299} & 1.6545 & 1.1161 & 1.0191 & \textbf{0.5310} & 4.3383 & 4.3309 & 3.8839 & \textbf{3.2264}\\
Mamba-2 & 0.0015 & 0.0005 & 0.0005 & \textbf{0.0001} & 0.0352 & 0.0412 & 0.0107 & \textbf{0.0034} & 0.8127 & 1.1788 & 0.3422 & \textbf{0.1511}\\
GLA & 0.0001 & \textbf{0.0000} & 0.0005 & 0.0001 & 0.0020 & \textbf{0.0019} & 0.0095 & 0.0029 & \textbf{0.1026} & 0.4735 & 0.2662 & 0.1129\\
GDN & 0.0001 & 0.0001 & 0.0007 & \textbf{0.0000} & 0.0048 & 0.0108 & 0.0187 & \textbf{0.0034} & \textbf{0.3911} & 1.3706 & 0.6502 & 0.4049\\
KDA & 0.0002 & \textbf{0.0000} & 0.0005 & 0.0001 & 0.0043 & \textbf{0.0026} & 0.0098 & 0.0035 & 0.2183 & 0.8500 & 0.3283 & \textbf{0.1758}\\
\midrule
Qwen-3.5 & \textbf{0.0001} & 0.0006 & 0.8767 & 0.0002 & \textbf{0.0016} & 0.0236 & 7.4927 & 0.1205 & \textbf{2.6319} & 5.4741 & 8.8185 & 7.5207\\
Nemotron-H & 0.0001 & \textbf{0.0000} & 6.5132 & 4.5026 & 0.0026 & \textbf{0.0011} & 7.5745 & 7.2219 & 0.5198 & \textbf{0.1926} & 9.4528 & 8.9028\\
Kimi-Linear & \textbf{0.0000} & \textbf{0.0000} & 0.0002 & \textbf{0.0000} & 0.0007 & 0.0015 & 0.0007 & \textbf{0.0003} & 0.1533 & 1.1315 & 0.1414 & \textbf{0.0484}\\
\bottomrule
\end{tabular}
\end{table}

\begin{table}[htbp]
\centering
\footnotesize
\setlength{\tabcolsep}{2pt}
\caption{One-step quantization error of the state: median of $\lVert Q(\bS)-\bS\rVert_F/\lVert\bS\rVert_F$ over layers, heads, and decode steps under SR.}
\label{tab:sq-oneshot-rot}
\begin{tabular}{@{}lrrrrrrrrrrrr@{}}\toprule
& \multicolumn{4}{c}{INT8} & \multicolumn{4}{c}{INT6} & \multicolumn{4}{c}{INT4}\\
\cmidrule(lr){2-5}\cmidrule(lr){6-9}\cmidrule(lr){10-13}
model & \numhead{key} & \numhead{$+\bR_v$} & \numhead{value} & \numhead{$+\bR_k$} & \numhead{key} & \numhead{$+\bR_v$} & \numhead{value} & \numhead{$+\bR_k$} & \numhead{key} & \numhead{$+\bR_v$} & \numhead{value} & \numhead{$+\bR_k$}\\\midrule
Mamba-2 & 0.013 & 0.009 & 0.020 & 0.008 & 0.052 & 0.036 & 0.082 & 0.034 & 0.229 & 0.159 & 0.351 & 0.152\\
GLA & 0.013 & 0.009 & 0.018 & 0.009 & 0.052 & 0.036 & 0.073 & 0.035 & 0.230 & 0.159 & 0.305 & 0.154\\
GDN & 0.012 & 0.009 & 0.015 & 0.009 & 0.049 & 0.036 & 0.060 & 0.036 & 0.216 & 0.161 & 0.263 & 0.160\\
KDA & 0.012 & 0.009 & 0.021 & 0.008 & 0.049 & 0.036 & 0.083 & 0.033 & 0.217 & 0.160 & 0.329 & 0.148\\
\midrule
Qwen-3.5 & 0.027 & 0.008 & 0.016 & 0.009 & 0.106 & 0.031 & 0.067 & 0.036 & 0.407 & 0.137 & 0.290 & 0.159\\
Nemotron-H & 0.016 & 0.007 & 0.014 & 0.009 & 0.064 & 0.030 & 0.057 & 0.035 & 0.264 & 0.132 & 0.236 & 0.157\\
Kimi-Linear & 0.015 & 0.009 & 0.021 & 0.008 & 0.060 & 0.037 & 0.084 & 0.034 & 0.259 & 0.162 & 0.330 & 0.152\\
\bottomrule\end{tabular}
\end{table}

\begin{table}[htbp]
\centering
\small
\setlength{\tabcolsep}{4pt}
\caption{Median cross-entropy increase $\Delta\ell$ for INT8 and E4M3 under RTN and SR with uniform grouping and symmetric grids. Bold marks best per row.}
\label{tab:sq-fp-rounding}
\begin{tabular}{@{}lrrrrrrrr@{}}\toprule
& \multicolumn{4}{c}{value} & \multicolumn{4}{c}{key}\\
\cmidrule(lr){2-5}\cmidrule(lr){6-9}
& \multicolumn{2}{c}{INT8} & \multicolumn{2}{c}{E4M3} & \multicolumn{2}{c}{INT8} & \multicolumn{2}{c}{E4M3}\\
\cmidrule(lr){2-3}\cmidrule(lr){4-5}\cmidrule(lr){6-7}\cmidrule(lr){8-9}
model & \numhead{RTN} & \numhead{SR} & \numhead{RTN} & \numhead{SR} & \numhead{RTN} & \numhead{SR} & \numhead{RTN} & \numhead{SR}\\\midrule
Linear & 0.1246 & 0.2229 & 0.0722 & 0.0273 & 0.3317 & 0.1349 & 0.0937 & \textbf{0.0102}\\
Mamba-2 & 0.0026 & 0.0015 & 0.0071 & 0.0019 & 0.0105 & \textbf{0.0005} & 0.0118 & 0.0010\\
GLA & 0.0002 & \textbf{0.0001} & 0.0040 & 0.0007 & 0.0047 & 0.0005 & 0.0064 & 0.0007\\
GDN & \textbf{0.0000} & 0.0001 & 0.0017 & 0.0016 & 0.0048 & 0.0007 & 0.0052 & 0.0013\\
KDA & \textbf{0.0002} & \textbf{0.0002} & 0.0031 & 0.0006 & 0.0071 & 0.0005 & 0.0073 & 0.0009\\
\midrule
Qwen-3.5 & 0.0005 & \textbf{0.0001} & 0.0010 & 0.0002 & 0.0100 & 0.8767 & 0.0027 & 0.0004\\
Nemotron-H & 0.0006 & \textbf{0.0001} & 0.0009 & 0.0015 & 0.0193 & 6.5132 & 0.0037 & 0.0429\\
Kimi-Linear & 0.0005 & \textbf{0.0000} & 0.0015 & 0.0001 & 0.0013 & 0.0002 & 0.0037 & 0.0001\\
\bottomrule
\end{tabular}
\end{table}

\section{Scale feedback and query alignment under rotations}
\label{app:sq-scale-feedback}

\begin{figure}[ht]
\centering
\input{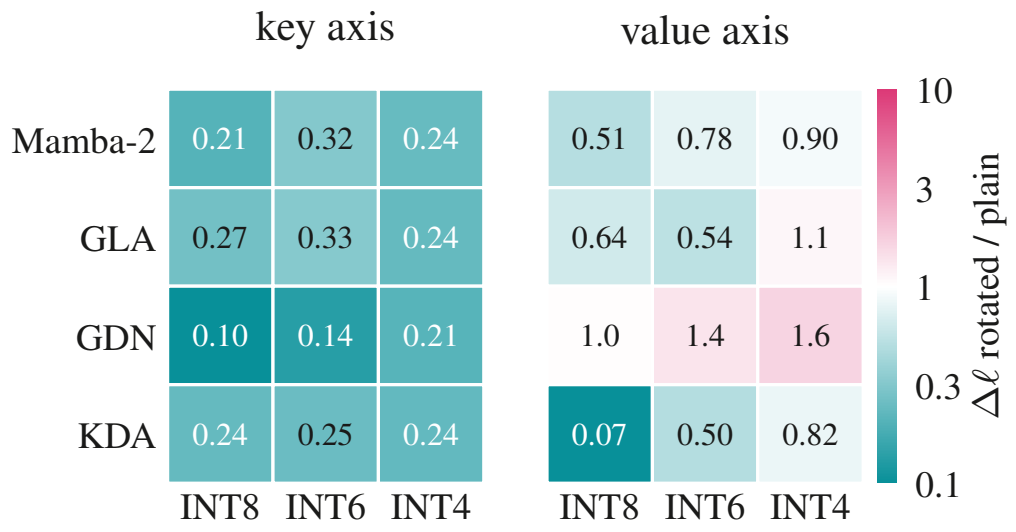}
\caption{$\Delta\ell$ with rotation divided by $\Delta\ell$ without it, as in Figure~\ref{fig:sq-rotation-effect}(a), with every scale group taking its scale from the FP32 copy of the state.}
\label{fig:sq-rot-oracle-full}
\end{figure}

\textbf{Scale feedback.} Section~\ref{sec:sq-rotations} partially attributes the $\Delta\ell$ penalty of $\bR_k$ under value-axis grouping to the quantization scale, which is computed from the stored state and therefore carries the accumulated rounding error. Table~\ref{tab:sq-rot-clean-scale-e2e} reports an intervention. Every scale group takes its scale from an FP32 copy of the state that is advanced with the same operands but never rounded. Figure~\ref{fig:sq-rot-oracle-full} shows the resulting $\Delta\ell$ ratios for both grouping axes.

\begin{table}[ht]
\centering
\footnotesize
\setlength{\tabcolsep}{2pt}
\caption{Median cross-entropy increase $\Delta\ell$ under within-group rotations for SR with symmetric grids, with every scale group taking its scale from the FP32 copy of the state; mixer-zoo models. Bold marks best per precision.}
\label{tab:sq-rot-clean-scale-e2e}
\begin{tabular}{@{}lrrrrrrrrrrrr@{}}\toprule
& \multicolumn{4}{c}{INT8} & \multicolumn{4}{c}{INT6} & \multicolumn{4}{c}{INT4}\\
\cmidrule(lr){2-5}\cmidrule(lr){6-9}\cmidrule(lr){10-13}
model & \numhead{value} & \numhead{$+\bR_k$} & \numhead{key} & \numhead{$+\bR_v$} & \numhead{value} & \numhead{$+\bR_k$} & \numhead{key} & \numhead{$+\bR_v$} & \numhead{value} & \numhead{$+\bR_k$} & \numhead{key} & \numhead{$+\bR_v$}\\\midrule
Mamba-2 & 0.0014 & 0.0007 & 0.0005 & \textbf{0.0001} & 0.0365 & 0.0284 & 0.0110 & \textbf{0.0035} & 0.4878 & 0.4407 & 0.2238 & \textbf{0.0541}\\
GLA & 0.0000 & \textbf{0.0000} & 0.0004 & 0.0001 & 0.0021 & \textbf{0.0011} & 0.0096 & 0.0031 & 0.0488 & 0.0538 & 0.1794 & \textbf{0.0432}\\
GDN & 0.0002 & 0.0002 & 0.0009 & \textbf{0.0001} & 0.0048 & 0.0065 & 0.0166 & \textbf{0.0022} & 0.1000 & 0.1644 & 0.3033 & \textbf{0.0640}\\
KDA & 0.0001 & \textbf{0.0000} & 0.0004 & 0.0001 & 0.0047 & \textbf{0.0023} & 0.0100 & 0.0025 & 0.0970 & 0.0799 & 0.2184 & \textbf{0.0517}\\
\bottomrule
\end{tabular}
\end{table}

\begin{figure}[htbp]
\centering
\input{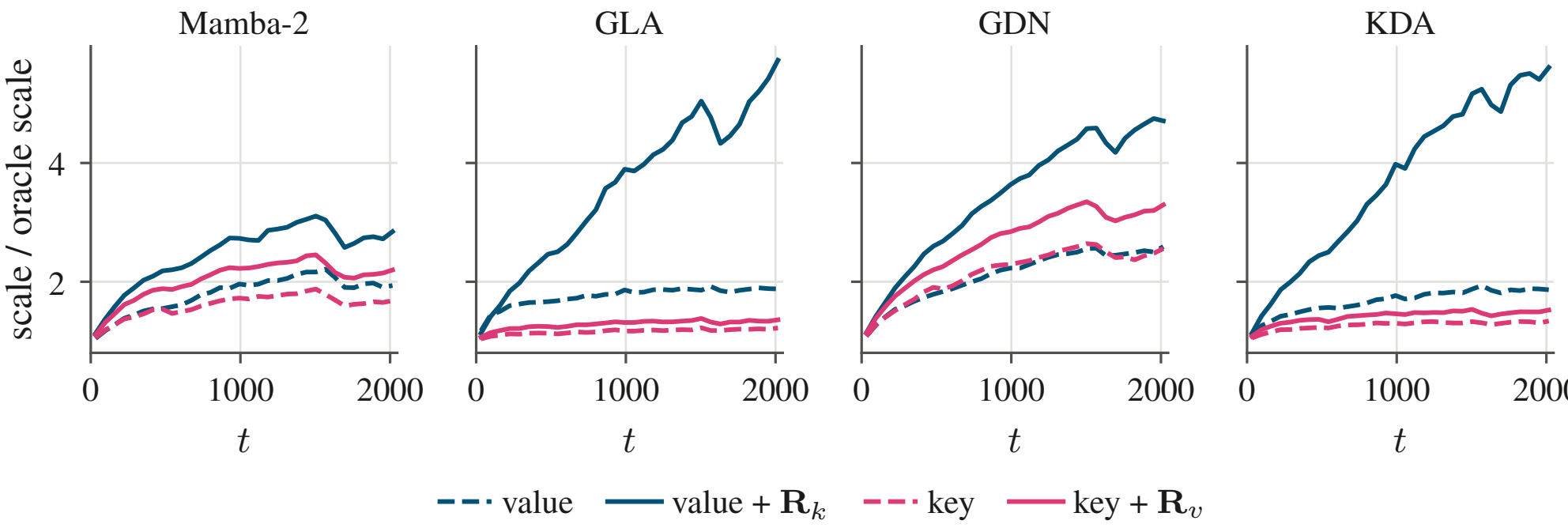}
\caption{Quantization scale over the scale of the oracle copy of the state against the decode step, INT4 SR, plain (dashed) and rotated (solid). Median of the per-step mean over the head's scale groups within 64 steps, then the 90th percentile over heads (24 layers, 384 heads). Value-axis grouping with $\bR_k$, key-axis grouping with $\bR_v$.}
\label{fig:sq-rot-inflation-time}
\end{figure}

Figure~\ref{fig:sq-rot-inflation-time} extends the scale-inflation trajectories in Figure~\ref{fig:sq-rotation-effect}(b) to all four decayed recurrences. Rotations generally amplify scale inflation more under value-axis grouping than under key-axis grouping.

\paragraph{Query alignment.}
We replay the first eight recurrent layers on one $2048$-token \WT{} sequence using INT4 SR with oracle scales. We sample all heads at four decode steps, spaced $512$ steps apart starting at step $256$. At each snapshot, we evaluate quantization error and alignment statistics from exact SR moments at each run's own state before rounding. We average each head's statistics over the snapshots and classify it as worse under $\bR_k$ if rotation increases its expected squared readout error.

\begin{figure}[htbp]
\centering
\input{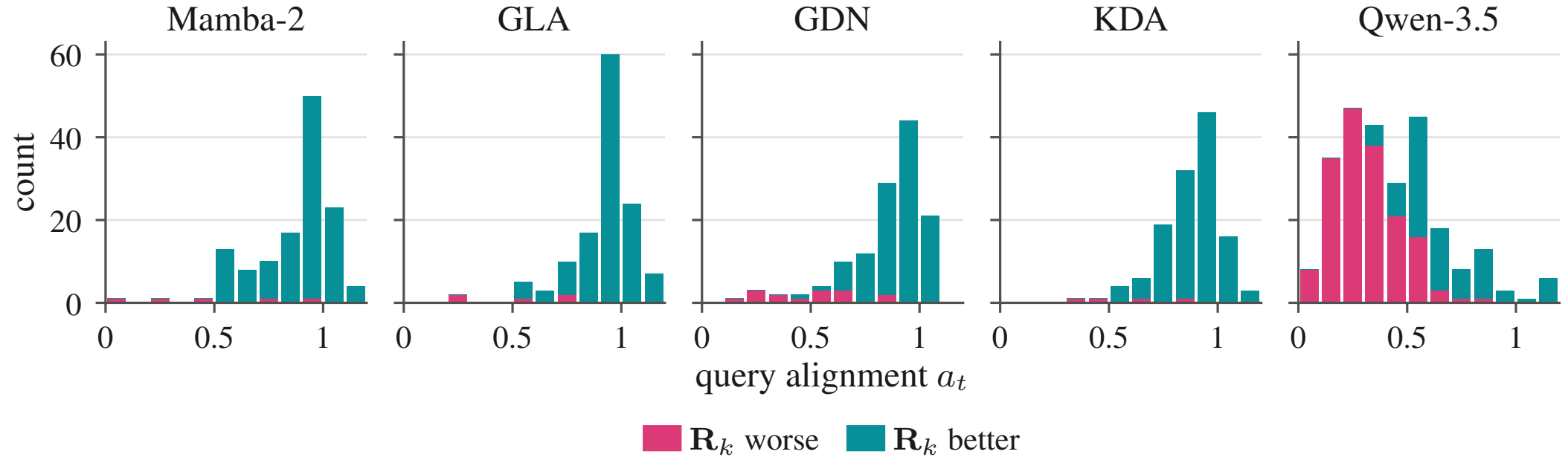}
\caption{As Figure~\ref{fig:sq-rot-query-alignment}, more models.}
\label{fig:sq-rot-query-alignment-full}
\end{figure}

Figure~\ref{fig:sq-rot-query-alignment-full} shows that rotation hurts most often in heads whose plain error has low query alignment. This pattern is pronounced in Qwen-3.5, whereas most Mamba-2 heads have plain alignment near unity and benefit from rotation.

\section{End-to-End Evaluation}
\label{app:end-to-end}

This section extends the decode results presented in Section~\ref{sec:recipes}.

\subsection{Configuration comparisons and promotion sweeps}

Table~\ref{tab:e2e-ranking} compares value- and key-axis grouping under stochastic rounding.

\begin{table}[htbp]
\centering
\caption{GSM8K accuracy (\%) under SR. FP32 is the unquantized-state baseline.}
\label{tab:e2e-ranking}
\centering
\footnotesize
\setlength{\tabcolsep}{2.5pt}
\renewcommand{\arraystretch}{1.10}
\begin{tabular}{@{}lrrrrrrr@{}}
\toprule
& & \multicolumn{3}{c}{value} & \multicolumn{3}{c}{key}\\
\cmidrule(lr){3-5}\cmidrule(lr){6-8}
model & \numhead{FP32} & \numhead{INT8} & \numhead{INT6} & \numhead{INT4} & \numhead{INT8} & \numhead{INT6} & \numhead{INT4}\\
\midrule
Qwen-3.5 & 95.9 & 96.4 & 95.3 & 8.0 & 47.8 & 1.5 & 0.7\\
Nemotron-H & 92.3 & 92.0 & 91.2 & 3.5 & 0.9 & 1.4 & 1.1\\
\bottomrule
\end{tabular}
\end{table}

Table~\ref{tab:aime-cube} compares RTN and SR at INT8 state on AIME. Stochastic rounding outperforms round-to-nearest for all three models.

\begin{table}[htbp]
\caption{AIME avg@32 accuracy (\%) at INT8 state with symmetric absmax scales. Value-axis grouping unless stated otherwise.}
\label{tab:aime-cube}
\centering
\small
\setlength{\tabcolsep}{6pt}
\begin{tabular}{@{}lrrrr@{}}
\toprule
 & \multicolumn{1}{c}{Nemotron-H} & \multicolumn{1}{c}{Qwen-3.5} & \multicolumn{2}{c}{Kimi-Linear} \\
\cmidrule(lr){2-2}\cmidrule(lr){3-3}\cmidrule(lr){4-5}
rounding & \numhead{value axis} & \numhead{value axis} & \numhead{value axis} & \numhead{key axis} \\
\midrule
RTN & $40.3$ & $54.0$ & $56.1$ & $53.9$ \\
SR  & $48.4$ & $79.5$ & $62.6$ & $60.5$ \\
\addlinespace
FP32 & $51.2$ & $82.3$ & $63.6$ & $63.6$ \\
\bottomrule
\end{tabular}

\end{table}

\begin{figure}[htbp]
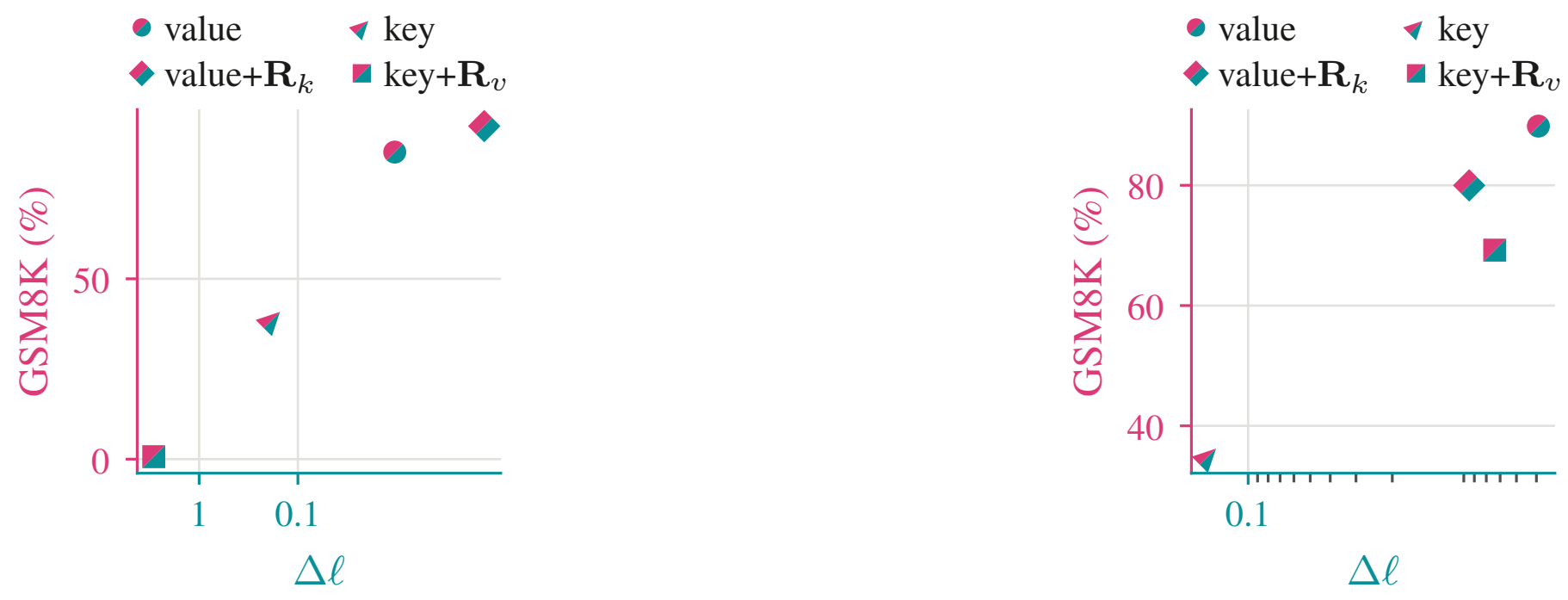

\centering
\begin{subfigure}[t]{0.42\linewidth}
\centering
\input{media/rank_slope_int6_nemo.pgf}
\caption{Nemotron-H.}
\label{fig:rank-slope-rtn-nemo}
\end{subfigure}\hfill
\begin{subfigure}[t]{0.42\linewidth}
\centering
\input{media/rank_slope_int6_app.pgf}
\caption{Qwen-3.5.}
\label{fig:rank-slope-rtn-qwen}
\end{subfigure}
\caption{Teacher-forced $\Delta\ell$ versus full-test GSM8K accuracy for four INT6 state configurations under round-to-nearest with symmetric grids. The horizontal axis is reversed so that better results lie to the right.}
\label{fig:rank-slope-rtn}
\end{figure}

\paragraph{Proxy-to-generation transfer.}
\label{app:rank-transfer-rtn}
For sequence $n$ and token $t$, let $\ell^{(n,t)}_{\mathrm{quant}}$ and $\ell^{(n,t)}_{\mathrm{FP}}$ denote the token-level cross-entropies obtained with quantized and FP32 states, respectively. We define
\begin{equation*}
    \Delta\ell \coloneqq
    \operatorname*{median}_{n,t}
    \bigl(
        \ell^{(n,t)}_{\mathrm{quant}}
        -
        \ell^{(n,t)}_{\mathrm{FP}}
    \bigr),
\end{equation*}
where the median is taken over all sequence--token pairs.
Figure~\ref{fig:rank-slope-rtn} compares the teacher-forced proxy $\Delta\ell$ with full-test GSM8K accuracy for four INT6 configurations under round-to-nearest with symmetric grids. The two metrics follow the same overall trend.

\paragraph{Calibration-to-generation transfer under promotion.}
\label{app:rank-slope}

\begin{figure}[htbp]
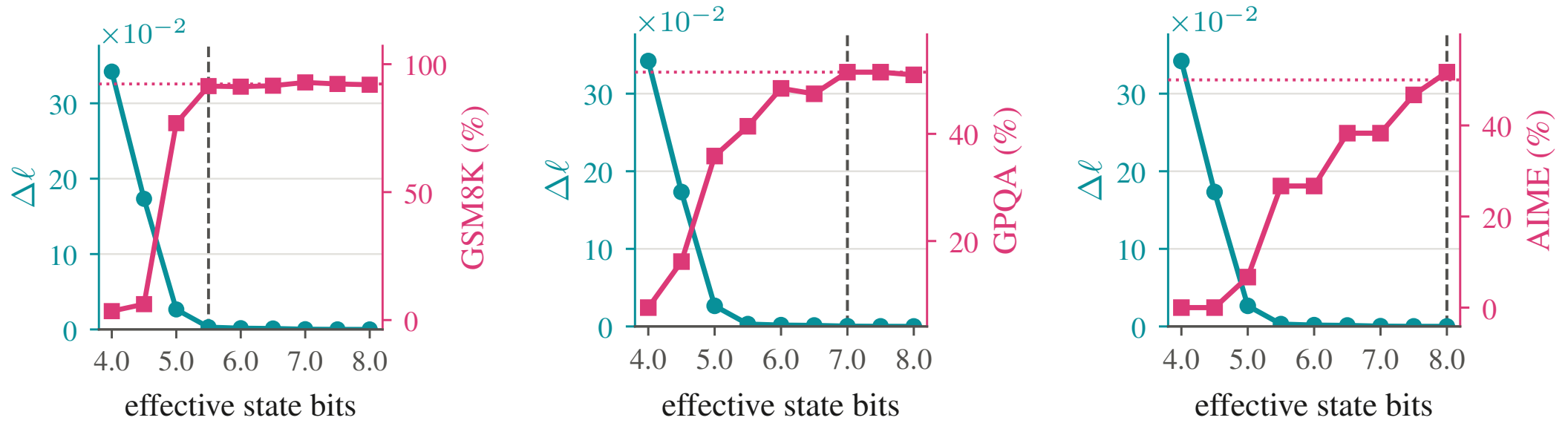
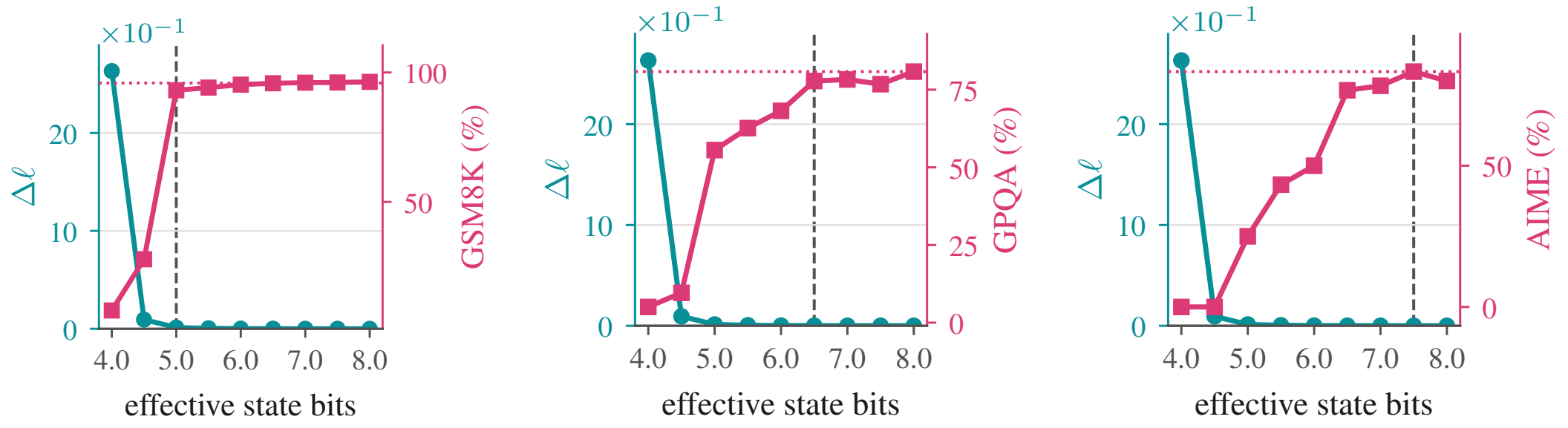

\centering
\begin{subfigure}[t]{\linewidth}
\centering
\resizebox{0.32\linewidth}{!}{\input{media/dce_gen_promotion_nemo.pgf}}\hfill
\resizebox{0.32\linewidth}{!}{\input{media/gpqa_effbits_nemo.pgf}}\hfill
\resizebox{0.32\linewidth}{!}{\input{media/aime_effbits_nemo.pgf}}
\caption{Nemotron-H.}
\label{fig:e2e-transfer}
\end{subfigure}
\vspace{0.5\baselineskip}
\begin{subfigure}[t]{\linewidth}
\centering
\resizebox{0.32\linewidth}{!}{\input{media/dce_gen_promotion_app.pgf}}\hfill
\resizebox{0.32\linewidth}{!}{\input{media/gpqa_effbits_qwen.pgf}}\hfill
\resizebox{0.32\linewidth}{!}{\input{media/aime_effbits_qwen.pgf}}
\caption{Qwen-3.5.}
\label{fig:e2e-transfer-app}
\end{subfigure}
\caption{Promotion sweeps on GSM8K, GPQA, and AIME. Teal denotes $\Delta\ell$, magenta accuracy, dotted lines the FP32-state baseline, and dashed lines the minimum precision under the $5\%$ criterion.}
\label{fig:e2e-transfer-joint}
\end{figure}

Figure~\ref{fig:e2e-transfer-joint} compares the teacher-forced proxy $\Delta\ell$ with generation accuracy as decay-ranked heads are promoted.

Table~\ref{tab:aime-sweeps} gives the complete accuracy sweeps. Its entries are exactly the accuracy points plotted in Figure~\ref{fig:e2e-transfer-joint} and Figure~\ref{fig:promo-relative}.

\begin{table}[htbp]
\centering
\small
\setlength{\tabcolsep}{4.5pt}
\caption{Generation accuracy (\%) as a function of effective bits. GPQA and AIME entries are pass@1.}
\label{tab:aime-sweeps}
\begin{tabular}{@{}rrrrrrr@{}}
\toprule
& \multicolumn{3}{c}{Qwen-3.5} & \multicolumn{3}{c}{Nemotron-H} \\
\cmidrule(lr){2-4}\cmidrule(lr){5-7}
\numhead{effective bits} & \numhead{GSM8K} & \numhead{GPQA} & \numhead{AIME} & \numhead{GSM8K} & \numhead{GPQA} & \numhead{AIME} \\
\midrule
$4.0$  & 8.0 & 5.1 & 0.0 & 3.5 & 7.6 & 0.0 \\
$4.5$  & 27.8 & 9.6 & 0.0 & 6.2 & 16.2 & 0.0 \\
$5.0$  & 93.1 & 55.6 & 25.0 & 77.0 & 35.9 & 6.7 \\
$5.5$  & 94.2 & 62.6 & 43.3 & 91.4 & 41.4 & 26.7 \\
$6.0$  & 95.3 & 68.2 & 50.0 & 91.2 & 48.5 & 26.7 \\
$6.5$  & 95.8 & 77.8 & 76.7 & 91.6 & 47.5 & 38.3 \\
$7.0$  & 96.1 & 78.3 & 78.3 & 92.9 & 51.5 & 38.3 \\
$7.5$  & 96.1 & 76.8 & 83.3 & 92.3 & 51.5 & 46.7 \\
$8.0$  & 96.4 & 80.8 & 80.0 & 92.0 & 51.0 & 51.7 \\
\addlinespace
FP32   & 95.9 & 80.8 & 83.3 & 92.3 & 51.5 & 50.0 \\
\bottomrule
\end{tabular}
\end{table}

\subsection{Generation length}
\label{app:generation-length}

Table~\ref{tab:genlen} shows that state INT8 can either lengthen or shorten the median response, depending on the model and benchmark. The accuracy results are therefore not an artifact of systematically shorter quantized generations. The AIME comparison below examines stochastic and nearest rounding.

\begin{table}[htbp]
\centering
\small
\setlength{\tabcolsep}{3.5pt}
\caption{Median response length in generated tokens for the FP32-state baseline and state INT8 at W/A $=16/16$. $\Delta$ is relative to FP32 state.}
\label{tab:genlen}
\begin{tabular}{@{}lrrrrrrrrr@{}}
\toprule
& \multicolumn{3}{c}{Qwen-3.5} & \multicolumn{3}{c}{Nemotron-H} & \multicolumn{3}{c}{Kimi-Linear} \\
\cmidrule(lr){2-4}\cmidrule(lr){5-7}\cmidrule(lr){8-10}
benchmark & \numhead{FP32} & \numhead{INT8} & \numhead{$\Delta$ (\%)} & \numhead{FP32} & \numhead{INT8} & \numhead{$\Delta$ (\%)} & \numhead{FP32} & \numhead{INT8} & \numhead{$\Delta$ (\%)} \\
\midrule
GSM8K & 2096 & 2212 & $+5.5$ & 592 & 588 & $-0.7$ & 89 & 88 & $-1.1$ \\
MMLU-Pro & 4050 & 4152 & $+2.5$ & 1112 & 1132 & $+1.8$ & 2080 & 2184 & $+5.0$ \\
GPQA & 9169 & 9134 & $-0.4$ & 5449 & 5312 & $-2.5$ & 7249 & 7356 & $+1.5$ \\
AIME & 25223 & 24950 & $-1.1$ & 10994 & 10964 & $-0.3$ & 11839 & 11516 & $-2.7$ \\
RULER & 16 & 16 & $0.0$ & 21 & 20 & $-4.8$ & 12 & 12 & $0.0$ \\
\bottomrule
\end{tabular}
\end{table}

Table~\ref{tab:aime-genlen} reports the generation lengths of the AIME runs behind Table~\ref{tab:aime-cube}. For Qwen-3.5 and Nemotron-H, stochastic rounding keeps mean response length close to the FP32-state baseline, whereas round-to-nearest lengthens responses and increases budget exhaustion. Kimi-Linear instead produces shorter responses under round-to-nearest with value-axis grouping, so budget exhaustion does not explain its accuracy loss. For all three models, stochastic rounding keeps mean response length closer to the FP32-state baseline than round-to-nearest.

\begin{table}[htbp]
\centering
\small
\setlength{\tabcolsep}{5pt}
\caption{Mean generated AIME tokens (and traces hitting the $65536$-token budget) for the arms of Table~\ref{tab:aime-cube}. Same generations as the reported accuracies. $n{=}1920$ per cell ($60$ problems $\times$ $32$ samples).}
\label{tab:aime-genlen}
\begin{tabular}{lrrrrrrrr}\toprule
& \multicolumn{2}{c}{Qwen-3.5} & \multicolumn{2}{c}{Nemotron-H} & \multicolumn{2}{c}{Kimi-Linear (value)} & \multicolumn{2}{c}{Kimi-Linear (key)}\\
\cmidrule(lr){2-3}\cmidrule(lr){4-5}\cmidrule(lr){6-7}\cmidrule(lr){8-9}
rounding & \numhead{tokens} & \numhead{capped} & \numhead{tokens} & \numhead{capped} & \numhead{tokens} & \numhead{capped} & \numhead{tokens} & \numhead{capped}\\\midrule
FP32 state & 29927 & 240 & 12106 & 26 & 22565 & 318 & 22565 & 318\\
\midrule
RTN & 46437 & 792 & 30136 & 459 & 21398 & 244 & 25728 & 322\\
SR & 30362 & 275 & 12470 & 36 & 22121 & 316 & 22507 & 365\\
\bottomrule
\end{tabular}
\end{table}

\section{Implementation and Efficiency}
\label{app:throughput}

\subsection{Kernel Throughput}
\label{app:kernel-design}

\begin{figure}[!htbp]
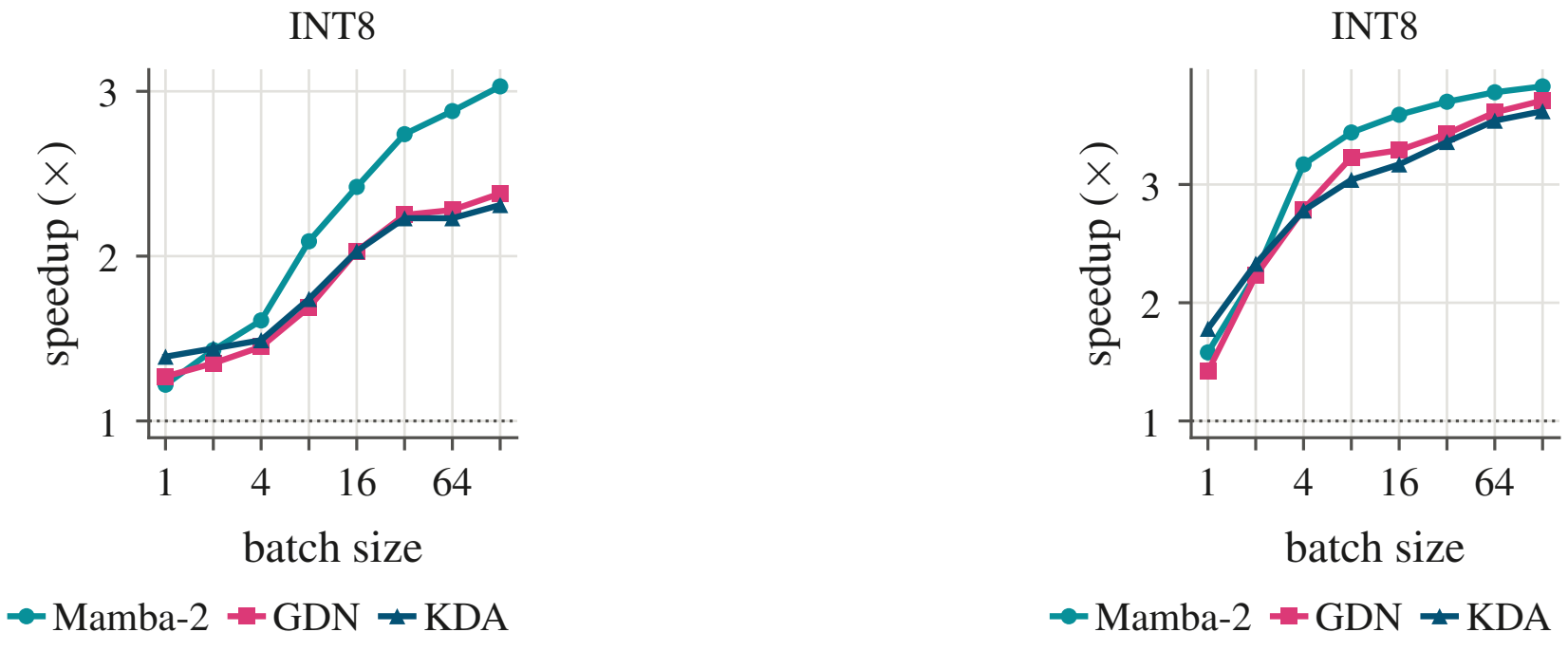

\centering
\begin{subfigure}[b]{0.48\linewidth}
\centering
\input{media/speed_kernel_h100.pgf}
\caption{H100.}
\label{fig:speed-kernel-h100-all}
\end{subfigure}\hfill
\begin{subfigure}[b]{0.48\linewidth}
\centering
\input{media/speed_kernel_rtx.pgf}
\caption{RTX~6000.}
\label{fig:speed-kernel-rtx}
\end{subfigure}
\caption{Speedup of vLLM mixer kernels using INT8 versus FP32 states.}
\label{fig:speed-kernel-all}
\end{figure}

The state-update kernels tile each head's state $\bS\in\mathbb R^{d_v,d_k}$ over value rows. Value-axis scales are local to each tile, so quantization and dequantization fuse into the decode kernel. Key-axis scales are not local to these tiles, complicating an efficient implementation. We benchmark vLLM's FP32-state decode kernels against our INT8-state variants with equal state element counts: GDN and KDA use 32 heads with $d_k=d_v=128$, while Mamba-2 uses 64 heads with $d_k=128$ and $d_v=64$. The INT8 variants use stochastic rounding. A CUDA graph cycles through a working set larger than the L2 cache, and each kernel is launch-tuned at every batch size. The selected configurations are then timed in seven repetitions, and speedup is the ratio of median kernel times. Figure~\ref{fig:speed-kernel-all} compares the H100 with the lower-bandwidth RTX~6000. The larger gains on the RTX~6000 are consistent with state traffic accounting for a larger share of decode cost.

\subsection{End-to-end throughput}
\label{app:speed-raw}

\label{app:speed-protocol}
We time Nemotron-H-8B-Reasoning-128K and Qwen-3.5 9B with vLLM on a dedicated H100~SXM5, using CUDA graphs and asynchronous decode scheduling. The weight settings are the vendor BF16 checkpoint and rotated, weight-only GPTQ at INT4 with group size~128. Activations remain in BF16. Online rotations remain active for INT4 weights during timing and use Dao-AILab's fast CUDA Hadamard kernels.\footnote{\url{https://github.com/Dao-AILab/fast-hadamard-transform}} We compare FP32 recurrent states with INT8 states using stochastic rounding and value-axis grouping.

Both sweeps use fixed batches of $c$ requests with synthetic token prompts and greedy decoding, forcing 32 output tokens per request. All prefills and the first output token finish before timing starts. We then measure engine wall time for the remaining 31 decode steps, synchronizing the GPU at both timing boundaries, and divide by 31 to obtain TPOT. Prefill, tokenization and HTTP overhead are excluded. %

The batch-size sweep uses $c\in\{1,2,4,\dots,256\}$ (up to 64 with INT4 weights) and two-token prompts. The context-length sweep uses $c\in\{2,4,8,16,32\}$ and prompt lengths $\{16,32,512,1024,4096\}$, plus 8160 tokens for Nemotron-H or 8192 for Qwen-3.5. Each cell has two warmup batches followed by 16 measured batches per repetition. FP32 and INT8 states are measured in seven paired repetitions on the same GPU, with fresh engine processes, matching prompts and seeds, and alternating measurement order. Reported TPOT is the median of the seven per-repetition means. Speedup is the ratio of the corresponding median TPOTs.

Launch configurations are tuned separately for the FP32 and INT8 decode kernels at each batch size. Attention block sizes are selected automatically, except in the INT4-weight pinned-block control, which fixes both state formats to the FP32-state block size: 1040 tokens for Nemotron-H and 528 for Qwen-3.5.

\begin{figure}[!htbp]
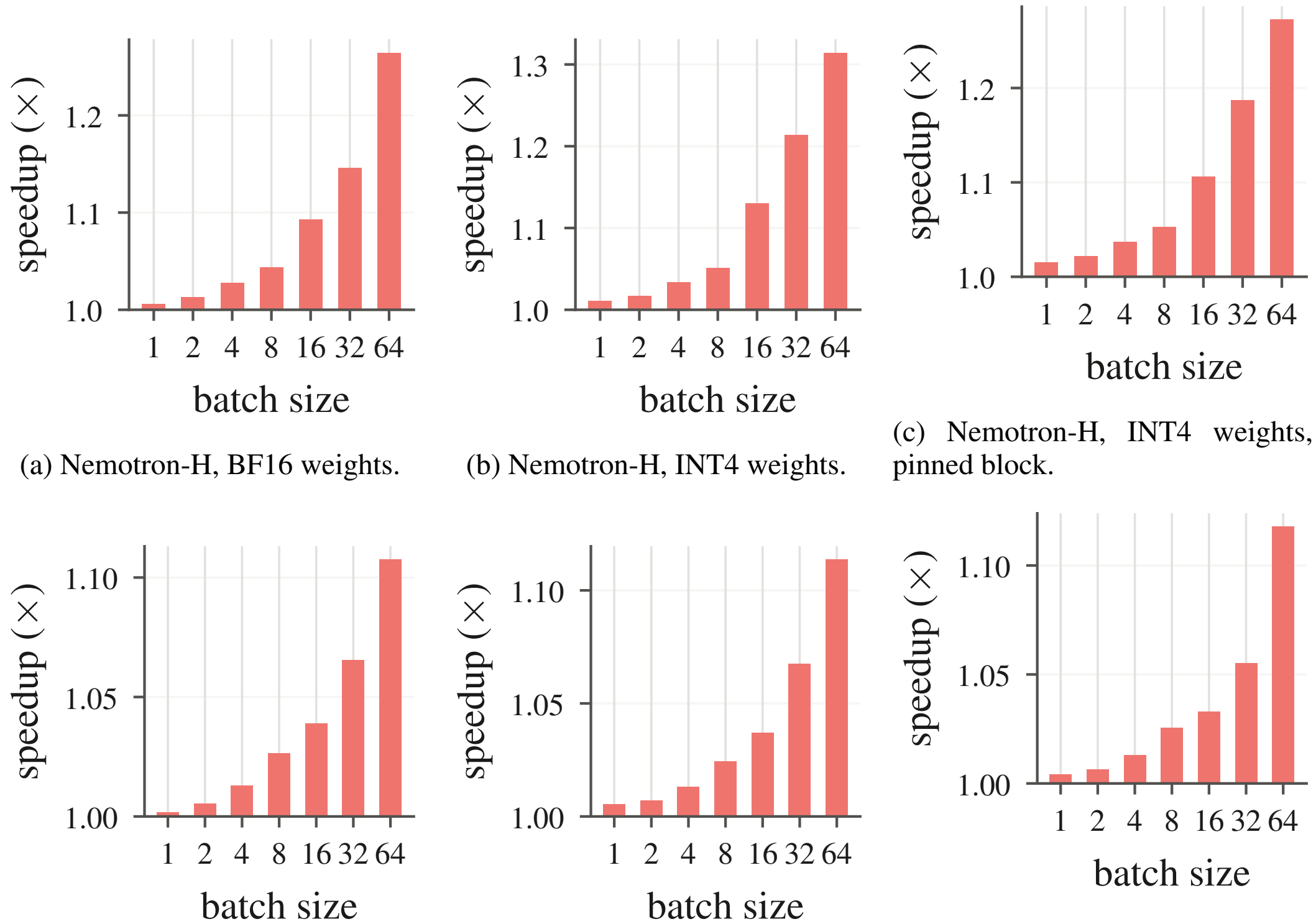

\centering
\begin{subfigure}{0.32\linewidth}
\resizebox{\linewidth}{!}{\input{media/speed_vllm_bars_bf16_a3nemo.pgf}}
\caption{Nemotron-H, BF16 weights.}
\label{fig:speed-vllm-bars-nemo-bf16}
\end{subfigure}
\hfill
\begin{subfigure}{0.32\linewidth}
\resizebox{\linewidth}{!}{\input{media/speed_vllm_bars_w4_a3nemo.pgf}}
\caption{Nemotron-H, INT4 weights.}
\label{fig:speed-vllm-bars-nemo-w4}
\end{subfigure}
\hfill
\begin{subfigure}{0.32\linewidth}
\resizebox{\linewidth}{!}{\input{media/speed_vllm_bars_w4_a3nemoPIN.pgf}}
\caption{Nemotron-H, INT4 weights, pinned block.}
\label{fig:speed-vllm-bars-nemo-w4-pinned}
\end{subfigure}
\par\medskip
\begin{subfigure}{0.32\linewidth}
\resizebox{\linewidth}{!}{\input{media/speed_vllm_bars_bf16_a3qwen.pgf}}
\caption{Qwen-3.5, BF16 weights.}
\label{fig:speed-vllm-bars-qwen-bf16}
\end{subfigure}
\hfill
\begin{subfigure}{0.32\linewidth}
\resizebox{\linewidth}{!}{\input{media/speed_vllm_bars_w4_a3qwen.pgf}}
\caption{Qwen-3.5, INT4 weights.}
\label{fig:speed-vllm-bars-qwen-w4}
\end{subfigure}
\hfill
\begin{subfigure}{0.32\linewidth}
\resizebox{\linewidth}{!}{\input{media/speed_vllm_bars_w4_a3qwenPIN.pgf}}
\caption{Qwen-3.5, INT4 weights, pinned block.}
\label{fig:speed-vllm-bars-qwen-w4-pinned}
\end{subfigure}
\caption{INT8-state TPOT speedup over an FP32 state, by batch size. Top row Nemotron-H, bottom row Qwen-3.5. Columns BF16 weights, INT4 weights, and INT4 weights with the attention block size pinned to its FP32-state value.}
\label{fig:speed-vllm-bars}
\end{figure}

Figure~\ref{fig:speed-vllm-bars} visualizes INT8-state speedup over an FP32 state at the same weight setting. Table~\ref{tab:vllm-absolute} gives the corresponding TPOT for the automatic-block sweeps.

\begin{table}[!htbp]
\caption{Absolute TPOT (ms), reported as the median of seven repetitions.}
\label{tab:vllm-absolute}
\centering
\small
\setlength{\tabcolsep}{3.8pt}
\begin{tabular}{@{}lrrrrrrrrr@{}}
\toprule
W/A/S & \multicolumn{9}{c}{batch size} \\
\cmidrule(lr){2-10}
& \numhead{1} & \numhead{2} & \numhead{4} & \numhead{8} & \numhead{16} & \numhead{32} & \numhead{64} & \numhead{128} & \numhead{256} \\
\midrule
\multicolumn{10}{@{}l}{\textit{Nemotron-H}} \\
$16/16/32$ & 6.31 & 6.41 & 6.62 & 6.96 & 7.63 & 8.78 & 11.40 & 16.83 & 29.35 \\
$16/16/8$ & 6.27 & 6.32 & 6.44 & 6.67 & 6.98 & 7.66 & 9.01 & 11.18 & 17.32 \\
\addlinespace[2pt]
$4/16/32$ & 4.46 & 4.52 & 4.73 & 4.99 & 5.62 & 6.77 & 9.86 & -- & -- \\
$4/16/8$ & 4.41 & 4.44 & 4.57 & 4.74 & 4.98 & 5.57 & 7.50 & -- & -- \\
\midrule
\multicolumn{10}{@{}l}{\textit{Qwen-3.5}} \\
$16/16/32$ & 6.99 & 7.05 & 7.15 & 7.39 & 7.77 & 8.44 & 9.60 & 12.19 & 18.23 \\
$16/16/8$ & 6.98 & 7.01 & 7.06 & 7.20 & 7.48 & 7.92 & 8.67 & 9.99 & 14.21 \\
\addlinespace[2pt]
$4/16/32$ & 5.70 & 5.79 & 5.89 & 6.11 & 6.46 & 7.22 & 9.57 & -- & -- \\
$4/16/8$ & 5.67 & 5.74 & 5.81 & 5.96 & 6.23 & 6.76 & 8.59 & -- & -- \\
\bottomrule
\end{tabular}
\end{table}

\begin{figure}[!htbp]
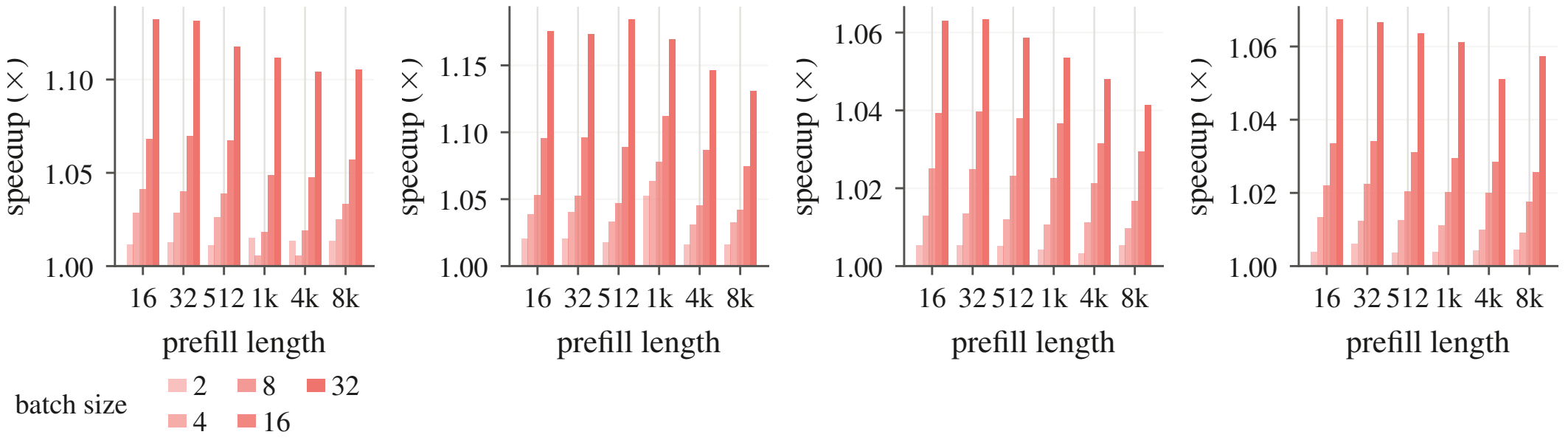

\centering
\begin{subfigure}{0.24\linewidth}
\resizebox{\linewidth}{!}{\input{media/speed_vllm_b3_nemo_bf16_auto.pgf}}
\caption{Nemotron-H, BF16 weights.}
\label{fig:speed-vllm-dilution-nemo-bf16}
\end{subfigure}
\hfill
\begin{subfigure}{0.24\linewidth}
\resizebox{\linewidth}{!}{\input{media/speed_vllm_b3_nemo_state_only_auto.pgf}}
\caption{Nemotron-H, INT4 weights.}
\label{fig:speed-vllm-dilution-nemo-w4}
\end{subfigure}
\hfill
\begin{subfigure}{0.24\linewidth}
\resizebox{\linewidth}{!}{\input{media/speed_vllm_b3_qwen_bf16_auto.pgf}}
\caption{Qwen-3.5, BF16 weights.}
\label{fig:speed-vllm-dilution-qwen-bf16}
\end{subfigure}
\hfill
\begin{subfigure}{0.24\linewidth}
\resizebox{\linewidth}{!}{\input{media/speed_vllm_b3_qwen_state_only_auto.pgf}}
\caption{Qwen-3.5, INT4 weights.}
\label{fig:speed-vllm-dilution-qwen-w4}
\end{subfigure}
\caption{INT8-state TPOT speedup over the same model and weight setting with an FP32 state, across prefill length. Bar shade gives the batch size.}
\label{fig:speed-vllm-dilution-app}
\end{figure}

Figure~\ref{fig:speed-vllm-dilution-app} and Table~\ref{tab:vllm-dilution} report the context sweep.

\begin{table}[!htbp]
\centering
\small
\setlength{\tabcolsep}{3.5pt}
\caption{Absolute TPOT (ms, median of seven repetitions) as prefill length increases. The 8k column uses the longest benchmarked prompt: $8160$ tokens for Nemotron-H and $8192$ for Qwen-3.5.}
\label{tab:vllm-dilution}
\begin{tabular}{@{}lrrrrrrr@{}}
\toprule
W/A/S & \numhead{batch size} & \multicolumn{6}{c}{prefill length} \\
\cmidrule(lr){3-8}
& & \numhead{16} & \numhead{32} & \numhead{512} & \numhead{1024} & \numhead{4096} & \numhead{8k} \\
\midrule
\multicolumn{8}{@{}l}{\textit{Nemotron-H}} \\
\multirow{5}{*}{$16/16/32$} & 2 & 6.4 & 6.4 & 6.5 & 6.4 & 6.3 & 6.4 \\
 & 4 & 6.6 & 6.6 & 6.7 & 6.6 & 6.7 & 6.8 \\
 & 8 & 7.0 & 7.0 & 7.1 & 6.9 & 7.2 & 7.4 \\
 & 16 & 7.7 & 7.7 & 7.8 & 7.8 & 8.1 & 8.5 \\
 & 32 & 8.9 & 9.0 & 9.0 & 9.1 & 9.7 & 10.5 \\
\addlinespace[2pt]
\multirow{5}{*}{$16/16/8$} & 2 & 6.3 & 6.3 & 6.4 & 6.3 & 6.2 & 6.3 \\
 & 4 & 6.5 & 6.5 & 6.5 & 6.6 & 6.7 & 6.7 \\
 & 8 & 6.7 & 6.7 & 6.8 & 6.8 & 7.0 & 7.1 \\
 & 16 & 7.2 & 7.2 & 7.3 & 7.4 & 7.7 & 8.1 \\
 & 32 & 7.9 & 7.9 & 8.1 & 8.2 & 8.8 & 9.5 \\
\addlinespace[2pt]
\multirow{5}{*}{$4/16/32$} & 2 & 4.5 & 4.5 & 4.6 & 4.6 & 4.6 & 4.6 \\
 & 4 & 4.7 & 4.7 & 4.8 & 4.9 & 4.8 & 4.9 \\
 & 8 & 4.9 & 5.0 & 5.1 & 5.2 & 5.2 & 5.4 \\
 & 16 & 5.6 & 5.6 & 5.8 & 5.8 & 6.0 & 6.4 \\
 & 32 & 6.7 & 6.8 & 6.9 & 6.9 & 7.5 & 8.3 \\
\addlinespace[2pt]
\multirow{5}{*}{$4/16/8$} & 2 & 4.4 & 4.4 & 4.5 & 4.4 & 4.5 & 4.5 \\
 & 4 & 4.5 & 4.5 & 4.7 & 4.6 & 4.7 & 4.8 \\
 & 8 & 4.7 & 4.7 & 4.9 & 4.8 & 5.0 & 5.2 \\
 & 16 & 5.1 & 5.1 & 5.3 & 5.2 & 5.5 & 6.0 \\
 & 32 & 5.7 & 5.8 & 5.9 & 5.9 & 6.6 & 7.4 \\
\midrule
\multicolumn{8}{@{}l}{\textit{Qwen-3.5}} \\
\multirow{5}{*}{$16/16/32$} & 2 & 6.8 & 6.8 & 6.8 & 6.9 & 7.0 & 7.1 \\
 & 4 & 6.9 & 6.9 & 7.0 & 7.1 & 7.2 & 7.5 \\
 & 8 & 7.2 & 7.2 & 7.3 & 7.4 & 7.7 & 8.1 \\
 & 16 & 7.6 & 7.6 & 7.8 & 7.9 & 8.5 & 9.2 \\
 & 32 & 8.2 & 8.3 & 8.5 & 8.7 & 9.9 & 11.4 \\
\addlinespace[2pt]
\multirow{5}{*}{$16/16/8$} & 2 & 6.8 & 6.8 & 6.8 & 6.9 & 7.0 & 7.1 \\
 & 4 & 6.8 & 6.8 & 7.0 & 7.0 & 7.2 & 7.4 \\
 & 8 & 7.0 & 7.0 & 7.2 & 7.2 & 7.6 & 7.9 \\
 & 16 & 7.3 & 7.3 & 7.5 & 7.6 & 8.2 & 9.0 \\
 & 32 & 7.7 & 7.8 & 8.0 & 8.3 & 9.4 & 11.0 \\
\addlinespace[2pt]
\multirow{5}{*}{$4/16/32$} & 2 & 5.8 & 5.8 & 5.9 & 6.0 & 6.1 & 6.2 \\
 & 4 & 5.9 & 5.9 & 6.1 & 6.2 & 6.3 & 6.5 \\
 & 8 & 6.1 & 6.2 & 6.4 & 6.4 & 6.8 & 7.2 \\
 & 16 & 6.5 & 6.5 & 6.7 & 6.9 & 7.5 & 8.3 \\
 & 32 & 7.2 & 7.3 & 7.6 & 7.8 & 8.9 & 10.6 \\
\addlinespace[2pt]
\multirow{5}{*}{$4/16/8$} & 2 & 5.7 & 5.7 & 5.9 & 6.0 & 6.1 & 6.2 \\
 & 4 & 5.8 & 5.8 & 6.0 & 6.1 & 6.2 & 6.5 \\
 & 8 & 6.0 & 6.0 & 6.3 & 6.3 & 6.6 & 7.0 \\
 & 16 & 6.3 & 6.3 & 6.5 & 6.7 & 7.3 & 8.0 \\
 & 32 & 6.8 & 6.8 & 7.1 & 7.4 & 8.5 & 10.0 \\
\bottomrule
\end{tabular}
\end{table}

\begin{figure}[!htbp]
\centering
\input{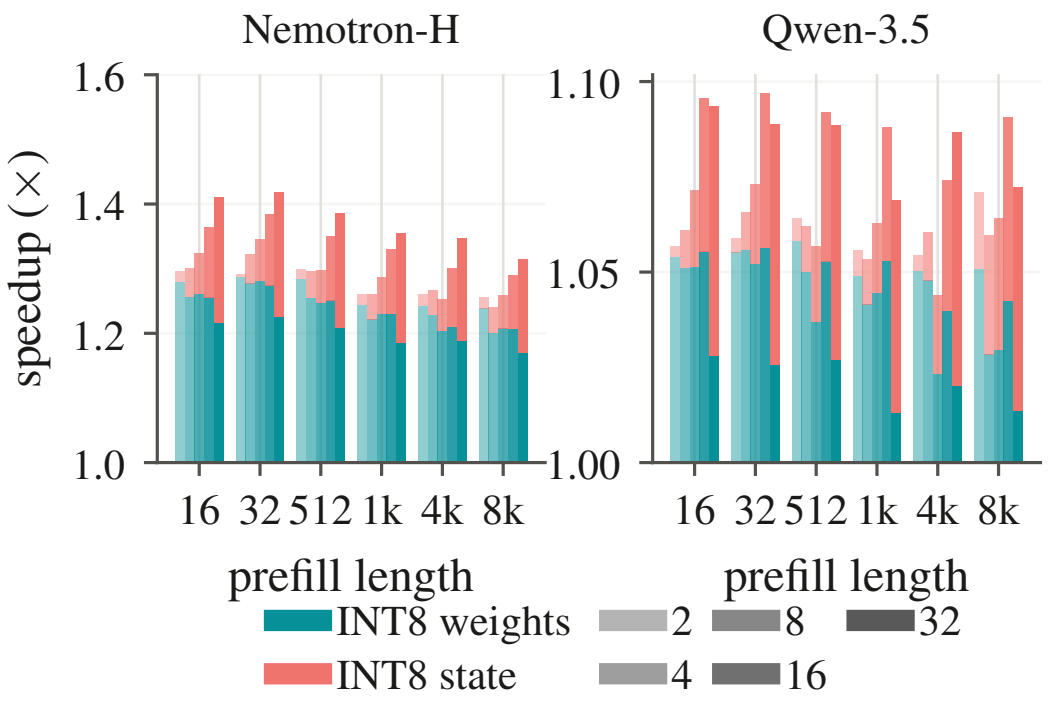}
\caption{INT8-weight timing without online rotations. TPOT speedup over BF16 weights with an FP32 state across prefill length. Teal denotes the INT8-weight contribution, salmon the additional INT8-state contribution, and shade batch size. Exported weights remain rotated.}
\label{fig:speed-vllm-w8-no-rot}
\end{figure}

\FloatBarrier
\subsection*{AI use statement}

In this work, we use generative AI tools to formulate mathematical claims, provide critical ingredients for proving them, to design or give feedback on our methodology and experiments, to create or edit code, implement methods, to propose or refine hypotheses, to support qualitative and thematic data analysis, interpret results, to help develop theoretical models or conceptual frameworks, and to do a literature review.
Additionally, we used generative AI tools for \LaTeX{} formatting, to draft parts of the appendix, and for brainstorming. 
We have not used generative AI tools for generating synthetic datasets, to assist with translation, or to clean or reformat datasets.
We have reviewed all AI-assisted work, checked the code as well as the reformatted \LaTeX{} tables. We take responsibility for the final content of this work, including text, claims or artifacts produced with the aid of generative AI.

\end{document}